\documentclass[final,3p,times]{elsarticle} % for TCS final format
\makeatletter
\def\ps@pprintTitle{%
 \let\@oddhead\@empty
 \let\@evenhead\@empty
 \def\@oddfoot{}%
 \let\@evenfoot\@oddfoot}
\makeatother
\usepackage{amsfonts}
\usepackage{amsmath}
\usepackage{amssymb}
\usepackage{mathtools}
\usepackage{xfrac}
\usepackage{fp}
\usepackage{bm}
\usepackage{bbm}
\usepackage{ifthen}
\usepackage{hyperref}
\usepackage{url}
\usepackage{ dsfont } %for mathds
\usepackage{xspace}

\usepackage{theorem}
\usepackage{makecell}
\usepackage[disable]{todonotes}

\usepackage{subcaption}
\usepackage[autolanguage,np]{numprint}
\usepackage[export]{adjustbox} % <- to aligned subfigures vertically

\usepackage{tikz}
\usetikzlibrary{positioning}
\usetikzlibrary{chains}
\usetikzlibrary{shapes}
\usetikzlibrary{calc}
\usetikzlibrary{arrows.meta}
\usetikzlibrary{overlay-beamer-styles}
\usetikzlibrary{decorations.pathreplacing}

 \newtheorem{theorem}{Theorem} 
  \newtheorem{definition}{Definition} 
 \newtheorem{lemma}{Lemma}
  \newtheorem{axiom}{Axiom}  
   
 \newproof{pf}{Proof}
  \newproof{spf}{Sketch of proof}
\newcommand{\eop}{$\quad\blacksquare$}

\newcommand{\metapath}{meta path}

\newcommand{\ts}[2]{ \text{#1}^{\text{#2}}\text{ copy}}

\newcommand{\reals}{\mathbb{R}}
\newcommand{\naturals}{\mathbb{N}}

\newcommand{\itemset}{I}
\newcommand{\typeset}{T}
\newcommand{\belongs}{\tau}

\newcommand{\richness}{R}
\newcommand{\entropy}{H}
\newcommand{\HHI}{\text{HHI}}
\newcommand{\gini}{\text{Gini}}
\newcommand{\BPI}{\text{BPI}}

\newcommand{\A}{\mathcal{A}}
\newcommand{\R}{\mathcal{R}}
\newcommand{\vertexlabel}{A}
\newcommand{\edgelabel}{R}
\newcommand{\edgemap}{\psi}
\newcommand{\vertexmap}{\varphi}

\newcommand{\Vstart}{V_\text{start}}
\newcommand{\Vend}{V_\text{end}}

\newcommand{\V}{\bm{V}}
\newcommand{\E}{\bm{E}}

\newcommand{\sv}[1]{v_\text{src}(#1)}
\newcommand{\sV}[1]{V_\text{src}(#1)}
\newcommand{\dv}[1]{v_\text{dst}(#1)}
\newcommand{\dV}[1]{V_\text{dst}(#1)}

\newcommand{\calV}{\mathcal{V}}
\newcommand{\calE}{\mathcal{E}}
\newcommand{\calG}{\mathcal{G}}

\newcommand{\given}{\:\vert\:}
\newcommand{\relativeto}{\:\|\:}
\newcommand{\D}[1]{D_{#1}}

\newcommand{\subPi}[2]{\Pi_{({#1},{#2})} }

\newcommand{\susers}[1]{#1_\text{users}}
\newcommand{\sitems}[1]{#1_\text{items}}
\newcommand{\stypes}[1]{#1_\text{types}}

\newcommand{\srec}{E_\text{recommended}}
\newcommand{\schosen}{E_\text{chosen}}
\newcommand{\styped}{E_\text{types}}

\newcommand{\numusers}[1]{#1_\text{users}}
\newcommand{\numsongs}[1]{#1_\text{songs}}
\newcommand{\numtags}[1]{#1_\text{tags}}
\newcommand{\numconsumed}{E_\text{consumed}}
\newcommand{\numtagged}{E_\text{tagged}}

\newcommand{\U}{V_{{U}}}
\newcommand{\IOne}{V_{{I_1}}}
\newcommand{\ITwo}{V_{{I_2}}}
\newcommand{\TOne}{V_{{T_1}}}
\newcommand{\TTwo}{V_{{T_2}}}
\newcommand{\GOne}{V_{{G_1}}}
\newcommand{\GTwo}{V_{{G_2}}}

\newcommand{\XU}{X_{{U}}}
\newcommand{\XIOne}{X_{{I_1}}}
\newcommand{\XITwo}{X_{{I_2}}}
\newcommand{\XTOne}{X_{{T_1}}}
\newcommand{\XTTwo}{X_{{T_2}}}
\newcommand{\XGOne}{X_{{G_1}}}

\newcommand{\EU}{E_{U}}
\newcommand{\GOneU}{E_{G_1}}
\newcommand{\GTwoU}{E_{G_2}}

\newcommand{\UlikeIOne}{E_\text{like}}
\newcommand{\UseenIOne}{E_\text{seen}}
\newcommand{\UrecIOne}{E_{\text{rec},1}}

\newcommand{\UrateITwo}{E_\text{rate}}
\newcommand{\UrecITwo}{E_{\text{rec},2}}

\newcommand{\IOneTOne}{E_{\tau_{11}}}
\newcommand{\IOneTTwo}{E_{\tau_{12}}}
\newcommand{\ITwoTTwo}{E_{\tau_{22}}}

\newcommand{\users}{V_\text{users}}
\newcommand{\affiliations}{V_\text{affiliations}}
\newcommand{\posts}{V_\text{posts}}
\newcommand{\medias}{V_\text{medias}}
\newcommand{\tags}{V_\text{tags}}
\newcommand{\topics}{V_\text{topics}}
\newcommand{\articles}{V_\text{articles}}

\newcommand{\Xusers}{X_\text{users}}
\newcommand{\Xaffiliations}{X_\text{affiliations}}
\newcommand{\Xposts}{X_\text{posts}}
\newcommand{\Xmedias}{X_\text{medias}}

\newcommand{\Xtopics}{X_\text{topics}}
\newcommand{\Xarticles}{X_\text{articles}}

\newcommand{\Ufollow}{E_\text{follow}}
\newcommand{\Uaff}{E_\text{identify}}
\newcommand{\Maff}{E_\text{affiliate}}
\newcommand{\Upub}{E_\text{post}}
\newcommand{\Ushares}{E_\text{share}}
\newcommand{\Pmentions}{E_\text{mention}}
\newcommand{\Part}{E_\text{link}}
\newcommand{\Amed}{E_\text{belong}}
\newcommand{\Ptag}{E_\text{use}}
\newcommand{\Atop}{E_\text{treat}}
\newcommand{\Ptop}{E_\text{include}}
\newcommand{\Ttop}{E_\text{refer}}

\newcommand{\habitats}[1]{#1_\text{habitats}}
\newcommand{\species}[1]{#1_\text{species}}
\newcommand{\genera}[1]{#1_\text{genera}}
\newcommand{\families}[1]{#1_\text{families}}
\newcommand{\individuals}[1]{#1_\text{individuals}}

\newcommand{\connect}{E_\text{connect}}
\newcommand{\belongS}{E_\text{belong,1}}
\newcommand{\inhabit}{E_\text{inhabit}}
\newcommand{\eat}{E_\text{eat}}
\newcommand{\parasite}{E_\text{parasite}}
\newcommand{\belongG}{E_\text{belong,2}}
\newcommand{\belongF}{E_\text{belong,3}}
\newcommand{\belongBeyond}{E_\text{belong,4}}

\newcommand{\units}[1]{{#1}_\text{units}}
\newcommand{\firms}[1]{{#1}_\text{firms}}
\newcommand{\persons}[1]{{#1}_\text{persons}}

\newcommand{\produced}{E_\text{produced}}
\newcommand{\control}{E_\text{control}}
\newcommand{\own}{E_\text{own}}
\newcommand{\crossown}{E_\text{cross-own}}
\newcommand{\crosscontrol}{E_\text{cross-control}}

\newcommand{\authors}[1]{{#1}_\text{authors}}

\newcommand{\arts}[1]{{#1}_\text{papers}}
\newcommand{\journals}[1]{{#1}_\text{journals}}
\newcommand{\domains}[1]{{#1}_\text{domains}}
\newcommand{\keywords}[1]{{#1}_\text{keywords}}
\newcommand{\institutions}[1]{{#1}_\text{lab.}}

\newcommand{\papercite}{E_\text{cite}}
\newcommand{\publish}{E_\text{publish}}

\newcommand{\paperwrite}{E_\text{write}}
\newcommand{\aff}{E_\text{aff.}}
\newcommand{\edit}{E_\text{edit}}
\newcommand{\usekey}{E_\text{use}}
\newcommand{\belongdomain}{E_\text{belong}}
\newcommand{\treatdomain}{E_\text{treat}}

\newcommand{\eg}{{\em e.g.}\xspace}
\newcommand{\ie}{{\em i.e.}\xspace}

\tikzset{
  > = {Stealth [inset = 0pt, length = 5pt, angle' = 30, round]},
  vertexset/.style = {minimum width = 0.75cm, minimum height = 0.75cm, inner sep = 0pt},
  vertex/.style = {minimum width = 0.75cm, minimum height = 0.75cm, inner sep = 0pt, draw, circle},
  marked/.style = {draw = blue, color = blue, line width = (4*#1), text = blue},
}

\newcommand{\drawLink}[4][]{
  \ifthenelse{\equal{#4}{1}}{\drawSimpleLink[#1]{#2}{#3}}
  {\ifthenelse{\equal{#4}{2}}{\drawDoubleLink[#1]{#2}{#3}}
    {\ifthenelse{\equal{#4}{3}}{\drawTripleLink[#1]{#2}{#3}}}
  }
  
}

\newcommand{\drawSimpleLink}[3][]{
  \draw [->,#1] (#2) -- (#3);
}

\newcommand{\drawDoubleLink}[3][]{
  \draw [->,#1] (#2) to [bend left=7.5] (#3);
  \draw [->,#1] (#2) to [bend right=7.5] (#3);
}

\newcommand{\drawTripleLink}[3][]{
  \draw [->,#1] (#2) to [bend left=15] (#3);
  \draw [->,#1] (#2) -- (#3);
  \draw [->,#1] (#2) to [bend right=15] (#3);
}

\begin{document}

%\maketitle

\begin{frontmatter}
\journal{Theoretical Computer Science}
\title{Measuring Diversity in Heterogeneous Information Networks}
% Authors

\author{ Pedro Ramaciotti Morales\footnote{Corresponding author: pedro.ramaciottimorales{@}sciencespo.fr}}
\address{Sciences Po, m\'edialab, Paris, France \& Sorbonne Universit\'e, CNRS, LIP6, F-75005 Paris, France}

\author{Robin Lamarche-Perrin}
\address{CNRS, ISC-PIF, France}

\author{ Rapha\"el Fournier-S'niehotta}
\address{CEDRIC, CNAM, Paris, France}

\author{ R\'emy Poulain}
\author{ Lionel Tabourier}
\address{Sorbonne Universit\'e, CNRS, LIP6, F-75005 Paris, France}

\author{ Fabien Tarissan}
\address{Universit\'e Paris-Saclay, CNRS, ISP, ENS Paris-Saclay, Cachan, France}

\begin{abstract}
Diversity is a concept relevant to numerous domains of research varying from ecology, to information theory, and to economics, to cite a few. 
It is a notion that is steadily gaining attention in the information retrieval, network analysis, and artificial neural networks communities.
While the use of diversity measures in network-structured data counts a growing number of applications, no clear and comprehensive description is available for the different ways in which diversities can be measured. % in data described by these structures.
In this article, we develop a formal framework for the application of a large family of diversity measures to heterogeneous information networks (HINs), a flexible, widely-used network data formalism.
This extends the application of diversity measures, from systems of classifications and apportionments, to more complex relations that can be better modeled by networks.
In doing so, we not only provide an effective organization of multiple practices from different domains, but also unearth new observables in systems modeled by heterogeneous information networks.
We illustrate the pertinence of our approach by developing different applications related to various domains concerned by both diversity and networks.
In particular, we illustrate the usefulness of these new proposed observables in the domains of recommender systems and social media studies, among other fields.
\end{abstract}


\end{frontmatter}

\clearpage

\setcounter{tocdepth}{2}
\tableofcontents

\clearpage

\section{Introduction}
\label{sec:part1}

Diversity is a concept of importance in several different domains of research, such as ecology \cite{mccann2000diversity}, economy \cite{geroski1992choice}, public policy \cite{geroski1992choice}, information theory \cite{aczel1975measures,renyi1961measures}, social media studies \cite{nikolov2015measuring,kulshrestha2015characterizing}, and complex systems \cite{ursem2002diversity,riget2002diversity}, among many others.
Across the full range of domains where it is used, diversity refers to some combination of three properties of systems including classifications of items, identified as \emph{variety} (the number of types of entities in the system), \emph{balance} (the distribution of entities into types), and \emph{disparity} (how different types of entities are between them) \cite{stirling2007general}.
Diversity measures are quantitative indices for these properties.
Prominent examples are Shannon's entropy in information theory \cite{shannon1948mathematical}, the Gini Index in economy \cite{gini1921measurement}, and the Herfindahl-Hirschmann Index \cite{rhoades1993herfindahl} in competition law.
Examples of the application of these indices can be found in the measurement of biodiversity in ecology \cite{hill1973diversity}, industrial concentration in economics~\cite{encaoua1980degree,chakravarty1991axiomatic}, and online social phenomena such as \textit{filter bubbles} and \textit{echo chambers} \cite{nikolov2015measuring}.
The notion of diversity has recently become central as well in the context of digital platforms and online media.  
The fact that digital platforms increasingly resort to algorithmic recommendations to drive the choices of users has led the scientific community to analyze the impact of recommendations made to users. 
Although one can argue that this recent development provides users with useful information, the phenomenon also feeds into fears of unpredictable outcomes over the long term, the most debated being the emergence of so-called filter bubbles \cite{blackboxsoc, pariser,
2018discrimination}.
In this context, while the need to measure and audit recommendation
systems is commonly agreed upon~\cite{detective, ziegler2005improving}, there is no
consensus on how to properly measure the impact of recommendations
on users. 
On the other hand, many studies have highlighted the need to explore diversity or serendipity (the fortunate discovery of unexpected items) in the way information is exposed to users~\cite{div2, recdiv, divmusic2}.

Diversity measures can be computed over different types of data in a multitude of contexts.
Access to data traces of different real phenomena has enabled for a tremendous extension of the reach of quantitative studies in many disciplines.
One particular type of data over which diversity measures can be computed is network-structured data, best represented using graph formalisms. 
Recently, formalisms such as \emph{heterogeneous information networks} (HINs) \cite{sun2013mining,sun2009ranking}
have been successfully used to provide ontologies for unstructured data, especially in the contexts of information retrieval~\cite{sun2009ranking} and recommender systems~\cite{yu2014personalized}, as well as in the artificial intelligence and representation learning communities \cite{fu2017hin2vec,sydow2013notion,chen2017hine}.

Much of the success of these representations and their precursors -- such as \emph{multi-layer graphs} \cite{de2013mathematical,de2015ranking}~-- is due to the way in which semantic relations can be mapped to sets of paths between groups of entities.
These sets of paths are called \emph{meta paths} and can be easily exploited by algorithms.
One prominent way of exploiting meta paths is by constraining random walks to them (\ie, constraining random walks to paths contained in a given meta path). 
This procedure has been extensively used in the computation of similarity \cite{sun2011pathsim,lao2010relational,xiong2014top} or for recommendation purposes~\cite{yu2014personalized,yu2013recommendation,shi2018heterogeneous}.
While the application of diversity measurements to graph structures is not new \cite{yu2006soft,li2012scalable}, it is gaining widespread use in different communities \cite{hristova2016measuring}, and in particular in the information retrieval and recommender systems communities \cite{king2014heterogeneous}.
Few studies have hinted at the application of \textit{entropy} \cite{shannon1948mathematical} (one prominent diversity measure) to distributions computable from meta path structures in heterogeneous information networks.
This application of entropy has been done to provide diverse recommendations \cite{nandanwar2018fusing}. 
In similarity searches (the search for similar items in information retrieval), entropy has also been used to measure information gain in the selection of different meta paths \cite{vahedian2016meta,yu2012user}.
However, no clear and comprehensive description is currently available for the different ways in which diversity measures can be computed from data described with network-structured data.
Several communities interested in both network representation models and diversity measures have limited -- or no -- examples of application at their disposal, let alone any theory or a framework on which to develop applications.

In this article we develop \emph{network diversity measures}: a comprehensive theory of diversity and a formal framework for its application to network-structured data.
This framework relies on modeling data with heterogeneous information networks using multigraphs for generality.
Doing so, we collect and unify a wide range of results on quantitative diversity measures across different disciplines covering most practical uses.
And in developing this formal framework, we also provide a unified reformulation of several practices existing in scientific literature.
In addition, we point to new information that may be extracted by measuring the diversity of previously unconsidered observables in network-structured data.
One of the main applications of \emph{network diversity measures} is the extension of existing diversity measures, from relatively simple systems of classification and apportionment (\eg, species in ecosystems, units produced by firms) to more complex data, best modeled by network structures.
The relevance and usefulness of these new \emph{network diversity measurements} are illustrated by the development of practical examples in different domains of research, including recommender systems, social media and platforms, and ecology, among others.

The main contributions of this article are:
\begin{itemize}
\item a new organization of an axiomatic theory of diversity measures encompassing most uses across several disciplines;
\item a formalization of concepts emerging in graph theory (especially in applications in recommender systems, information retrieval, and representation learning communities), in particular that of \metapath{}s and observables computable from meta paths;
\item the proposal of several \emph{network diversity measures}, resulting from applying diversity measures to distribution probabilities computable in the heterogeneous information network formalism;
\item the application of these network diversity measures to previously existing quantitative observables in different research domains and the development of new applications through examples.
\end{itemize}

In Section~\ref{sec:part2}, we provide a framework to organize diversity measures found in the literature.
This new framework has the advantage of covering a large part of existing concepts relating to diversity, and of formalizing the algebraic properties that they obey.
Then, in Section~\ref{sec:part3}, we define random walks in the context of heterogeneous information networks.
In particular, we formalize the concept of meta path.
Constrained random walks along particular meta paths will play a central role in the rest of the article when computing diversity in systems represented by networks.
Indeed, in Section~\ref{sec:part4}, we combine diversity measures described within the framework with different observables computed from constrained random walks in order to derive families of interpretable network diversity measures.
Finally, in Section~\ref{sec:part5} we illustrate the relevance of these measures using them in applications in various fields concerned by the concept of diversity.

\section{The Concept of Diversity}
\label{sec:part2}

In general, diversity refers to certain properties of a system that contains items that are classified into types.
These properties are related to the number of types used, the way in which items are classified into types, and how different types are from one another.
This simple model of items classified into types accounts for the usage of diversity in many domains of research. Prominent examples are units of wealth or revenue classified as belonging to different persons (in economics), the number of individuals classified into different species (in ecology), or produced units of a commodity classified by firms (in competition law).

%%%%%%%%%%%%%%%%%%%%%%%%%%%%%%%%%%%%%%%%%%%%%%%%%%%%%%%%%%
\subsection{Items, types, and classifications}
\label{subsec:items_types_classifications}

Let us consider a system made of a set $\itemset$ of items, a set $\typeset$ of types, and a membership relation $\belongs\subseteq\itemset\times\typeset$ indicating the way items are classified according to types: item $i\in\itemset$ is classified as being of type $t\in\typeset$ if and only if $(i,t)\in\belongs$.
The use membership relations allows for an item to have more than one type.
The diversity measures considered in this article are functions $D:\itemset\times\typeset\rightarrow\reals^+$ that map any such system to a diversity value $d$, \ie, $D: \belongs \mapsto d\in\reals^{+}$.

We do not consider the problem of identification, \ie, what should be considered as an item in a universe of possible elements and what types should be considered in a classification.
This identification problem is an important question, however it deals with the meaning of the system's elements and its semantic content, which is beyond the scope of this work.

We define the {\em abundance} of type $t\in\typeset$ as the number of items of that type: $a_\belongs(t) = |\{i \in \itemset : (i,t)\in\belongs \}|$, and the {\em proportional abundance} as $p_\belongs(t) = \frac{a_\belongs(t)}{|\belongs|}$.
Using these definitions, we further narrow our consideration of diversity measures to functions that map proportional abundances resulting from a classification to non-negative real values: $D(\tau) = D(p_\tau(t_1),\ldots,p_\tau(t_k))$ with $k=|\typeset|$ being the number of types.
Hence, a diversity measure $D$ is an application from $\Delta^*$ to $\reals^{+}$, where $\Delta^* = \cup_{k \geq 0} \Delta^k$ is the union of all standard $k$-simplices, that is the set of probability distributions on discrete spaces of size $k+1$:
$$\Delta^k = \left\{ (p_1, \ldots, p_{k+1}) \in \mathbb{R}^{k+1} : \forall i \leq k, \; p_i \in [0,1] \; \text{and} \; \textstyle\sum_{i \leq k+1} {p_i} = 1 \right\} \text{.}$$

%%%%%%%%%%%%%%%%%%%%%%%%%%%%%%%%%%%%%%%%%%%%%%%%%%%%%%%%%%
\subsection{The diversity of diversity measures}
\label{subsec:diversity_of_diversities}

As stated in the previous subsection, the term \textit{diversity} is used to designate various properties of dissimilarity in a range of domains, such as ecology \cite{helfman2009diversity,mcnaughton1977diversity,may1975patterns,mccann2000diversity},
life sciences \cite{smith1989trees}, economics \cite{geroski1992choice,rhoades1993herfindahl}, public policy \cite{gillett2003praise,silverberg94innovation,nowotny2001re}, information theory \cite{aczel1975measures,renyi1961measures}, internet \& media studies \cite{nikolov2015measuring,kulshrestha2015characterizing}, physics \cite{shevchenko2006structural,schneider1994life}, social sciences \cite{grabher1997organizing}, complexity sciences \cite{ashby1991requisite,dyson1962statistical,ursem2002diversity,riget2002diversity}, and opinion dynamics \cite{yang2009effects}. 
This term refers to different properties of systems of items classified into types. 
Accordingly, diversity measures are functions assigning to each system a diversity value, intended to be a quantitative measurement of these different properties.

The properties referred to by the term \emph{diversity} across the full range of sciences are some combination of three properties, identified as \emph{variety}, \emph{balance}, and \emph{disparity} \cite{stirling2007general}: 

\begin{itemize}
  \item \emph{variety} is the number of types into which items of a system can be classified;
\item \emph{balance} is a measure of the extent to which the pattern of proportional abundances resulting from a classification of items into types is evenly distributed (\ie, balanced);
  \item and \emph{disparity} is the degree to which types can be differentiated according to a metric on the set of types $\typeset$.
\end{itemize}

The reader is referred to \cite{stirling1998economics} for an extended discussion of these properties.

We illustrate the concept of \emph{diversity} through classic examples of diversity measures present in works from different fields. 
For this purpose, we consider a proportional abundance distribution $p=\left(p_1,p_2,\dots,p_{|\typeset|} \right)$ resulting from the classification of items $I$ into types $T$.

\textbf{Richness} \cite{macarthur1965patterns,gotelli2011estimating} is a common diversity measure only related to the property of \emph{variety}. 
Often used in ecology, it simply measures the number of types \emph{effectively} used to classify items. If a bookcase contains novels, comics, and travel books, its richness is equal to 3, regardless of proportions.

  $$\richness(p) = |\left\{i \in \{1,2,...,|T|\} : p_i > 0\right\}|.$$

Richness only counts types that are effectively used in a classification. If one considers a typology of 20 possible types to examine two bookcases, the first containing books of 3 different types and the second book of 4 different types, the second one will be more diverse under this measure. 
The property captured by this measure coincides with the property identified as \emph{variety}. 
Richness may serve as a basis for other measures, such as, the ratio between richness and the number of classified items \cite[Section 9]{odum1959fundamentals}.

\textbf{Shannon entropy} \cite{shannon1948mathematical,shannon1963mathematical}, here denoted by $H$ and related to the \textit{variety} and \emph{balance} properties, is a cross-disciplinary diversity measure, most often used in the field of information theory. It quantifies the uncertainty in predicting the type of an item taken at random. If one knows the proportional abundance of types of books in a bookcase, and if one draws books from it at random, Shannon entropy is the average number of binary type-checks (\ie, ``yes or no'' questions about the book belonging to a given type) one would have to make per book in order to determine its type: 

  $$\entropy(p) = - \sum\limits^{|T|}_{i=1} {p_i \log_2 p_i}.$$

Many classical diversity measures are functions of the properties identified as \emph{variety} and \emph{balance}. Shannon entropy, introduced in the context of channel capacity in telecommunications, is clearly affected by proportional abundances, and thus by their \emph{balance}, but also by their \emph{variety}: according to Shannon entropy, a bookcase with books that are uniformly distributed among 5 types is more diverse  than a bookcase with books that are uniformly distributed among 4 types. By applying normalization, one may restrain the measurement to the property of \emph{balance}. The diversity measure known as \textbf{Shannon Evenness} \cite{pielou1969introduction} in ecology, for example, consists of the ratio between measured entropy and maximal entropy for the same number of effective types and only accounts for the property of \emph{balance}.

Shannon entropy has found renewed use by the information retrieval and artificial intelligence communities.
In information retrieval, some recommender systems exploit Shannon entropy to improve performance of algorithmic recommendations \cite{wulfmeier2015maximum}.
In deep learning methods for artificial intelligence, Shannon entropy is often used for quantifying information gain \cite{wang2015collaborative}. 

The \textbf{Herfindahl-Hirschman Index} \cite{rhoades1993herfindahl}, here denoted as $\HHI$, is mainly used in competition law or antitrust regulation in economy. 
It is intended to measure the degree of concentration of items into types.
If one takes 2 books from a bookcase at random, Herfindahl-Hirschman Index is the probability of them belonging to the same type:
  $$\HHI(p) = \sum\limits^{|T|}_{i=1} {p_i^2}.$$

\noindent{}Related to the \emph{variety} and \emph{balance} properties, this index (also known as the \textbf{Simpson Index} \cite{simpson1949measurement}), was first introduced by Hirschman \cite{hirschman1945national} and later by Herfindahl \cite{herfindahl1950concentration} in the study of the concentration of industrial production.
Concentration and diversity are opposite and complementary concepts.
Higher diversity means lower concentration and \emph{vice versa}.

A related diversity measure, the \textbf{Gini-Simpson Index} \cite{gini1921measurement} (also called the \textbf{Gibbs-Martin Index} in sociology and psychology \cite{gibbs1962urbanization} and \textbf{Population Heterozygosity} in genetics \cite{nei1978estimation}) is another prominent example of a measure accounting for \textit{variety} and \emph{balance}. Also known as the probability of interspecific encounters in ecology \cite{hurlbert1971nonconcept}, it is the probability of the complementary event associated with the Herfindahl-Hirschman Index, \ie, the probability of randomly selecting two items with different types. 

This is not to be confused with the \textbf{Gini Coefficient} \cite{sen1997economic}, commonly used in economics, which is a \emph{balance}-only diversity measure that may be interpreted as a measure of inequality where items are units of wealth distributed into types. One of the formulations of the Gini Coefficient is given by
$$\gini(p)=\frac{1}{2|T|}\sum\limits^{|T|}_{i=1}\sum\limits^{|T|}_{j=1}|p_i-p_j|.$$

Other diversity measures address only the property of \emph{balance}. The \textbf{Berger-Parker Index} \cite{berger1970diversity}, here denoted as $\BPI$, is another prominent example. Also common in ecology, it measures the proportional abundance of the most abundant type. If 90\% of the books in a bookcase are comics, its Berger-Parker Index will be 0.9, regardless of how the remaining 10\% of books are classified:
  $$\BPI(p) = \max_{i \in \{1,2,...,|T|\}} {p_i}.$$
  It is easy to see that only the \emph{balance} property affects this diversity measure. 
 If the books of a first bookcase are classified as 90-10\% into two types and those in a second bookcase as 90-5-5\% into three types, both bookcases still have the same diversity according to this measure.

Another group of existing diversity measures addresses the \emph{disparity} property. In its most general form, a pure-\emph{disparity} diversity measure is a function of the pairwise distance between types of $T$ in some disparity space \cite{weitzman1992diversity}. One example of a measure of \emph{disparity} is proposed in \cite{solow1994measuring}:
$$
\text{Disparity}(T) = \frac{1}{|T|\,(|T|-1)}\sum\limits_{t,t'\in T} d(t,t'),
$$
where $d$ is a metric on the set $T$ of types.
\emph{Disparity} is the underlying property in some use cases of the notion of diversity. Examples may be found in fields such as paleontology \cite{williams1991measuring}, economics \cite{nguyen2005variety}, and biology \cite{runnegar1987rates}. Furthermore, diversity measures accounting for \emph{disparity} as well as \emph{variety} and \textit{balance} exist \cite{junge1994diversity}.

While the measurement of \emph{disparity} relies on the existence of topological or metrical structures for the set of types $\typeset$, that of \emph{variety} and \emph{balance} relies solely on the establishment of identification and classification in a system of items and types, which is the setting of many studies and applications.
As indicated in the previous subsection, we focus in this article on diversity measures for this latter setting, thus setting aside \emph{disparity}-related diversity measures.

%%%%%%%%%%%%%%%%%%%%%%%%%%%%%%%%%%%%%%%%%%%%%%%%%%%%%%%%%%
\subsection{A theory of diversity measures}
\label{subsec:theory_of_diversity}

In Section~\ref{subsec:items_types_classifications}, we first limited the scope of diversity measures to that of functions mapping systems with given items, types, and classification, to non-negative real numbers.
Then we further limited the scope to only functions mapping probability distributions to non-negative real numbers.
In this section, we further reduce the scope of diversity measures by prescribing axioms reflecting the desired properties such measures should have.
%
% By imposing axioms it will be shown that new desired properties emerge as a consequence.

In the domain of information theory, there are several possible axiomatic theories that give rise to entropies and diversity measures (cf.~\cite{rao2014rao,csiszar2008axiomatic,wang2005axiomatic,aczel1987synthesizing}). 
Drawing from these existing axiomatizations, we propose an organization of axioms suited for the purposes of this article.

We first introduce four axioms that encode properties which are necessary for a diversity measure, \ie, \emph{symmetry}, \emph{expansibility}, \emph{transferability}, and \emph{normalization}.
Then, we present a family of functions that satisfy these properties.
By imposing an additional property known as \emph{replicability} in the form of an axiom, the resulting measures of the theory correspond to the family of functions known as \emph{true diversities}.
One member of this family, closely related to Shannon entropy, has additional properties of interest for the measurement of diversity in networks.

%%%%%%%%%%%%%%%%%%%%%%%%%%%%%%%%%%%%%%%%%%%%%%%
\subsubsection{Properties of diversity measures}

Let us consider a diversity measure $D:\Delta^{k-1} \rightarrow \reals^{+}$, a probability distribution $p=(p_1,...,p_k)\in\Delta^{k-1}$, and some properties of interest in the form of axioms for a theory of diversity.

A first property, called {\it symmetry} (or {\it anonymity}), is said to be satisfied by a diversity measure if it is invariable to permutation of types.
For instance, a bookcase with 25\% comics and 75\% novels has the same diversity as a bookcase with 75\% comics and 25\% novels using a symmetric diversity measure. 
This means that symmetric diversity measures are blind to the nature of types.

\begin{axiom}[Symmetry]
\label{ax:symmetry}
For any permutation $\sigma$ on the set $\{1,2,...,k\}$, a diversity measure $D$ is symmetric if and only if
$$
D(p_1,p_2,...,p_k)=D(p_{\sigma(1)},p_{\sigma(2)},...,p_{\sigma(k)}).
$$
\end{axiom}

We also require that diversity measures be {\it expansible}, or {\it invariant to non-effective types}, that is, invariant to the addition of types with no items.
Adding a type with no items does not impact diversity: considering the type ``dictionaries'' which does not contain any books does not change the diversity of a bookcase.

\begin{axiom}[Expansibility]
\label{ax:expansibility}
A diversity measure $D$ is expansible if and only if
$$
D(\underbrace{p_1,p_2,...,p_k}_{k\text{ entries}})=D(\underbrace{p_1,p_2,...,p_k,0}_{k+1\text{ entries}}).
$$
\end{axiom}

For a diversity measure to be a measure of balance it needs to satisfy the {\it transfer principle}, also called the {\it Pigou-Dalton principle}~\cite{dalton1920measurement}: if a bookcase has more novels than comics, replacing some novels with new comics should increase its diversity (if the new number of comics does not surpass the new number of novels).

\begin{axiom}[Transfer Principle]
\label{ax:transfer}
A diversity measure $D$ satisfies the transfer principle if and only if, for all $i,j$ in $\{1,...,k\}$, if $p_i>p_j$, then 
$$
 \forall \epsilon \leq \frac{p_i - p_j}{2}, \quad 
 D(\underbrace{\ldots,p_i-\epsilon, \ldots, p_j+\epsilon, \ldots}_{k\text{ entries}}) 
 \geq 
 D(\underbrace{\ldots, p_i, \ldots, p_j, \ldots}_{k\text{ entries}}).
$$
\end{axiom}

It is easy to verify that axioms \ref{ax:symmetry}, \ref{ax:expansibility}, and \ref{ax:transfer} imply the following {\it merging} property.

\begin{theorem}[Merging]
\label{thm:merging}
A diversity measure $D$ that satisfies axioms \ref{ax:symmetry}, \ref{ax:expansibility} \& \ref{ax:transfer} is such that
$$
D(\underbrace{\ldots,p_i,p_{i+1},\ldots}_{k\text{ entries}}) \quad \geq \quad D(\underbrace{\ldots,p_i+p_{i+1},\ldots}_{k-1 \text{ entries}}).
$$
\end{theorem}

\begin{pf}
By the application of Axiom~\ref{ax:expansibility} and Axiom~\ref{ax:symmetry}, the claim of the theorem is equivalent to
$$D(\underbrace{\ldots,p_i,p_{i+1},\ldots}_{k\text{ entries}}) \quad \geq \quad D(\underbrace{\ldots,p_i+p_{i+1},0,\ldots}_{k \text{ entries}}).$$
Without loss of generality, let us suppose that $p_i\geq p_{i+1}$, and let us apply the transfer principle of Axiom~\ref{ax:transfer} to this distribution $(\ldots,p_i+p_{i+1},0,\ldots)$ of $k$ entries.
The first condition for its application is always satisfied, \ie, $p_i+p_{i+1}>0$ (if $p_i+p_{i+1}=0$ the theorem is trivially assured by Axioms~\ref{ax:expansibility} \&{} \ref{ax:symmetry}).
Choosing $\epsilon=p_{i+1}$ satisfies the second condition of application, because $p_{i+1}\leq(p_i+p_{i+1}-0)/2$ if $p_i \geq p_{i+1}$. Finally, the application of Axiom~\ref{ax:transfer} gives the desired result.\eop
\end{pf}

These first three axioms also imply that diversity measures of the theory are bounded.

\begin{theorem}[Bounds for diversities measures]
\label{thm:bounds_for_measures}
A diversity measure $D$ that satisfies axioms \ref{ax:symmetry}, \ref{ax:expansibility} \& \ref{ax:transfer} is such that
$$
D(\underbrace{1/k,1/k,\ldots,1/k}_{k\text{ entries}}) \quad \geq \quad D(p_1,p_2,\ldots,p_k) \quad \geq \quad D(\underbrace{1,0,\ldots}_{k\text{ entries}}).
$$
\end{theorem}

\begin{pf}
The second inequality is warranted by Theorem~\ref{thm:merging}.
If distribution $p=(p_1,\ldots,p_k)\in\Delta^{k-1}$ is the uniform distribution, that we shall denote by $u$, the first inequality is trivialy satisfied. 
If, on the other hand, $p$ is any distribution that is not uniform, we will show that Axiom~\ref{ax:transfer} assures the construction of a sequence of $m$ distributions $p^1,\ldots,p^m$ in $\Delta^{k-1}$, such that $p^0 = p$, $p^m$ is the uniform distribution, and $D(p^1)\leq D(p^2) \leq \ldots \leq D(p^m)$, thus assuring that $D(u)\geq D(p)$.
We do this by adapting the proof of \cite[Thm. 1]{encaoua1980degree} developed for measures of concentration.
Because of Axiom~\ref{ax:symmetry}, we can set, without loss of generality, $p^1$ as the distribution that results from ordering the entries of $p$ in decreasing order, still resulting in $D(p^1)=D(p)$.
Given a non-uniform distribution $p^l$ of the sequence, and assuming that its entries are arranged in decreasing order, we will show how to compute the next distribution of the sequence, $p^{l+1}$, so that $D\left(p^{l+1}\right)\geq D\left(p^l\right)$, using Axiom~\ref{ax:transfer}.
Let $\delta^l$ be a vector in $\reals^k$ resulting from subtracting $p^l$ and $u$ element-wise: $\delta^l_i = p^l_i-u_i=p^l_i-1/k$.
Now let $i^-$ be the first negative entry of $\delta^l$:  $i^- =\text{min}\left\{1\leq i \leq k:\delta^l_i<0\right\}$.
Because entries of $p^l$ cannot all be less than $1/k$, we know that $i^-$ is never the first entry ($i^- > 1$), and because entries cannot all be greater than $1/k$, we know that $i^-$ can always be determined ($1<i^- \leq k$), as long as $p^l$ is non-uniform.
Because $p^l$ is not uniform, we know that $p^l_1>1/k$, and so we transfer the quantity $\text{min}(\delta^l_1,-\delta^l_{i^-})$ from entry $p^l_1$ to entry $p^-$ to compute a distribution $\overline{p}^{l+1}$.
The components of $\overline{p}^{l+1}$ are computed as:  $\overline{p}^{l+1}_1 = p^l_1 - \text{min}(\delta^l_1,-\delta^l_{i^-})$, $\overline{p}^{l+1}_{i^-} = p^l_{i^-} + \text{min}(\delta^l_1,-\delta^l_{i^-})$, and $\overline{p}^{l+1}_i = p^l_i$ for $i\notin\{1,i^-\}$.
Next, we compute $p^{l+1}$ as the distribution resulting from arranging the elements of $\overline{p}^{l+1}$ in decreasing order.
Because either entry $i=1$ or entry $i=i^-$ was set to $1/k$, a new entry is now $1/k$ (as entries in $u$). 
And because, at each step in the sequence, a new entry is set to $1/k$, we know that this sequence is finite.\eop
\end{pf}

In order for diversity measures to have a scale for measurement, we need to impose values of minimal and maximal diversity \cite{chakravarty1991axiomatic}. We establish this as a property, called the {\it normalization} principle. Normalization means that if all types of books are equally abundant in a bookcase, its diversity is equal to the number of effective types.

\begin{axiom}[Normalization]
\label{ax:normalization}
A diversity measure $D$ satisfies the normalization principle if and only if
$$
D(\underbrace{1/k,...,1/k}_{k\text{ entries}})=k.
$$
\end{axiom}

It is easy to see that values of diversity measures of the theory are bounded as a consequence of the normalization axiom.

\begin{theorem}[Bounds for diversity values]
\label{thm:bounds_for_diversities}
A diversity measure $D$ that satisfies axioms \ref{ax:symmetry}, \ref{ax:expansibility}, \ref{ax:transfer} \& \ref{ax:normalization} is such that, for all $p\in\Delta^{k-1}$, we have $k \geq D(p) \geq 1$.
\end{theorem}

%%%%%%%%%%%%%%%%%%%%%%%%%%%%%%%%%%%%%%%%%%%%%%%
\subsubsection{Self-weighted quasilinear means}

One of the advantages of restricting the scope of diversity measures to functions of distributions $p\in\Delta^*$, is that they may then be used in conjunction with probability computations, as will be shown in Section~\ref{sec:part4}.
The measures considered thus far also belong to the more general class of \textit{aggregation functions}.
The most general form of aggregation functions that is compatible with the axioms of probability \cite{hoffmann2006concavity} is the family of \textit{quasilinear means} (developpd by Kolmogorov \cite{kolmogorov1930notion} and Nagumo \cite{nagumo1930klasse}).
Quasilinear means of a probability distribution are central to the quantification of information in information theory \cite{aczel1975measures}, and are of the form 
$$
\phi^{-1}\left(\sum\limits^k_{i=1}w_i\phi(p_i) \right),$$
with weights $w_i$ such that $\forall i\in\{1,\ldots,k\}, (0\leq w_i\leq 1)$ with $\sum\limits^k_{i=1}w_i=1$, and for $\phi$ a strictly monotonic increasing continuous function.

A sub-family of quasilinear means, the so-called self-weighted quasilinear means~\cite{pursiainen2008consistency}, has additional properties that will be of interest in what follows.

\begin{definition}[Self-weighted quasilinear means \cite{pursiainen2008consistency}]
\label{def:SWQLM}
A function $S:\Delta^*\rightarrow\reals^+$ is a self-weighted quasilinear mean if it is of the form
$$
S(p)=\phi^{-1}\left(\sum\limits^k_{i=1}p_i\phi(p_i)\right),
$$
\noindent with $\phi$ a strictly monotonic increasing continuous function.
\end{definition}

Further restrictions of the considered diversity measures, described by the following theorem, result in a family of functions that simultaneously satisfy the above properties described by the axioms.

\begin{theorem}[Reciprocal self-weighted quasilinear means are diversity measures of the theory]
\label{thm:SWQLM}

A reciprocal self-weighted quasilinear mean $D=1/S$ such that $h(t)=t\, \phi(t)$ is concave (with function $\phi$ from Definition~\ref{def:SWQLM}), satisfies Axioms \ref{ax:symmetry}, \ref{ax:expansibility}, \ref{ax:transfer} \& \ref{ax:normalization}.
\end{theorem}
\begin{pf}
Let us consider a diversity measure $D$ in the form of the reciprocal of a self-weighted quasilinear mean: $D(p)=\frac{1}{S(p)}$, with $S$ of the form given in Definition~\ref{def:SWQLM}, with $\phi$ continuous strictly increasing such that $h(t)=t\phi(t)$ is concave.
It is easy to check that $D$ satisfies Axiom~\ref{ax:symmetry} (symmetry) because of the commutativity of the sum.
Because the summands are self-weighted, adding new zero-valued entries results in zero-valued summands, which assures that Axiom~\ref{ax:expansibility} (expansibility) is satisfied.
%
%Being strictly increasing, $\phi^{-1}$ is well defined and $\phi^{-1} \!\circ \phi$ yields the identity. 
By construction, uniform distributions of $k$ entries have diversity $\frac{1}{\phi^{-1}\left(k\cdot (1/k)\cdot \phi (1/k) \right) }=k$, assuring that $D$ satisfies Axiom~\ref{ax:normalization} (normalization).

Finally, given $p=(\ldots,p_i,\ldots,p_j,\ldots)$ with $p_i > p_j$ and $\epsilon \leq (p_i-p_j)/2$, let us consider $\tilde{p}=(\ldots,p_i-\epsilon,\ldots,p_j+\epsilon,\ldots)$. 
If $S(\tilde{p})\leq S(p)$, then $D(\tilde{p})\geq D(p)$ and $D$ would satisfy Axiom~\ref{ax:transfer} (transfer principle).
Because $\phi$ is monotonic strictly increasing, $\phi^{-1}$ also is, and $S(\tilde{p})\leq S(p)$ if
$ \sum\nolimits^{k}_{l=1} h(p_l) \geq \sum\nolimits^{k}_{l=1} h(\tilde{p}_l) $. Because $p$ and $\tilde{p}$ share all but the $i$-th and $j$-th entries, this last inequality is assured by
$$
h(p_i)- h(p_i-\epsilon) + h(p_j) -h(p_j+\epsilon) \geq 0.
$$
To see that this inequality holds, let us note that we can always compute $\theta=(p_i-p_j-\epsilon)/(p_i-p_j-2\epsilon)$, with $\theta\in (0,1)$, so that $p_i = \theta (p_i-\epsilon) + (1-\theta)(p_j+\epsilon)$ and $p_j = (1-\theta) (p_i-\epsilon) + \theta(p_j+\epsilon)$. This step is adapted from the proof of \cite[Thm. 3]{encaoua1980degree}.
$\theta$ is always positive by the restrictions on $\epsilon$ required by Axiom~\ref{ax:transfer}.
By concavity of $h$ we can write inequalities for $h(p_i)$ and $h(p_j)$:
\begin{equation*}
\begin{split}
h(p_i) = h\left( \theta(p_i-\epsilon) + (1-\theta)(p_j+\epsilon) \right) \geq \theta h(p_i-\epsilon) + (1-\theta) h(p_j+\epsilon),\\
h(p_j) = h\left( (1-\theta)(p_i-\epsilon) + \theta(p_j+\epsilon) \right) \geq (1-\theta) h(p_i-\epsilon) + \theta h(p_j+\epsilon).
\end{split}
\end{equation*}
Additioning these two inequalities we obtain $h(p_i)- h(p_i-\epsilon) + h(p_j) -h(p_j+\epsilon) \geq 0$.\eop
\end{pf}
  
Theorem~\ref{thm:SWQLM} provides us with an explicit expression for functions that satisfy axioms \ref{ax:symmetry}, \ref{ax:expansibility}, \ref{ax:transfer} \& \ref{ax:normalization}.
The use of self-weighted quasilinear means yields, however, a subset of the functions defined by these axioms. 
Indeed, there are diversity measures that satisfy these axioms but cannot be expressed as self-weighted quasilinear means (\eg, Hall-Tideman Index \cite{hall1967measures}).

%%%%%%%%%%%%%%%%%%%%%%%%%%%%%%%%%%%%%%%%%%%%%%%%%%%%%%%%%%

\subsubsection{True diversities}

An additional property, the {\it replication principle}, captures a characteristic of some diversity measures according to which, if types are replicated $m$ times, diversity is multiplied by $m$ \cite{chakravarty1991axiomatic}. 
Let us suppose, for example, that a bookcase contains 25\% comics and 75\% novels. 
Let us also suppose that we add new items from a different bookcase, in which 25\% of books are dictionaries and 75\% of books are photo albums. 
The diversity of the new --replicated-- bookcase with four types of books is double that of the original bookcase. 

\begin{axiom}[Replication]
\label{ax:replication}
A diversity measure $D$ satisfies the replication principle if it is such that
$$
D\left(\underbrace{\frac{p_1}{m},\frac{p_2}{m},...,\frac{p_k}{m}}_{\ts{1}{st}},\underbrace{\frac{p_1}{m},\frac{p_2}{m},...,\frac{p_k}{m}}_{\ts{2}{nd}},...,\underbrace{\frac{p_1}{m},\frac{p_2}{m},...,\frac{p_k}{m}}_{\ts{m}{th}}\right)=m \, D(p_1,...,p_k).
$$
\end{axiom}

The addition of the replication principle to the theory of diversity uniquely defines a sub-family within that of reciprocal self-weighted quasilinear means, called \emph{true diversities}.

\begin{definition}[True diversity of order $\alpha$]
\label{def:true_diversities}
The $\alpha$-order true diversity, denoted $\D{\alpha}$, is the application $\D{\alpha}:\Delta^*\rightarrow\reals^+$, such that, given $p=(p_1,\ldots,p_k)\in\Delta^*$ and $\alpha\in\reals^+$,
$$
\D{\alpha} (p) = \left( \sum\limits^k_{i=1} {p_i^\alpha} \right)^{\frac{1}{1-\alpha}} \text{\ if\ } \alpha \neq 1, \quad \text{and }\quad \D{1}(p) = \left( \prod\limits^{k}_{i=1} {p_i^{p_i}} \right)^{-1},\quad\text{ with } {p_i}^{p_i}\coloneqq 1\text{ if } p_i=0.
$$
\end{definition}

True diversities were first introduced as the Hill Number \cite{hill1973diversity} and named \emph{true diversity} in \cite{jost2006entropy}. Variants of true diversities exist in different domains. The Hannah-Kay concentration index of order $\alpha$ \cite{hannah1977concentration} is the reciprocal of $\D{\alpha}$. In information theory, R\'enyi Entropy \cite{renyi1961measures} of order $\alpha$, denoted by $H_\alpha$ , is the natural logarithm of $\D{\alpha}$: $H_\alpha(p)=\text{ln} \D{\alpha}(p)$.

\begin{theorem}[Diversity measures that satisfy the replication principle are true diversities \cite{chakravarty1991axiomatic}]
\label{thm:div_measures_are_true_diversities}
Suppose that a diversity measure $D$ can be represented as a reciprocal self-weighted quasilinear mean as in Theorem~\ref{thm:SWQLM}, then $D$ is a true diversity for some order $\alpha$ if and only if $D$ satisfies the replication principle of Axiom~\ref{ax:replication}.
\end{theorem}

The reader is referred to \cite[Theorem 3.1]{chakravarty1991axiomatic} for the proof.

The replication principle may be needed to avoid otherwise paradoxical results in many applications. 
Let us consider for example a library with 3 bookcases, each containing items of 3 different types: 9 types of items organized in 3 bookcases, with no types dispersed in multiple bookcases.
Let us also suppose that on each one of the 3 bookcases, the distribution of items into the 3 types is the same: 10-20-70\%, \ie, $p=(0.1,0.2,0.7)$ for each bookcase and $p=\left(\frac{0.1}{3},\frac{0.2}{3},\frac{0.7}{3},\frac{0.1}{3},\frac{0.2}{3},\frac{0.7}{3},\frac{0.1}{3},\frac{0.2}{3},\frac{0.7}{3}\right)$ for the library of 3 bookcases.
Finally, let us now suppose that, due to maintenance costs, 2 of our 3 bookcases will have to be discarded, and that we are interested in measuring the diversity that will be lost, and the diversity we will manage to preserve.
If we consider the Gini-Simpson Index (cf. Section~\ref{subsec:diversity_of_diversities}), we measure the initial diversity of our 3 bookcases at 0.82, the diversity of the saved bookcase at 0.46, and the diversity of the 2 lost bookcases at 0.73.
Paradoxically, because the Gini-Simpson Index does not satisfy the replication principle, we have saved about 56.1\% ($\frac{0.46}{0.82}$) of the initial diversity, but we have lost about 89\% ($\frac{0.73}{0.82}$) of it.
Had we taken the true diversity of order 1 for measurements, the initial diversity would have been of 6.69, while that of the saved bookcase would have been 2.23, and that of the 2 lost bookcases would have been 4.46.
Because true diversities satisfy the replication principle, this would have yielded no paradox: we would have measured a lost of 2/3 of the diversity while measuring 1/3 of the initial diversity saved.
The replication principle allows for interesting algebraic properties of diversity measures: when gathering or disassembling multiple distributions, this principle ensures that sum of diversities is preserved.
For further examples, and for a discussion of the implications of the replication principle in ecology, the reader is refered to \cite{jost2010partitioning,daly2018ecological}.

True diversities are related to several of the diversities used in different domains and identified in Section~\ref{subsec:diversity_of_diversities}.
\emph{Richness} of a distribution $p$ can be computed as the limit of $\D{\alpha}(p)$ when $\alpha\rightarrow 0^+$, observing that $p_i^\alpha \rightarrow 1$ if $p_i>0$, thus resulting in the count of effective types. We thus identify richness with 0-order true diversity, calling it \textbf{Richness diversity}. 
$\D{1}(p)$, 1-order true diversity (or \textbf{Shannon diversity}), also called {\it perplexity} \cite{ramaciotti2019role}, is related to Shannon entropy $H(p)$ of $p$ by exponentiation: $\D{1}(p)=2^{H(p)}$ when entropy is computed in base~2. 
$\D{2}(p)$, 2-order true diversity (or \textbf{Herfindhal diversity}), is the reciprocal of the Herfindhal-Hirschman Index: $\D{2}(p)={1/\HHI(p)}$. 
The Berger-Parker Index is also identified with the result of a limit process. By observing that $\D{\alpha}\xrightarrow{\alpha\rightarrow\infty} 1/\text{max}\{p_1,\ldots,p_k\}$ (Section 5.4 of \cite{aczel1975measures}) we can define 
$$
\D{\infty}(p)\coloneqq \frac{1}{\max\{p_1,\ldots,p_k\}},
$$
and thus conclude that $\D{\infty}(p)=1/\BPI(p)$ (here called \textbf{Berger diversity}).
These relations are summarized in Table~\ref{tab:summary_diversity_measures}.
In previous relations, the fact that the Herfindhal-Hirschman Index and the Berger-Parker Index are reciprocal to true diversities underlines that they are intended to measure concentration.

\begin{table}[!h]
\scriptsize
\centering
\caption{Summary of true diversities of order 0, 1, 2, and $\infty$, and their relation to classic diversity measures.}
\label{tab:summary_diversity_measures}
\begin{tabular}{|l|l|l|l|l|}
\hline
Order ($\alpha$) & Name &  \makecell[l]{True\\diversity} & Expression & \makecell[c]{Relation to other diversity measures} \\
\hline
& & & & \\
0        & Richness diversity            & $\D{0}(p)$      & $\left|\left\{i\in\{1,\ldots,k\} : p_i>0\in\right\} \right|$ & Same as richness \cite{macarthur1965patterns,gotelli2011estimating}. \\
& & & & \\
1        & Shannon diversity    & $\D{1}(p)$      & $\left(\prod\limits^{k}_{\substack{i=1\\p_i\neq 0}} p_i^{p_i} \right)^{-1}$ 
& Exponential of Shannon entropy \cite{shannon1948mathematical,shannon1963mathematical}: $H(p)=\log_2 \left(\D{1}(p) \right)$, with $H$ in base 2. \\
& & & & \\
2        & Herfindahl diversity & $\D{2}(p)$      & $\left(\sum\limits^k_{i=1}p^2_i\right)^{-1} $
& Reciprocal of the Herfindahl-Hirschman Index \cite{rhoades1993herfindahl}: $\HHI(p)=1/\D{2}(p)$. \\
& & & & \\
$\infty$ & Berger diversity     & $\D{\infty}(p)$ & $\left(\max\limits_{i\in\{1,...,k\}}\{p_i\}\right)^{-1}$ & Reciprocal of the Berger-Parker Index \cite{berger1970diversity}: $\BPI(p)=1/\D{\infty}(p)$. \\
& & & & \\
\hline
\end{tabular}
\end{table}

\begin{figure}[!h]
  \centering
  \input{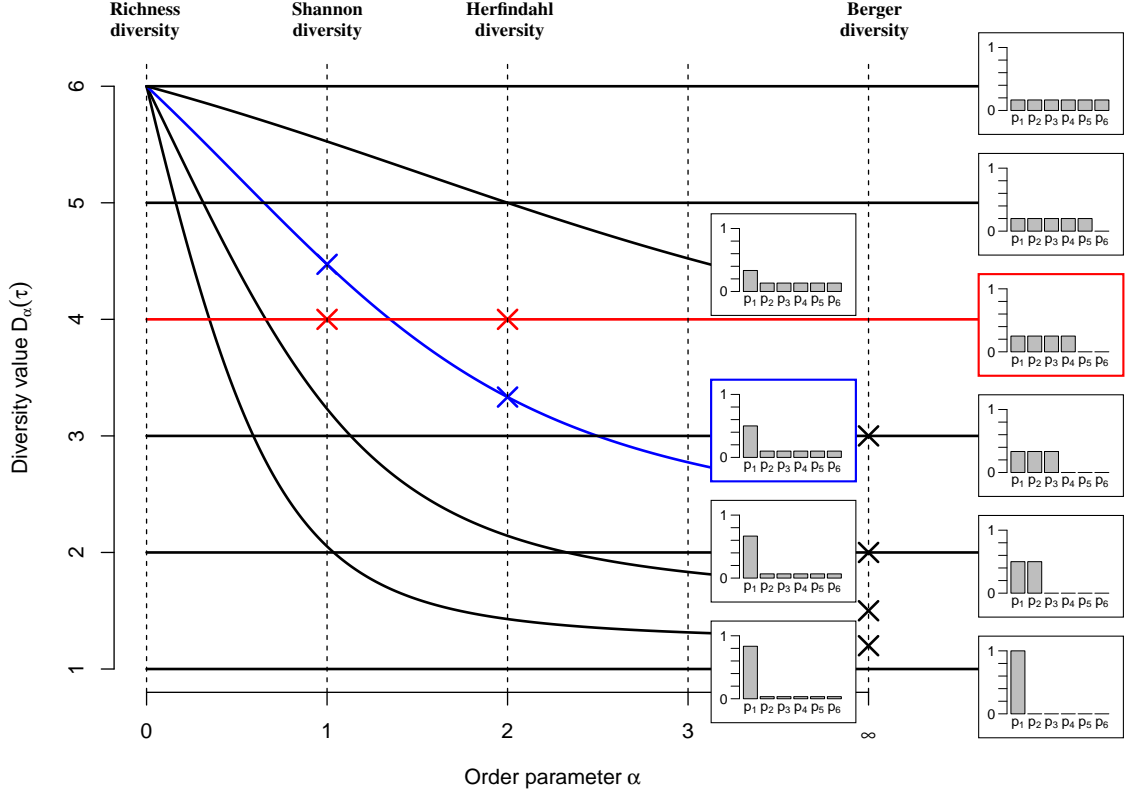}
  \caption{
    Values of different true diversities, depending on order $\alpha$, for different distributions.
  }
  \label{fig:true_diversities}
\end{figure}

Let us illustrate some of these properties in Figure~\ref{fig:true_diversities}.
By virtue of the axioms of the theory, all true diversities have equal values for uniform distributions with the same number of effective (non-empty) types. 
In this case, diversity is the number of effective types (horizontal lines in Figure~\ref{fig:true_diversities}).
However, when the distribution into types is not uniform, these measures behave differently (decreasing curves in Figure~\ref{fig:true_diversities}).
In this case, parameter $\alpha$ expresses the way non-uniformity, or \emph{balance}, is taken into account.
If $\alpha$ is low, inequalities in a distribution will only have a weak impact on diversity values, and in the extreme case where $\alpha = 0$ (\ie, for richness), inequalities in proportional abundances are not at all taken into account.
Conversely, if $\alpha$ is high, inequalities in a distribution will have a strong impact on diversity values, and in the extreme case where $\alpha \rightarrow \infty$ (\ie, for Berger diversity), only the highest abundance is taken into account.
Red and blue curves in Figure~\ref{fig:true_diversities} illustrate how parameter $\alpha$ can modulate the relative importance given to \emph{variety} and \emph{balance} (cf. Section~\ref{subsec:diversity_of_diversities}): a distribution with 6 types could be evaluated less diverse than one with 4 types if it is sufficiently unbalanced for a given value of $\alpha$.
True diversities hence allow us to have a continuum of measures which give a different weight to the \emph{variety} and \emph{balance} of distributions: $\alpha\rightarrow 0$ means that diversity takes only \emph{variety} into account, while $\alpha\rightarrow \infty$ means that diversity takes only \emph{balance} into account.

%%%%%%%%%%%%%%%%%%%%%%%%%%%%%%%%%%%%%%%%%%%%%%%%%%%%%%%%%%%%%%%%%%
\subsection{Relative true diversities}
\label{subsec:relative_true_diversities}

As with R\'enyi entropy, true diversities can be generalized to form a family of divergence measures.
\emph{Relative true diversities} generalize the family of true diversities by allowing them  to take any baseline other than the uniform distribution (that is, the distribution with maximal diversity).
In different applications, it might be interesting to measure diversity with respect to another reference distribution.
In Bayesian inference, for example, divergence of the posterior, relative to the prior probability distribution, is a measure of gained information.
Relative true diversities generalize this notion using true diversity.

This generalization is analogous to the well-known generalization of the family of \emph{R\'enyi entropies} to the family of \emph{R\'enyi divergences} \cite{renyi1961measures,van2014renyi}.
Among these generalizations, a well-known special case is the generalization of \emph{Shannon entropy} to \emph{Kullback-Leibler divergence} (also known as \emph{relative entropy})~\cite{Kullback1951, Cover2006}.

Abusing notation, we also denote $\D{\alpha}$ the $\alpha$-order relative true diversity between two distributions $p,q\in\Delta^{k-1}$, as described below.

\begin{definition}[Relative true diversity]
\label{def:relative_true_diversity}
The relative true diversity of order $\alpha$ is the application $\D{\alpha} : \Delta^* \times \Delta^* \rightarrow \mathbb{R}^+$ such that, given $p=(p_1,\ldots,p_k)\in\Delta^{k-1}$, $q=(q_1,\ldots,q_k)\in\Delta^{k-1}$, with $p_i=0$ whenever $q_i=0$, and $\alpha\in\reals^+$,
$$
\D{\alpha} (p \relativeto q) = \left( \sum\limits^k_{\substack{i=1\\q_i\neq0}} p_i^\alpha q_i^{1-\alpha} \right)^{\frac{1}{\alpha-1}} \quad \text{ if }\alpha\neq 1.
$$
As with true diversities, extreme values are defined as the result of limit processes (cf. Theorems 4, 5, \& 6 of \cite{van2014renyi}):
$$
\D{0}(p\relativeto q) \coloneqq \left|\left\{ i\in\{1,\ldots,k\} : p_i \neq 0 \text{ and } q_i \neq 0 \right\} \right|,
$$
$$
\D{1}(p \relativeto q) \coloneqq \left( \prod\limits^k_{\substack{i=1\\q_i\neq0}} {\left(\frac{p_i}{q_i}\right)^{p_i}} \right)^{-1} \text{\ with\ } {p_i}^{p_i}\coloneqq 1 \text{ if }p_i=0,\quad \text{and}\quad\D{\infty}(p \relativeto q) \coloneqq  \left( \max_{\substack{i \leq k\\q_i\neq0}} \frac{p_i}{q_i} \right)^{-1} \text{.}
$$
\end{definition}

This definition is analogous to that of true diversities with respect to R\'enyi entropy: $\D{\alpha}(p\relativeto q) = e^{H_\alpha(p\relativeto q)}$.
Thus, relative true diversities satisfy analogous properties.
If $u=(1/k,\ldots,1/k)$ is the uniform distribution, then, for $p\in\Delta^{k-1}$ we have $\D{\alpha}(p\relativeto u) = k/\D{\alpha}(p)$, and thus $\D{\alpha}(p\relativeto u)\in [1,k ]$ ($1$ when $p$ is also uniform and $k$ when $\D{\alpha}(p)$ is minimal, \ie, equal to $1$).
For a fixed $k$ and a fixed $p \in \Delta^{k-1}$, a relative true diversity is only minimal when distributions are equal. For all $p,q\in\Delta^{k-1}$
$$
\D{\alpha}(p\relativeto q)\geq \D{\alpha}(p\relativeto p),
$$
and its minimal value is $\D{\alpha}(p\relativeto p)=1$.

%%%%%%%%%%%%%%%%%%%%%%%%%%%%%%%%%%%%%%%%%%%%%%%%%%%%%%%%%%%%%%%%%%%%%%%%
\subsection{Joint distributions, additivity, and Shannon entropy}
\label{subsec:joint_distributions_additivity}

Other relevant properties of diversity measures are related to situations in which we have concurrent classifications.
Following the notation from Section~\ref{subsec:items_types_classifications}, let us consider a system in which items are classified according to two criteria, giving rise to two relations: $\tau_1\subseteq I\times T_1$ and $\tau_2\subseteq I\times T_2$.
For instance, books in a bookcase may be classified according to their genre (\eg, comics, novels) but also according to their author.

Let us define the \emph{joint membership relation} $\tau_1{\times}\tau_2 \subseteq I \times (T_1 \times T_2)$ such that $(i,(t_1,t_2)) \in \tau_1{\times}\tau_2 \Leftrightarrow (i,t_1) \in \tau_1 \wedge (i,t_2) \in \tau_2$.
Let us also define the \emph{conditional membership relation} $(\tau_2 \given t_1) \subseteq I \times T_2$ such that $(i,t_2) \in (\tau_2 \given t_1) \Leftrightarrow (i,(t_1,t_2)) \in \tau_1{\times}\tau_2$.

As in Section~\ref{subsec:items_types_classifications}, let us consider the following distributions: $p_{\tau_1}(t)=a_{\tau_1}(t)/|\tau_1|$ and $p_{\tau_2}(t)=a_{\tau_2}(t)/|\tau_2|$, resulting in $p_{\tau_1}\in\Delta^{|T_1|-1}$ and $p_{\tau_2}\in\Delta^{|T_2|-1}$.
Similarly, we define joint and conditional distributions.
We define the \emph{joint distribution} over $T_1$ and $T_2$ as 
$$p_{\tau_1\times\tau_2}(t)=\frac{a_{\tau_1\times\tau_2}(t)}{|\tau_1\times\tau_2|}, \quad\text{ with }\,p_{\tau_1\times\tau_2}\in\Delta^{(|T_1|-1)(|T_2|-1)},$$
\noindent and the \emph{conditional distribution} over $T_2$ given $t_1\in T_1$ as
$$p_{(\tau_2\given t_1)}(t)=\frac{a_{(\tau_2 \given t_1)}(t)}{\left|(\tau_2 \given t_1)\right|}, \quad \text{ for }t_1\in T_1,\text{ with }\,p_{(\tau_2\given t_1)}\in\Delta^{|T_2|-1}.$$

The first of two additivity principles considered in this article is the \emph{weak additivity principle}. 

\begin{definition}[Weak additivity]
\label{def:weak_additivity}
A diversity measure $D$ is weakly additive if and only if, for all $\tau_1$ and $\tau_2$ such that $p_{\tau_1{\times}\tau_2}(t_1,t_2) = p_{\tau_1}(t_1) p_{\tau_2}(t_2)$, we have $D\left(p_{\tau_1{\times}\tau_2}\right) = D\left(p_{\tau_1}\right) D\left(p_{\tau_2}\right)$.
\end{definition}

In other words, if two classifications are independent, then the diversity of the joint classification is equal to the product of the diversities of each separate one.

\begin{theorem}[True diversities satisfy the principle of weak additivity \cite{csiszar2008axiomatic}]
\label{thm:weak_additivity}
True diversities $\D{\alpha}$ satisfy the principle of weak additivity.
\end{theorem}

Theorem~\ref{thm:weak_additivity} is equivalent to the expression of joint R\'enyi entropy for independent variables.

A stronger property, called \emph{strong additivity principle}, and not restricted to independence between $\tau_1$ and $\tau_2$, is verified for the particular case of 1-order true diversity, that is Shannon diversity.

\begin{definition}[Strong additivity]
\label{def:strong_additivity}
A diversity measure $D$ is strongly additive if and only if, for all $\tau_1$ and $\tau_2$, we have $D\left(p_{\tau_1{\times}\tau_2}\right) = D\left(p_{\tau_1}\right) D\left(p_{\tau_2 \given \tau_1}\right)$ where $D\left(p_{\tau_2 \given \tau_1}\right) = \prod_{t_1 \in T_1} {{D\left(p_{\tau_2 \given t_1}\right)}^{p_{\tau_1}\left(t_1\right)}}$.
\end{definition}

In other words, the diversity of the joint classification is equal to the diversity of the first classification multiplied by the diversity of the second classification conditioned by the knowledge of the first one.
\emph{Conditional diversity} is the weighted geometric mean of the diversities of conditional distributions.

\begin{theorem}[1-order true diversity is strongly additive~\cite{csiszar2008axiomatic}]
\label{thm:true_diversities_are_strongly_additive}
1-order true diversity $\D{1}$ satisfies the principle of strong additivity.
\end{theorem}

The principle of strong additivity is analogous to the well-known \emph{chain rule} between \emph{conditional entropy} and \emph{joint entropy} in information theory (cf. Section 2.5 in~\cite{Cover2006}):  $H(X,Y) = H(X) + H(Y|X)$ for random variables $X$ and $Y$.

\begin{figure}
  \centering
  \begin{tikzpicture}[scale=0.6]

  % Outer shell: application
  \draw [thick] (-2,1.5) ellipse (7.5cm and 6cm);
  \node [align=center, font=\bf\footnotesize] at (-2,6.6) {{  $D:\Delta^*\rightarrow\reals^+$}\\{Applications from simplices to $\reals^+$}};

  % First shell: 4 axioms
  \draw [thick] (-2,0.85) ellipse (6.5cm and 5cm);
  \node [align=center, font=\bf\footnotesize] at (-2,4.9) {{Diversities $D$ that} \\{satisfy Ax. \ref{ax:symmetry}, \ref{ax:expansibility}, \ref{ax:transfer} \& \ref{ax:normalization}}};
\node [align=center, font=\bf\footnotesize, blue] (axiom1) at (-11.5,6.5) {{ Symmetry, Expansibility}\\ {\footnotesize Normalisation, Transfer principle} \\ { $\Rightarrow$ Merging (Th.~\ref{thm:merging}) \& Bounds (Th.\ref{thm:bounds_for_diversities})}};
  \draw [->, thick, blue] (axiom1.south) to +(5.5,-1.75);

  % Second shell: SWQLM
  \draw [thick] (-2,0.20) ellipse (5.5cm and 4.0cm);
  \node [align=center, font=\bf\footnotesize] at (-2,3.15) {{Reciprocal self-weighted}\\{quasilinear means}};
  %\node [align=center, font=\small] at (-0.5,3.55) {$\phi^{-1} \! \left( \displaystyle\sum_{i \leq k} {p_i \phi (p_i)} \right)$};

  \draw [thick] (-2,-0.5) ellipse (4.5cm and 3cm);
  \node [align=center, font=\bf\footnotesize] at (-2,1.8) {True diversities};

  \node [align=center, font=\bf\footnotesize, blue] (axiom2) at (-8.5,-4.25) {{ Replication principle}\\ {$\Rightarrow$ Weak additivity (Th.~\ref{thm:weak_additivity})}};
  \draw [->, thick, blue] (axiom2.north) to +(4.1,1.1);

  \node [align=center, font=\normalsize] at (-3.5,1) {{\footnotesize Richness}};
  \node [align=center, font=\normalsize] at (-4.2,-0.3) {{\footnotesize Herfindahl diversity}};
  \node [align=center, font=\normalsize] at (-3.5,-1.6) {{\footnotesize Berger diversity}};

  \draw [thick] (0,-0.5) ellipse (1.5cm and 1cm);
  \node [align=center, font=\normalsize] at (0,-0.5) {{\footnotesize Shannon}\\{\footnotesize  diversity}};

  \node [align=center, font=\bf\footnotesize, blue] (axiom3) at (4.5,-4) {{Strong additivity (Th.~\ref{thm:true_diversities_are_strongly_additive})}};
  \draw [->, thick, blue] (axiom3.north) to +(-3.5,3);
\end{tikzpicture}

%%% Local Variables:
%%% mode: latex
%%% TeX-master: "../main"
%%% End:
  \caption{
    Relations between the different families of diversity measures, and their most important properties.
  }
  \label{fig:all_diversities}
\end{figure}
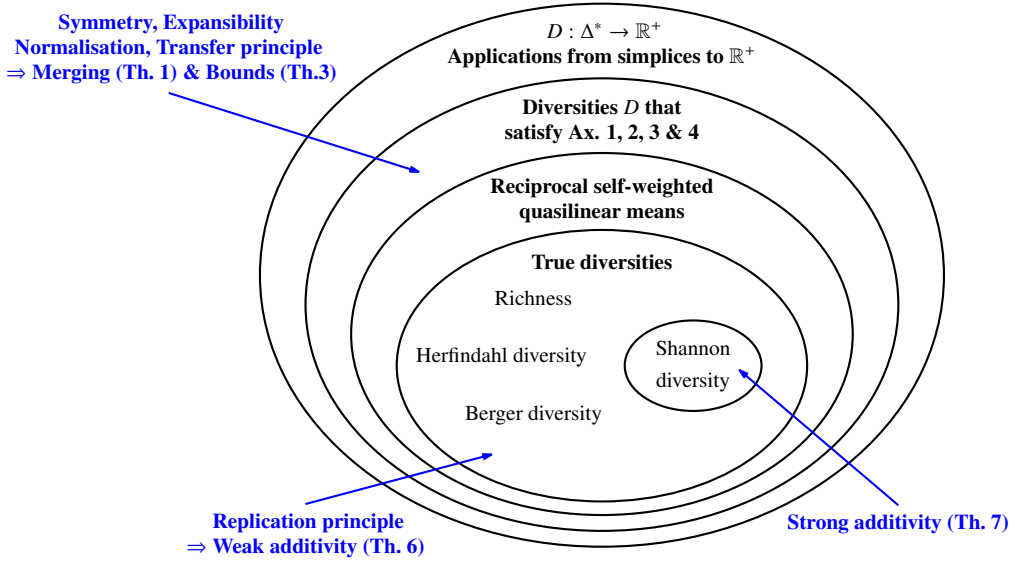

Theorem~\ref{thm:true_diversities_are_strongly_additive} will justify the use of 1-order diversities in some results regarding the relations of different \emph{network diversity measures} in the next sections.
Figure~\ref{fig:all_diversities} summarizes and illustrates the relations between the different families of diversity measures from this section, along with their most important properties.

\section{Random Walks in Heterogeneous Information Networks}
\label{sec:part3}

In the previous section, we presented a broad definition of diversity, which we then narrowed to a particular family of measures that share relevant properties captured by axioms.
%
% These properties were taken from requirements found in the scientific literature.
%
The functions of the theory determined by these axioms resulted in true diversities, which are connected to many of the diversity measures used in different domains of research.

When considering complex systems and network-structured data, different distribution functions can be computed.
One example is probability distributions over vertices resulting from a random walk.
Different diversity measures may be computed over these distributions.
In this article, we develop a single framework for both operations, effectively covering and summarizing the measurement of diversity in networks in several domains.
In order to do so, we develop in this section a formalism for the treatment of networks that are relevant for fields concerned by the concept of diversity.

Developed within graph theory, heterogeneous information networks \cite{sun2013mining,sun2009ranking} (equivalent to directed graphs with colored vertices and edges) have recently been used to provide ontologies to represent complex unstructured data in a wide gamut of applications (knowledge graphs are prominent examples of their flexibility~\cite{bollacker2008freebase,miller1995wordnet,ashburner2000gene}).
In this work, we will consider an extended model of heterogeneous information network, using multigraphs (graphs for which multiple edges might exist between any given couple of vertices), for the development of a framework for measuring diversity in networks.
As we shall see in more detail in Section~\ref{sec:part5}, many situations encountered in practice can be represented using heterogeneous information networks.
For example, when modeling the consumption of news on a website, the situation may be represented as users selecting articles, and articles having specific categories (business, culture, sports, etc.). 
This translates to a heterogeneous information network with three vertex types (users, articles, categories) and two edge types (users select articles, articles belong to categories).

%%%%%%%%%%%%%%%%%%%%%%%%%%%%%%%%%%
\subsection{Preliminary notations}
\label{subsec:notations}

We consider a multigraph composed of a set of nodes $\V$, linked by a set of directed edges $\E$.
We propose the following system of capitalization and typefaces to reference different objects:
\begin{itemize}
  \item \emph{vertices} and \emph{edges} are designated by {lowercase} letters, $v$ and $e$;
  \item a set of \emph{types of vertex} is designated by $\A$;
  \item a set of \emph{types of edge} is designated by $\R$;
  \item types in $\A$ are notated with uppercase letters $\vertexlabel$, types in $\R$ are denoted with uppercase letters $\edgelabel$;
	\item \emph{vertex sets} and \emph{edge sets} are notated by {uppercase} letters $V$ and $E$;
	\item sets of \emph{vertex types} and \emph{edge types} labels are notated by {calligraphic} letters $\calV$ and $\calE$;
  \item random variables with support on sets of vertices are notated by the capital letter $X$.
\end{itemize}

%%%%%%%%%%%%%%%%%%%%%%%%%%%%%%%%%%%%%%%%%%%%%%%%%%
\subsection{Heterogeneous information networks}
\label{subsec:hins}

In contrast to traditional formalizations of heterogeneous information networks \cite{sun2013mining,sun2009ranking}, we propose the use of multigraphs for generality.
A \emph{multigraph} $G$ is a couple $(\V,\E)$ where $\V = \{v_1,\ldots,v_n\}$ is a set of vertices and $\E = \{e_1,\ldots,e_m\}$ is a set of directed edges that is a multisubset of $\V \times \V$.
Given an edge $e \in \E$, we denote $\sv{e}$ its source vertex and $\dv{e}$ its destination vertex such that $(\sv{e},\dv{e}) \in \V \times \V$.

We also denote $\epsilon : \V \times \V \rightarrow \mathbb{N}$ the multiplicity function of edges, that is the function counting the number of edges in $\E$ that link any two vertices:
$\epsilon(v_1,v_2) = | \{ e \in \E : \sv{e} = v_1 \wedge \dv{e} = v_2 \} |$.
We also define:
\begin{itemize}
\item $\epsilon(v_1,-) \coloneqq \displaystyle\sum_{v_2 \in \V} {\epsilon(v_1,v_2)}$ the \emph{out-degree} of vertex $v_1$;
\item $\epsilon(-,v_2) \coloneqq \displaystyle\sum_{v_1 \in \V} {\epsilon(v_1,v_2)}$ the \emph{in-degree} of vertex $v_2$;
\item $\epsilon(-,-) \coloneqq \displaystyle\sum_{(v_1,v_2) \in \V \times \V} {\epsilon(v_1,v_2)}$ the \emph{total number} of edges.
\end{itemize}

We now define heterogeneous information networks using multigraphs.
Classical heterogeneous information networks can be easily accounted for by constraining the multiplicity of edges.

\begin{definition}[Heterogeneous information network]
\label{def:hins}
A \emph{heterogeneous information network} $\calG = (\V,\E,\A,\R,\vertexmap,\edgemap)$ is a multigraph $(\V,\E)$, with a vertex labeling function $\vertexmap:\V\rightarrow\A$ and an edge labeling function $\edgemap:\E\rightarrow\R$, such that edges with the same type in $\R$ have their source vertices mapped to the same type in $\A$ and their destination vertices mapped to the same type in $\A$: 
$$
\forall \, e,e' \in \E, \; \left(\quad \edgemap(e) = \edgemap(e') \quad \Rightarrow \quad  \left(\; \vertexmap (\sv{e}) = \vertexmap (\sv{e'}) \quad \wedge \quad \vertexmap (\dv{e}) = \vertexmap (\dv{e'})\;\right)\quad \right)\text{.}
$$
\end{definition}

Label functions $\vertexmap$ and $\edgemap$, that map vertices to vertex types and edges to edge types, induce a partition in the set of vertices and a partition in the set of edges.
If $\A=\{A_1,\ldots,A_N\}$ and $\R=\{R_1,\ldots,R_M\}$, $\vertexmap$ and $\edgemap$ induce partitions $\calV=\{V_1,\ldots,V_N\}$ on $\V$ and $\calE=\{E_1,\ldots,E_M\}$ on $\E$.
These partitions are such that $\forall v\in\V, \;(\vertexmap(v)=A_i \Leftrightarrow v\in V_i)$ and $\forall e\in\V,\;(\vertexmap(e)=R_j \Leftrightarrow e\in E_j)$.
Thus, abusing notation, we make indistinct use of types in $\A$ and sets in $\calV$, and of types in $\R$ and sets in $\calE$ when this is not ambiguous.

Given an edge type $E \in \calE$, we denote $\sV{E} \in \calV$ its source-vertex type and $\dV{E} \in \calV$ its destination-vertex type.
We also denote $\epsilon_{E} : \sV{E} \times \dV{E} \rightarrow \mathbb{N}$ the specialization of $\epsilon$ on $E$, that is, the function counting the number of edges in $E$ going from a given vertex in $\sV{E}$ to a given vertex in $\dV{E}$:
$$\epsilon_{E}(v_1,v_2) = | \{ e \in E : \sv{e} = v_1 \wedge \dv{e} = v_2 \} | \text{.}$$
As before, we also define:
\begin{itemize}
\item $\epsilon_E(v_1,-) \coloneqq \displaystyle\sum_{v_2 \in \dV{E}} {\epsilon_E(v_1,v_2)}$ is the \emph{out-degree} of $v_1$ among edges in $E$;
\item $\epsilon_E(-,v_2) \coloneqq \displaystyle\sum_{v_1 \in \sV{E}} {\epsilon_E(v_1,v_2)}$ is the \emph{in-degree} of $v_2$ among edges in $E$;
\item $\epsilon_E(-,-) \coloneqq \displaystyle\sum_{(v_1,v_2) \in \sV{E} \times \dV{E}} {\epsilon_E(v_1,v_2)}$ is the \emph{number} of edges in $E$.
\end{itemize}

Following the example of existing definitions for heterogeneous information networks~\cite{sun2013mining,sun2009ranking,shi2016survey}, we define the \emph{network schema}.
Consistency in the direction of edges belonging to the same edge type allows for the definition of schemas as proper directed graphs.
Figure~\ref{fig:schema_and_path} illustrates a heterogeneous information network and its network schema.

\begin{definition}[Network schema]
\label{def:network_schema}
The network schema of a heterogeneous information network $\calG = (\V,\E,\A,\R,\vertexmap,\edgemap)$ is the directed graph $\mathcal{S}=(\calE,\calV)$ that has vertex types $\calV$ for vertices and edge types $\calE$ for edges.
\end{definition}

Knowledge graphs (\eg, Google's Knowledge Graph \cite{singhal2012introducing}) are knowledge-based systems closely related to heterogeneous information networks.
They are used to store complex structured and unstructured data in the form of a network, based on the Resource Description Framework (RDF) \cite{W3Crdf}, which models data as entries of the form
\textit{$\langle$Subject, Property, Object$\rangle$}.
If edges of a same type always link source vertices of the same type with target vertices of the same type (cf. Definition~\ref{def:hins}), it is easy to see that identifying a \textit{Property} in the RDF data model with an edge type allows for the identification of a knowledge graph with a heterogeneous information network \cite{shi2016survey}.
Early pairings of the two concepts were proposed to leverage the heterogeneous information network formalism in data mining tasks in knowledge graphs \cite{han2013mining}.
While some works have equated these two closely similar concepts \cite{nickel2015review}, most insist in differentiating heterogeneous information networks as a mathematical formalism suitable for the treatment of data mining problems using knowledge graph data \cite{zheng2017entity,cao2018heterogeneous}.

Let us now define the probability of transitioning between vertices randomly following the available directed edges from an edge type.

\begin{definition}[Probability of transitioning between vertices in an edge type]
\label{def:transition_probability_edge_type}
Given an edge type $E \in \calE$, assuming that each vertex in $\sV{E}$ is connected to at least one vertex in $\dV{E}$, \ie, $\forall v_1 \in \sV{E}\,\left(\epsilon_E(v_1,-) > 0\right)$, we denote by $p_{E} : \sV{E} \times \dV{E} \rightarrow [0,1]$ the \emph{transition probability} of the random walk following edges in $E$, for all $(v_1,v_2) \in \sV{E} \times \dV{E}$, as
$$p_{E} (v_2 \given v_1) \coloneqq \frac{\epsilon_{E}(v_1,v_2)} {\epsilon_{E}(v_1,-)} \text{.}$$
\end{definition}

\begin{definition}[Random transition between vertices in an edge type]
For an edge type $E\in\calE$ going from vertex type $\sV{E}$ to vertex type $\dV{E}$ in $\calV$, we denote the transition from a random vertex $X_{src}\in \sV{E}$ to a random vertex $X_{dst}\in \dV{E}$, following probability distribution $p_E$, as $X_{src} \xrightarrow{E} X_{dst}$.
\end{definition}

As a consequence of Definition~\ref{def:transition_probability_edge_type}, $\forall v_1 \in \sV{E}$,\, $p_{E}(\,\cdot \given v_1):\dV{E}\rightarrow \reals^+$ is a probability distribution on $\dV{E}$.
For all $v_2 \in \dV{E}$, we have $p_{E}(v_2 \given v_1) \in [0,1]$ and $\sum_{v_2 \in \dV{E}} {p_{E}(v_2 \given v_1)} = 1$.

In the case where vertex $v_1 \in \sV{E}$ is not connected to any vertex in $\dV{E}$ (\ie, when $\epsilon_E(v_1,-) = 0$), $p_{E}(v_2 \given v_1)$ cannot be defined as above.
This situation can be remedied by adding a \text{sink vertex} to each vertex type.
For every $E\in\calE$, an edge $e^s_E$ is added such that $\sv{e^s_E}$ is the sink vertex in $\sV{E}$ and such that $\dv{e^s_E}$ is the sink vertex in $\dV{E}$.
Then, vertices in $\sV{E}$ connected to no vertex in $\dV{E}$ can be connected to the sink vertex.
In the rest of this article we will assume that this procedure has been applied if needed and that for every $E\in\calE$ there are no vertices in $\sV{E}$ that are not connected to at least one vertex in $\dV{E}$.

%%%%%%%%%%%%%%%%%%%%%%%%%%%%%%%%%%%%%%%%%%%%%%%%%%
\subsection{Meta paths and constrained random walks}
\label{subsec:restricted_random_walks}

Random walks in heterogeneous information networks can be constrained \cite{lao2010relational,lao2010fast} to follow a specific sequence of edge types, called \emph{\metapath{}} \cite{shi2016survey,shi2017heterogeneous}.
This enables for the computation of the probability distribution of the ending vertex of a random walker constrained to a specific \metapath{}.
The variety and combinatorics of \metapath{}s will be the origin of the network diversity measures that we propose in the next section.

For the definition of meta paths, we will first consider sequences on the set $\R$ of edge types.
We denote by \emph{sequence of length $k$ for $M$} ($M=|\R|$) a $k$-tuple $r=(r_1,\ldots,r_k)$ such that for all $i\in\{1,\ldots,k\}$ we have $r_i\in\{1,\ldots,M\}$.

\begin{definition}[Meta path]
Given a heterogeneous information network  $\calG = (\V,\E,\A,\R,\vertexmap,\edgemap)$ and a sequence $r$ of length $k\in\naturals$ for $M=|\R|$, a \emph{\metapath{}} of length $k$ is the $k$-tuple $\Pi = (E_{r_1}, \ldots, E_{r_k}) \in \calE^k$ of $k$ edge types (with possible repetitions) such that the source vertex type of an edge type is the destination vertex type of the previous one in the $k$-tuple $\Pi$:
\ie, $\forall 1 \leq i \leq k, \; \sV{E_{r_i}} = \dV{E_{r_{i-1}}}$.
\end{definition}

\begin{figure}[h]
\centering
  \centering
\begin{tikzpicture}

  % Left subfigure (HIM)
  %%%%%%%%%%%%%%%%%%%%%%

  % vertices

  \node [vertex] (i1) {$v_0^1$};
  \node [vertex, right = 0.25cm of i1] (i2) {$v_0^2$};

  \node [vertex, below left = 1.25cm and 0.0cm of i1] (i3) {$v_1^1$};
  \node [vertex, right = 0.25cm of i3] (i4) {$v_1^2$};
  \node [vertex, right = 0.25cm of i4] (i5) {$v_1^3$};

  \node [vertex, below left = 1cm and 0.25cm of i3] (i6) {$v_2^1$};
  \node [vertex, below right = 0.25cm and 0.25cm of i6] (i7) {$v_2^2$};

  \node [vertex, below right = 1cm and 0.25cm of i5] (i8) {$v_3^2$};
  \node [vertex, below left = 0.25cm and 0.25cm of i8] (i9) {$v_3^1$};

  % ellipses

  \draw [dashed] (0.5,0) ellipse (1.1cm and 0.65cm);
  \draw [dashed] (0.5,-1.8) ellipse (1.6cm and 0.75cm);
  \draw [dashed,rotate=-45] (1.95,-3.25) ellipse (1.2cm and 0.65cm);
  \draw [dashed,rotate=45] (-1.35,-3.95) ellipse (1.2cm and 0.65cm);

  % edges

  \drawLink[]{i1}{i3}{1}; % E0
  \drawLink[]{i1}{i4}{1}; % E0
  \drawLink[]{i1}{i5}{1}; % E0
  \drawLink[]{i2}{i5}{2}; % E0

  \drawLink[]{i3}{i6}{2}; % E1
  \drawLink[]{i3}{i7}{1}; % E1
  \drawLink[]{i3}{i9}{1}; % E1

  \drawLink[]{i4}{i9}{1}; % E2
  \drawLink[]{i4}{i9}{1}; % E2
  \drawLink[]{i5}{i8}{1}; % E2

  % vertex/edge type names
  \node [vertexset, left = 0.35cm of i1] (V0) {$V_0$};
  \node [vertexset, left = 0.35cm of i3] (V1) {$V_1$};
  \node [vertexset, left = 0.35cm of i6] (V2) {$V_2$};
  \node [vertexset, right = 0.15cm of i8] (V3) {$V_3$};

  \node [vertexset, below left = 0.1cm and 0.25cm of i1] (E0) {$E_0$};
  \node [vertexset, below left = 0.00cm and 0.25cm of i3] (E1) {$E_1$};
  \node [vertexset, below right = 0.00cm and 0.25cm of i5] (E2) {$E_2$};

  % Left subfigure (HIM schema)
  %%%%%%%%%%%%%%%%%%%%%%%%%%%%%

  \begin{scope}[xshift = 6cm]
    
    \node [vertex] (v0) {$V_0$};
    \node [vertex, below = 1cm of v0 ] (v1) {$V_1$};
    \node [vertex, below left = 1.2cm and 0.5 of v1 ] (v2) {$V_2$};
    \node [vertex, below right = 1.2cm and 0.5 of v1 ] (v3) {$V_3$};

    \drawLink[]{v0}{v1}{1};
    \drawLink[]{v1}{v2}{1};
    \drawLink[]{v1}{v3}{1};

    \node [vertexset, below left = 0.1cm and 0.0cm of v0] (e0) {$E_0$};
    \node [vertexset, below left = 0.00cm and 0.25cm of v1] (e1) {$E_1$};
    \node [vertexset, below right = 0.00cm and 0.25cm of v1] (e2) {$E_2$};

  \end{scope}

  % Left subfigure (HIM schema)
  %%%%%%%%%%%%%%%%%%%%%%%%%%%%%

\begin{scope}[xshift = 11cm]
    
    \node [vertex,marked=1] (v0bis) {$V_0$};
    \node [vertex,marked=1, below = 1cm of v0bis ] (v1bis) {$V_1$};
    \node [vertex, below left = 1.2cm and 0.5 of v1bis ] (v2bis) {$V_2$};
    \node [vertex,marked=1, below right = 1.2cm and 0.5 of v1bis ] (v3bis) {$V_3$};

    \drawLink[marked=0.4]{v0bis}{v1bis}{1};
    \drawLink[]{v1bis}{v2bis}{1};
    \drawLink[,marked=0.4]{v1bis}{v3bis}{1};

    \node [vertexset, below left = 0.1cm and 0.0cm of v0bis] (e0bis) {$E_0$};
    \node [vertexset, below left = 0.00cm and 0.25cm of v1bis] (e1bis) {$E_1$};
    \node [vertexset, below right = 0.00cm and 0.25cm of v1bis] (e2bis) {$E_2$};

    \node [vertexset, below left = 0.1cm and 0.0cm of v3bis] (pi) {$\Pi=(E_0,E_2)$};

  \end{scope}

\end{tikzpicture}

%%% Local Variables:
%%% mode: latex
%%% TeX-master: "../main"
%%% End:
  \caption{A heterogeneous information network (left), its network schema (center), and a \metapath{} $\Pi$ on the network schema (right).}
\label{fig:schema_and_path}
\end{figure}
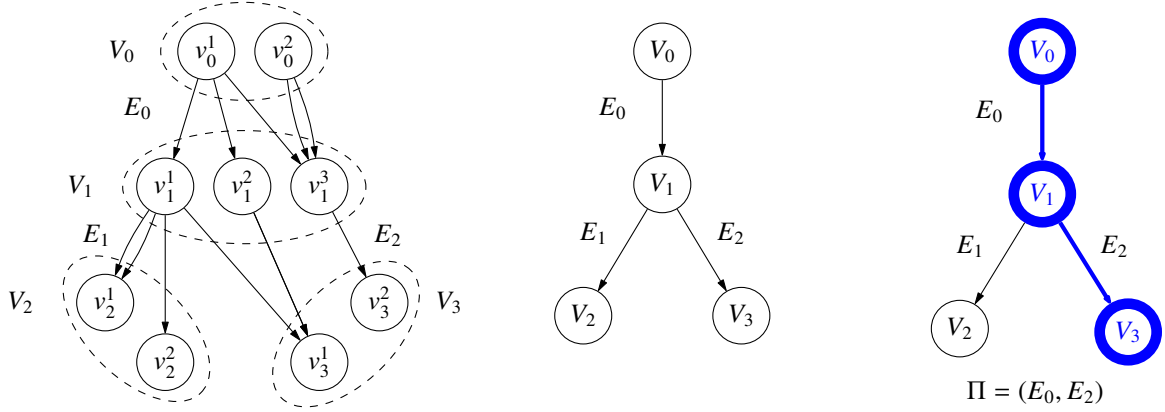

We denote by $\sV{\Pi} = \sV{E_{r_1}}$ the source vertex type of path type $\Pi$, and by $\dV{\Pi} = \dV{E_{r_k}}$ its destination vertex type.
Figure~\ref{fig:schema_and_path} provides an illustration of a heterogeneous information network and a \metapath{} on its network schema.

Using the notion of \metapath{}, we define a random walk restricted to it.

\begin{definition}[Random walk constrained to a \metapath{}]
\label{def:random_walk}
Given a \metapath{} $\Pi = (E_{r_1}, \ldots, E_{r_k})$ of length $k$ and a random variable $X_0 \in \sV{\Pi}$ representing the starting position of a random walk in vertex type $\sV{\Pi}$, the \emph{associated random walk restricted to $\Pi$} is a sequence of $k+1$ random variables $(X_0,X_1,\ldots,X_k)$ resulting from the sequential random transition between vertices in the edge types (cf. Definition~\ref{def:transition_probability_edge_type}) of $\Pi$:
$$
X_{0} \xrightarrow{E_{r_1}} X_{1} \xrightarrow{E_{r_2}} X_{2} \xrightarrow{E_{r_3}} \cdots \xrightarrow{E_{r_k}} X_{k},
$$
\noindent where, for all $i$, $X_i\in\dV{E_{r_i}}$.
\end{definition}

This is known as a \emph{path-constrained random walk} in the information retrieval community \cite{lao2010relational,lao2010fast}.
It follows from Definitions~\ref{def:transition_probability_edge_type} and \ref{def:random_walk} that a random walk restricted to a \metapath{} $\Pi$ of length $k$ is a Markov chain with transition probabilities defined as
$$
\Pr(X_i = v_i \given X_{i-1} = v_{i-1}) = P_{E_{r_i}}(v_i\given v_{i-1}),
$$
for $v_{i-1} \in \sV{E_{r_i}}$ and $v_{i} \in \dV{E_{r_i}}$.

For the next two definitions, we consider a \metapath{} $\Pi = (E_{r_1}, \ldots, E_{r_k})$ of length $k$ and its associated random walk restricted to $\Pi$, \ie, the sequence $(X_0,X_1,\ldots,X_k)$ of random variables.
The probability distribution in $\dV{\Pi}$ of the random walk's ending vertex plays a central role in the network diversity measures that will be proposed in the next section.
Let us define the conditional and the unconditional probability distributions.

\begin{definition}[Conditional probability distribution for random walks]
\label{def:conditional_probability_random_walk}
The conditional probability distribution of $X_k\in\dV{\Pi}$, that is, the destination vertex of the random walk constrained to $\Pi$, given that it started in $v_0\in\sV{\Pi}$ (\ie, $X_0=v_0$), is denoted by $p_\Pi(v_k\given v_0)$ for $v_k\in\dV{\Pi}$ and can be recursively computed as follows:
\begin{align*}
  p_{\Pi} (v_k\given v_0) \quad & = \quad \Pr(X_k=v_k\given X_0=v_0)\\
                      & = \quad \sum_{v_1 \in \dV{E_{r_1}}} { p_{(E_{r_2},\ldots,E_{r_k})}(v_k \given v_1) \; p_{E_{r_1}}(v_1|v_0)}.
\end{align*}
We will also designate by $p_{\Pi|v_0}(v_k)$ the distribution $p_{\Pi}(v_k|v_0)$ over the vertices of $\dV{\Pi}$.
\end{definition}

Using conditional probability distribution, the unconditional probability can be computed.

\begin{definition}[Unconditional probability distribution for random walks]
\label{def:unconditional_probability_random_walk}
The unconditional probability distribution of $X_k\in\dV{\Pi}$, that is, the destination vertex of the random walk restrained to $\Pi$, is denoted by $p_\Pi(v_k)$ for $v_k\in\dV{\Pi}$ and can be computed applying the law of total probability to conditional distribution $p_{\Pi\given v_0}$ as follows:
\begin{align*}
  p_{\Pi} (v_k) \quad & = \quad \Pr(X_k=v_k)\\
                      & = \quad \sum_{v_0 \in \sV{\Pi}} {p_{\Pi\given v_0}(v_k) \; \Pr(X_0=v_0)}.
\end{align*}
\end{definition}

In Definition~\ref{def:unconditional_probability_random_walk}, the dependence of $p_{\Pi}$ on $\Pr(X_0=v_0)$ (the probability distribution for the starting vertex) is explicit.

We now consider the edges resulting from the projection of all edge types in a \metapath{} $\Pi$.
This operation, related to the counting of paths in \metapath{}s, is used in the literature in related measures, such as the construction of similarity metrics for vertex searches \cite{sun2011pathsim} or for recommender systems \cite{yu2014personalized}.

\begin{definition}[Projection of a \metapath{}]
\label{def:metapath_projection}
Given a \metapath{} $\Pi = (E_{r_1}, \ldots, E_{r_k})$, we denote by $E_\Pi$ the set of edges going from vertices in $\sV{\Pi}$ to vertices in $\dV{\Pi}$, and resulting from the projection of all paths in \metapath{} $\Pi$.
We denote $\epsilon_\Pi(v_0,v_k)$ the number of paths starting at $v_0 \in \sV{\Pi}$ and ending at $v_k \in \dV{\Pi}$ that are part of \metapath{} $\Pi$. 
It is recursively computed as follows:
$$
\epsilon_{E_\Pi} (v_0, v_k) \quad = \quad \sum_{v_1 \in \dV{E_{r_1}}} {\epsilon_{E_{r_1}} (v_0, v_1) \; \epsilon_{(E_{r_1}, \ldots, E_{r_k})} (v_2, v_k)},
$$
with $\epsilon_{(E_{r_k},E_{r_k})} = \epsilon_{E_{r_k}}$.
\end{definition}

The projection is such that there is an edge in $E_\Pi$ for each path in $\Pi$.
This allows for the definition of a --one step-- random walk from $\sV{\Pi}$ to $\dV{\Pi}$.
Its probability distribution is denoted $p_{E_\Pi}$ and computed following Definition~\ref{def:transition_probability_edge_type}.
If random walk $X_0\xrightarrow{E_{r_1}}X_1\xrightarrow{E_{r_2}}\cdots\xrightarrow{E_{r_k}}X_k$ involves choosing a random edge at each vertex type $\sV{E_{r_i}}$, random walk $X_0\xrightarrow{E_{\Pi}}X_k$ involves randomly choosing one path among all possible paths in $\Pi$.

\section{Network Diversity Measures}
\label{sec:part4}

In the previous section, we established a formal framework for heterogeneous information networks within which we defined \metapath{}s and random walks constrained to them. 
This allowed us to consider different probability distributions related to these random walks.
In this section, we apply true diversity measures to these distributions, completing the framework for the measurement of diversity in heterogeneous information networks.

Depending on the chosen \metapath{}s, one can compute several diversities in a network.
These diversities will correspond to different concepts related to the structure of vertices and edges in the \metapath{}s: \emph{individual}, \emph{collective}, \emph{relative}, \emph{projected}, and \emph{backward} diversity.
All of these will be defined in this section.
These concepts will in turn have different semantical content depending on what is being modeled by the heterogeneous information network.
The way in which diversities associated with \metapath{}s may correspond to different concepts will be made clear in this section, and illustrated through different applications in the next section.

All definitions and results refer to a heterogeneous information network $\calG = (\V,\E,\A,\R,\vertexmap,\edgemap)$, and a \metapath{} $\Pi = (E_{r_1}, \ldots, E_{r_k})$ of length $k$ going from vertex type $\sV{\Pi}$ to vertex type $\dV{\Pi}$. 
In the scope of this section, let us define $\Vstart=\sV{\Pi}$ and $\Vend=\dV{\Pi}$ for ease of notation.
For diversities defined here, we will talk about the diversity of a given vertex type with respect to another one in a heterogeneous information network and along a given \metapath{}. In other words, given $\calG$, we define network diversities for a given $\Pi$ that are of the form: \emph{$\Vend\,$ diversity of $\Vstart\,$ along $\Pi$}. 

%%%%%%%%%%%%%%%%%%%%%%%%%%%%%%%%%%%%%%%%%%
\subsection{Collective and individual diversities}

The collective $\Vend$ diversity of vertices $\Vstart$ along \metapath{} $\Pi$ is the diversity of the probability distribution on vertices of $\Vend$ resulting from a random walk starting at a random vertex in $\Vstart$ and restricted to \metapath{} $\Pi$. Using previous definitions, we formally define this quantity.

\begin{definition}[Collective diversity]
\label{def:collective_diversity}
Given the probability distribution $\Pr(X_0)$ of starting at a random vertex $X_0 \in \Vstart$, we define the \emph{collective $\Vend$ diversity of $\Vstart$ along $\Pi$} as the true diversity of the probability distribution of the ending vertex of the constrained random walk.
We denote it as $\D{\alpha} \left( X_0 \xrightarrow{\Pi} X_k \right)$ and compute it as follows:
$$\D{\alpha} \left( X_0 \xrightarrow{\Pi} X_k \right) \quad = \quad \D{\alpha}(p_\Pi) \text{.}$$
\end{definition}

\begin{figure}
  \centering
  \centering
\begin{tikzpicture}
  \node [vertexset] (V0) {$V_0$};
  \node [vertexset, below = 0.5cm of V0] (V1) {$V_1$};
  \draw [->] (V0) -- node [right] {$E_1$} (V1);

  \begin{scope}[xshift = 0.5cm]
    \node [vertexset, opacity = 0] (V0) {};
    \node [vertexset, opacity = 0, below = 0.5cm of V0] (V1) {};
    
    \node [vertex, right = 1cm of V0] (i1) {$v_0^1$};
    \node [vertex, right = 0.5cm of i1] (i2) {$v_0^2$};

    \node [vertex, opacity = 0] (i) at ($0.5*(i1)+0.5*(i2)$) {};    
    \node [vertex, below = 0.5cm of i] (t2) {$v_1^2$};
    \node [vertex, left = 0.5cm of t2] (t1) {$v_1^1$};
    \node [vertex, right = 0.5cm of t2] (t3) {$v_1^3$};

    \drawLink[]{i1}{t1}{3};
    \drawLink[]{i1}{t2}{1};
    \drawLink[]{i2}{t2}{1};
    \drawLink[]{i2}{t3}{1};
  \end{scope}

  \begin{scope}[xshift = 5cm]
    \node [vertexset, opacity = 0] (V0) {};
    \node [vertexset, opacity = 0, below = 0.5cm of V0] (V1) {};
    
    \node [vertex, marked = 1/2, right = 1cm of V0] (i1) {$\frac{1}{2}$};
    \node [vertex, marked = 1/2, right = 0.5cm of i1] (i2) {$\frac{1}{2}$};

    \node [vertex, opacity = 0] (i) at ($0.5*(i1)+0.5*(i2)$) {};
    \node [vertex, marked = 3/8, below = 0.5cm of i, font=\small] (t2) {$\frac{1}{8}\!\!+\!\!\frac{2}{8}$};
    \node [vertex, marked = 3/8, left = 0.5cm of t2] (t1) {$\frac{3}{8}$};
    \node [vertex, marked = 1/4, right = 0.5cm of t2] (t3) {$\frac{2}{8}$};

    \drawLink[marked = 1/8]{i1}{t1}{3};
    \drawLink[marked = 1/8]{i1}{t2}{1};
    \drawLink[marked = 1/4]{i2}{t2}{1};
    \drawLink[marked = 1/4]{i2}{t3}{1};

    \node [above = 0.25cm of i] {$\Pr(X_0) = \left( \frac{1}{2}, \frac{1}{2} \right)$};
    \node [below = 0.25cm of t2] {$\Pr(X_1) = \left( \frac{3}{8}, \frac{3}{8}, \frac{2}{8} \right)$};
    \node [below = 1cm of t2] {$\D{1} \left( X_0 \xrightarrow{E_1} X_1 \right) \approx 2.95$};
  \end{scope}

  \begin{scope}[xshift = 9.5cm]
    \node [vertexset, opacity = 0] (V0) {};
    \node [vertexset, opacity = 0, below = 0.5cm of V0] (V1) {};
    
    \node [vertex, marked = 4/5, right = 1cm of V0] (i1) {$\frac{4}{5}$};
    \node [vertex, marked = 1/5, right = 0.5cm of i1] (i2) {$\frac{1}{5}$};

    \node [vertex, opacity = 0] (i) at ($0.5*(i1)+0.5*(i2)$) {};
    \node [vertex, marked = 3/10, below = 0.5cm of i, font=\small] (t2) {$\frac{2}{10}\!\!+\!\!\frac{1}{10}$};
    \node [vertex, marked = 3/5, left = 0.5cm of t2] (t1) {$\frac{6}{10}$};
    \node [vertex, marked = 1/10, right = 0.5cm of t2] (t3) {$\frac{1}{10}$};

    \drawLink[marked = 1/5]{i1}{t1}{3};
    \drawLink[marked = 1/5]{i1}{t2}{1};
    \drawLink[marked = 1/10]{i2}{t2}{1};
    \drawLink[marked = 1/10]{i2}{t3}{1};

    \node [above = 0.25cm of i] {$\Pr(X_0) = \left( \frac{4}{5}, \frac{1}{5} \right)$};
    \node [below = 0.25cm of t2] {$\Pr(X_1) = \left( \frac{6}{10}, \frac{3}{10}, \frac{1}{10} \right)$};
    \node [below = 1cm of t2] {$\D{1} \left( X_0 \xrightarrow{E_1} X_1 \right) \approx 2.45$};
  \end{scope}

\end{tikzpicture}

%%% Local Variables:
%%% mode: latex
%%% TeX-master: "../main"
%%% End:
  \caption{Computation of the collective $V_1$ diversity of $V_0$ along a simple \metapath{} made of only one edge type. Diversity along a path type depends on the starting probability distribution $\Pr(X_0)$ and on transition probabilities.}
  \label{fig:collective}
\end{figure}
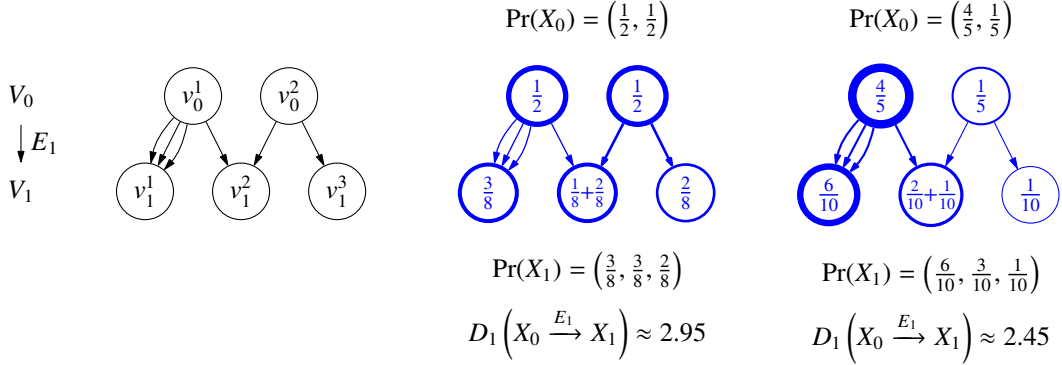

Note that this measure depends on the starting probability distribution $\Pr(X_0 = v_0)$ and on transition probabilities $p_{E_{r_i}}(v_i \given v_{i-1})$ for each $E_{r_i} \in \Pi$. Figure~\ref{fig:collective} provides an example of the measurement of collective diversity for a simple heterogeneous information network containing 5 vertices (represented as circles) and 6 edges (represented as arrows between circles), and using two different starting probability distributions $\Pr(X_0)$. 
In Figure~\ref{fig:collective}, vertices are organized into two vertex types $V_0 = \{v_0^1,v_0^2\}$ and $V_1 = \{v_1^1,v_1^2,v_1^3\}$ (represented as two horizontal layers) and edges are organized into a unique edge type $E_1$, going from $V_0$ to $V_1$.
Two examples of measurements are illustrated in blue for two different starting distributions (numbers within circles give the probabilities of the random walker's position during the different steps of the walk).

The choice of $X_0\sim\text{Uniform}(\Vstart)$ has a central role in many applications.
By giving each node in $\Vstart$ an equal chance of being the random walk's starting point, the resulting collective diversity will be that of the collective --equal-- contribution of all nodes in $\Vstart$.
Similarly, considering subset $\Vstart'\subset\Vstart$ and choosing $X_0\sim\text{Uniform}(\Vstart')$ allows us to define the collective diversity of the group of nodes $\Vstart'$.

\begin{figure}[!h]
  \centering
  \begin{center}
  \begin{tikzpicture}
    \node [vertexset] (V0) {$V_0$};
    \node [vertexset, below = 0.5cm of V0] (V1) {$V_1$};
    \draw [->] (V0) -- node [right] {$E_1$} (V1);

    \begin{scope}[xshift = 0.5cm]
      \node [vertexset, opacity = 0] (V0) {};
      \node [vertexset, opacity = 0, below = 0.5cm of V0] (V1) {};
      
      \node [vertex, right = 1cm of V0] (i1) {$v_0^1$};
      \node [vertex, right = 0.5cm of i1] (i2) {$v_0^2$};

      \node [vertex, opacity = 0] (i) at ($0.5*(i1)+0.5*(i2)$) {};    
      \node [vertex, below = 0.5cm of i] (t2) {$v_1^2$};
      \node [vertex, left = 0.5cm of t2] (t1) {$v_1^1$};
      \node [vertex, right = 0.5cm of t2] (t3) {$v_1^3$};

      \drawLink[]{i1}{t1}{3};
      \drawLink[]{i1}{t2}{1};
      \drawLink[]{i2}{t2}{1};
      \drawLink[]{i2}{t3}{1};
    \end{scope}

    \begin{scope}[xshift = 5cm]
      \node [vertexset, opacity = 0] (V0) {};
      \node [vertexset, opacity = 0, below = 0.5cm of V0] (V1) {};
      
      \node [vertex, marked = 1, right = 1cm of V0] (i1) {$1$};
      \node [vertex, right = 0.5cm of i1] (i2) {};

      \node [vertex, opacity = 0] (i) at ($0.5*(i1)+0.5*(i2)$) {};
      \node [vertex, marked = 1/4, below = 0.5cm of i] (t2) {$\frac{1}{4}$};
      \node [vertex, marked = 3/4, left = 0.5cm of t2] (t1) {$\frac{3}{4}$};
      \node [vertex, right = 0.5cm of t2] (t3) {0};

      \drawLink[marked = 1/4]{i1}{t1}{3};
      \drawLink[marked = 1/4]{i1}{t2}{1};
      \drawLink[]{i2}{t2}{1};
      \drawLink[]{i2}{t3}{1};

      \node [above = 0.25cm of i] {$X_0 = v_0^1$};
      \node [below = 0.25cm of t2] {$\Pr(X_1 \given X_0 = v_0^1) = \left( \frac{3}{4}, \frac{1}{4}, 0 \right)$};
      \node [below = 1cm of t2] {$\D{1} \left( X_0 \xrightarrow{E_1} X_1 \given X_0 = v_0^1 \right) \approx 1.75$};
    \end{scope}

    \begin{scope}[xshift = 9.5cm]
      \node [vertexset, opacity = 0] (V0) {};
      \node [vertexset, opacity = 0, below = 0.5cm of V0] (V1) {};
      
      \node [vertex,right = 1cm of V0] (i1) {};
      \node [vertex, marked = 1, right = 0.5cm of i1] (i2) {1};

      \node [vertex, opacity = 0] (i) at ($0.5*(i1)+0.5*(i2)$) {};
      \node [vertex, marked = 1/2, below = 0.5cm of i] (t2) {$\frac{1}{2}$};
      \node [vertex, left = 0.5cm of t2] (t1) {0};
      \node [vertex, marked = 1/2, right = 0.5cm of t2] (t3) {$\frac{1}{2}$};

      \drawLink[]{i1}{t1}{3};
      \drawLink[]{i1}{t2}{1};
      \drawLink[marked = 1/2]{i2}{t2}{1};
      \drawLink[marked = 1/2]{i2}{t3}{1};

      \node [above = 0.25cm of i] {$X_0 = v_0^2$};
      \node [below = 0.25cm of t2] {$\Pr(X_1 \given X_0 = v_0^2) = \left( 0, \frac{1}{2}, \frac{1}{2} \right)$};
      \node [below = 1cm of t2] {$\D{1} \left( X_0 \xrightarrow{E_1} X_1 \given X_0 = v_0^2 \right) = 2$};
    \end{scope}

  \end{tikzpicture}
\end{center}

$\Pr(X_0) = \left( \frac{1}{2}, \frac{1}{2} \right) \qquad \Rightarrow \qquad \D{1} \left( X_0 \xrightarrow{E_1} X_1 \given X_0 \right) \quad \approx \quad (1.75)^{\frac{1}{2}} \; (2)^{\frac{1}{2}} \quad \approx \quad 1.87$\\

$\Pr(X_0) = \left( \frac{4}{5}, \frac{1}{5} \right) \qquad \Rightarrow \qquad \D{1} \left( X_0 \xrightarrow{E_1} X_1 \given X_0 \right) \quad \approx \quad (1.75)^{\frac{4}{5}} \; (2)^{\frac{1}{5}} \quad \approx \quad 1.80$

%%% Local Variables:
%%% mode: latex
%%% TeX-master: "../main"
%%% End:  
  \caption{Examples of a heterogeneous information network (top left), the individual diversities of two vertices (top center and top right), and the mean individual diversities for two different starting probability distributions (bottom).}
  \label{fig:individual}
\end{figure}
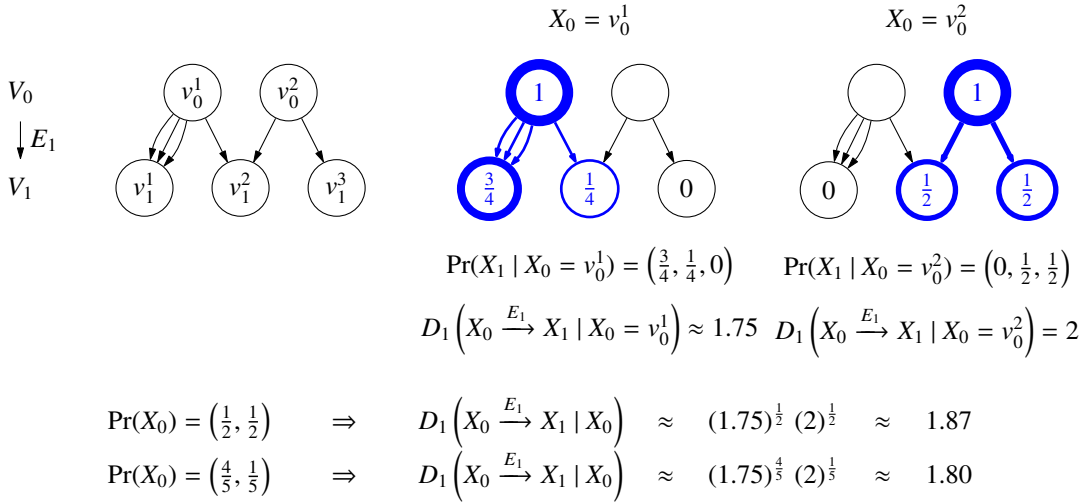

\emph{Conditioned} probabilities of random walks along a path $\Pi$ are also of interest, as they convey information about the structure of the network reachable from some vertices in $\Vstart$.
In particular, given a starting vertex $v_0 \in \Vstart$, we define the \emph{individual $\Vend$ diversity of $v_0$ along $\Pi$} as the true diversity of the probability distribution of $X_k\in \Vend$ at the end of the constrained random walk, knowing that it started at a given vertex $v_0\in \Vstart$ (\ie, $X_0 = v_0$). Figure~\ref{fig:individual} illustrates the measurement of individual diversity for two different vertices in $\Vstart$ in the case of a simple heterogeneous information network.

\begin{definition}[Individual diversity]
\label{def:individual_diversity}
Given a starting vertex $v_0 \in \Vstart$, we define the \emph{individual $\Vend$ diversity of $v_0$ along $\Pi$} as the true diversity of the probability distribution of the ending vertex of the constrained random walk. We denote it as $\D{\alpha} \left( X_0 \xrightarrow{\Pi} X_k  \given X_0 = v_0 \right)$ and compute it as follows:
$$\D{\alpha} \left( X_0 \xrightarrow{\Pi} X_k  \given X_0 = v_0 \right) \quad = \quad \D{\alpha}(p_{\Pi|v_0}) \text{.}$$
\end{definition}

An aggregation of individual diversities may be computed to represent the mean diversity of all (or many) of the vertices in the starting vertex type $\Vstart$.
Following the definition of conditional entropy in information theory (cf. Section 2.2 in \cite{Cover2006}), we define the \emph{mean $\Vend$ individual diversity of $\Vstart$ along $\Pi$} as the weighted geometric mean of individual diversities.

\begin{definition}[Mean individual diversity]
\label{def:mean_individual_diversity}
Given a starting vertex type $\Vstart$, we define the \emph{mean individual $\Vend$ diversity of $\Vstart$ along $\Pi$} as the weighted geometric mean of individual diversities.
We denote it by $\D{\alpha} \left( X_0 \xrightarrow{\Pi} X_k  \given X_0 \right)$ and compute it as follows:
$$
\D{\alpha} \left( X_0 \xrightarrow{\Pi} X_k  \given X_0 \right) \quad = \quad \prod\limits_{v_0\in \Vstart}\D{\alpha}(p_{\Pi|v_0})^{\Pr(X_0=v_0)} \text{}
$$
\end{definition}

This mean is weighted by --and so depends on-- the starting probability distribution $\Pr(X_0)$ over $V_0$, and it is minimal (\ie, equal to 1) when each individual diversity is minimal.
Mean individual diversity is a weighted geometric mean in the general case (\ie, for any distribution for $X_0$), and a --unweighted-- geometric mean when all vertices in $\Vstart$ have the same probability of being the starting point of the random walk (\ie, when $X_0\sim\text{Uniform}(\Vstart)$).
As with collective diversity, the mean individual diversity of a vertex group $\Vstart'\subset\Vstart$ can be considered choosing $X_0\sim\text{Uniform}(\Vstart')$.
Figure~\ref{fig:individual} illustrates the computation of individual and mean individual diversities in a simple heterogeneous information network.

\begin{figure}[!h]
  \centering
  \begin{tikzpicture}[scale=0.85]

  \begin{scope}
    \node [vertex] (i1) {};
    \node [vertex, right = 0.5cm of i1] (i2) {};
    \node [vertex, right = 0.5cm of i2] (i3) {};
    \node [vertex, right = 0.5cm of i3] (i4) {};

    \node [vertex, below = 1cm of i1] (t1) {};
    \node [vertex, below = 1cm of i2] (t2) {};
    \node [vertex, below = 1cm of i3] (t3) {};
    \node [vertex, below = 1cm of i4] (t4) {};

    \drawLink{i1}{t2}{2};
    \drawLink{i2}{t2}{2};
    \drawLink{i3}{t3}{2};
    \drawLink{i4}{t3}{2};
  \end{scope}

  \begin{scope}[xshift = 6cm]
    \node [vertex] (i1) {};
    \node [vertex, right = 0.5cm of i1] (i2) {};
    \node [vertex, right = 0.5cm of i2] (i3) {};
    \node [vertex, right = 0.5cm of i3] (i4) {};

    \node [vertex, below = 1cm of i1] (t1) {};
    \node [vertex, below = 1cm of i2] (t2) {};
    \node [vertex, below = 1cm of i3] (t3) {};
    \node [vertex, below = 1cm of i4] (t4) {};

    \drawLink{i1}{t1}{2};
    \drawLink{i2}{t2}{2};
    \drawLink{i3}{t3}{2};
    \drawLink{i4}{t4}{2};

    \draw [thick, decorate, decoration = {brace, amplitude = 10pt}] (5,0.5) -- (5,-2.5)
    node [midway, xshift = 1cm, rotate = -90, align = center] (ind1) {Mean of individual\\ diversities = 1};
    % $\D{1} \left( X_0 \xrightarrow{E_1} X_1 \given X_0 \right) = 1$
\end{scope}

  \begin{scope}[yshift = -4cm]
    \node [vertex] (i1) {};
    \node [vertex, right = 0.5cm of i1] (i2) {};
    \node [vertex, right = 0.5cm of i2] (i3) {};
    \node [vertex, right = 0.5cm of i3] (i4) {};

    \node [vertex, below = 1cm of i1] (t1) {};
    \node [vertex, below = 1cm of i2] (t2) {};
    \node [vertex, below = 1cm of i3] (t3) {};
    \node [vertex, below = 1cm of i4] (t4) {};

    \drawLink{i1}{t2}{1};
    \drawLink{i1}{t3}{1};
    \drawLink{i2}{t2}{1};
    \drawLink{i2}{t3}{1};
    \drawLink{i3}{t2}{1};
    \drawLink{i3}{t3}{1};
    \drawLink{i4}{t2}{1};
    \drawLink{i4}{t3}{1};

    \draw [thick, decorate, decoration = {brace, amplitude = 10pt}] (4.5,-3) -- (-0.5,-3)
    node [midway, yshift = -0.75cm, align = center] (coll1) {Collective diversity = 2};
  \end{scope}

  \begin{scope}[xshift = 6cm, yshift = -4cm]
    \node [vertex] (i1) {};
    \node [vertex, right = 0.5cm of i1] (i2) {};
    \node [vertex, right = 0.5cm of i2] (i3) {};
    \node [vertex, right = 0.5cm of i3] (i4) {};

    \node [vertex, below = 1cm of i1] (t1) {};
    \node [vertex, below = 1cm of i2] (t2) {};
    \node [vertex, below = 1cm of i3] (t3) {};
    \node [vertex, below = 1cm of i4] (t4) {};

    \drawLink{i1}{t1}{1};
    \drawLink{i1}{t2}{1};
    \drawLink{i2}{t1}{1};
    \drawLink{i2}{t3}{1};
    \drawLink{i3}{t2}{1};
    \drawLink{i3}{t4}{1};
    \drawLink{i4}{t3}{1};
    \drawLink{i4}{t4}{1};

    \draw [thick, decorate, decoration = {brace, amplitude = 10pt}] (4.5,-3) -- (-0.5,-3)
    node [midway, yshift = -0.75cm, align = center] (coll2) {Collective diversity = 4};

    \draw [thick, decorate, decoration = {brace, amplitude = 10pt}] (5,0.5) -- (5,-2.5)
    node [midway, xshift = 1cm, rotate = -90, align = center] (ind2) {Mean of individual\\ diversities = 2};
  \end{scope}

  \node [font=\huge] at ($0.5*(coll1)+0.5*(coll2)$) {$<$};
  \node [rotate = -90, font=\huge] at ($0.5*(ind1)+0.5*(ind2)$) {$<$};

\end{tikzpicture}

%%% Local Variables:
%%% mode: latex
%%% TeX-master: "../main"
%%% End:
  \caption{Different heterogeneous information networks with two vertex types, illustrating different relative ordering for collective and mean individual diversities.}
  \label{fig:individual_vs_collective}
\end{figure}
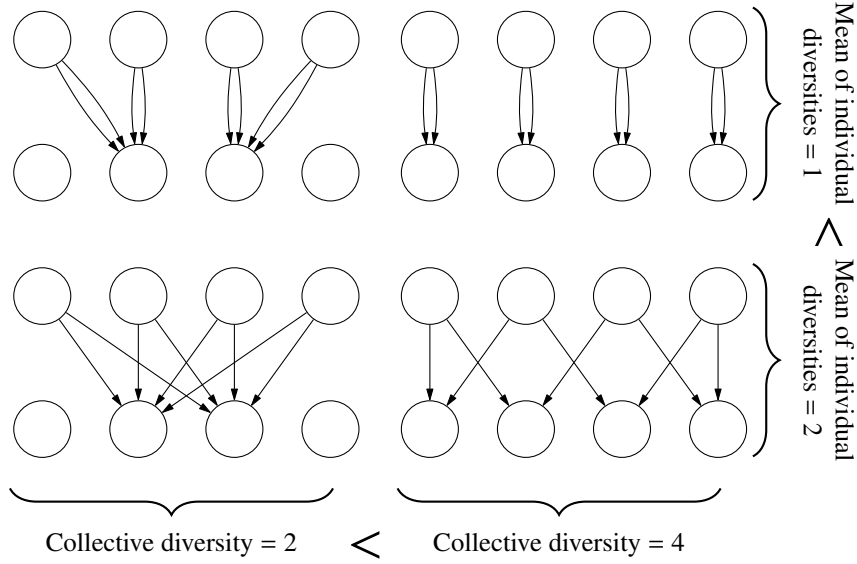

Individual and collective diversities are two complementary measures describing different properties of the system, as illustrated in Figure~\ref{fig:individual_vs_collective}. It is possible for a system to have a low mean individual diversity while having a high collective diversity (top-right in Figure~\ref{fig:individual_vs_collective}), or a high mean individual diversity while having a low collective diversity (bottom-left in Figure~\ref{fig:individual_vs_collective}).

%

%%%%%%%%%%%%%%%%%%%%%%%%%%%%%%%%%%%%%%%%%%%%%%%%%%%%%%%%%%%%%%
\subsection{Backward diversity}
\label{subsec:backward_diversity}

Backward diversity is related to random walks following directions and edges opposite to those of a given \metapath{}. In order to treat them formally, we first present the following definitions.

\begin{definition}[Transpose edge type]
\label{def:transpose_edge_type}
Let $E\in\calE$ be an edge type. We denote by $E^\intercal$ the set of edges resulting from inverting those of $E$:
$$
E^\intercal = \left\{ (\dv{e},\sv{e})\in \V\times\V: e\in E \right\}.
$$
\end{definition}

\begin{definition}[Transpose \metapath{}]
\label{def:transpose_metapath}
For a \metapath{} $\Pi = (E_{r_1}, \ldots, E_{r_k})$, we define its transpose \metapath{} $\Pi^\intercal$ as
$$
\Pi^\intercal = (E^\intercal_{r_k}, \ldots, E^\intercal_{r_1}).
$$
\end{definition}

Using random walks along a \metapath{} $\Pi$, we can also compute the probability distribution of the random walk's starting vertex $X_0\in \Vstart$ when its ending vertex $v_k\in \Vend$ is known.
True diversity of this distribution, called \emph{backward $\Vstart$ diversity of $v_k\in\Vend$ along $\Pi$}, provides a value for the diversity of starting points that can reach $v_k$ following $\Pi$.

\begin{definition}[Backward diversity]
\label{def:backward_diversity}
Given an ending vertex $v_k \in \Vend$, we define the \emph{individual backward $\Vstart$ diversity of $v_k$ along $\Pi$} as the true diversity of the distribution of starting vertex $X_0\in\Vstart$. We denote it by $\D{\alpha} \left( X_0 \given X_0 \xrightarrow{\Pi} X_k  = v_k \right)$ and compute it as follows:
$$\D{\alpha} \left( X_0 \given X_0 \xrightarrow{\Pi} X_k  = v_k \right) \; = \; \D{\alpha} \left( p_{\Pi^\intercal | v_k} \right).
$$

We denote by $\D{\alpha} \left( X_0 \given X_0 \xrightarrow{\Pi} X_k \right)$ the mean backward diversity and compute it as follows:
$$
\D{\alpha} \left( X_0 \given X_0 \xrightarrow{\Pi} X_k \right) \; = \; \prod\limits_{v_k\in\Vend}\D{\alpha} \left( p_{\Pi^\intercal | v_k} \right)^{\Pr(X_k=v_k)}.
$$
\end{definition}

%%%%%%%%%%%%%%%%%%%%%%%%%%%%%%%%%%%%%%%%%%
\subsection{Relative diversity}
\label{subsec:relative_diversity}

Once the notions of collective and individual diversities have been identified, it is natural to consider the diversity of an individual vertex relative to collective diversity.

\begin{definition}[Relative individual diversity]
\label{def:relative_individual_diversity}
We define the \emph{relative individual $\Vend$ diversity of $v_0\in \Vstart$ with respect to $\Vstart$ along $\Pi$} as the relative true diversity between the distribution resulting from a random walk starting at $v_0\in \Vstart$ (giving its individual diversity), and the distribution resulting from the unconditional random walk starting at random in $\Vstart$ (giving the collective diversity).
 We denote it by $\D{\alpha} \left( X_0 \xrightarrow{\Pi} X_k \given X_0 = v_0 \relativeto X_0 \xrightarrow{\Pi} X_k \right)$ and compute it as follows:
$$
\D{\alpha} \left( X_0 \xrightarrow{\Pi} X_k \given X_0 = v_0 \relativeto X_0 \xrightarrow{\Pi} X_k \right)=\D{\alpha}\left( p_{\Pi|v_0} \relativeto p_{\Pi} \right)\text{.}
$$
\end{definition}

Using relative true diversities from Section~\ref{subsec:relative_true_diversities}, other relative network diversities can be computed.
Let us consider for example two different \metapath{}s $\Pi_1$ and $\Pi_2$, such that $\Vstart=\sV{\Pi_1}=\sV{\Pi_2}$, and $\Vend=\dV{\Pi_1}=\dV{\Pi_2}$.
One diversity measure of interest when comparing diversities is the relative true diversity between distributions on $\Vend$ resulting from following different \metapath{}s:
$$
\D{\alpha}\left( X_0\xrightarrow{\Pi_1}X_k \relativeto X_0\xrightarrow{\Pi_2}X_k \right) = \D{\alpha}\left( p_{\Pi_1}\relativeto p_{\Pi_2}\right).
$$

Relative diversities (to be illustrated in Section~\ref{sec:part5}) are useful whenever we want to compare the diversity related to some \metapath{} with a baseline resulting with another one.
Similarly, though not developed in this article, these computations could be extended to the relative mean individual diversity, and backward diversity.

%%%%%%%%%%%%%%%%%%%%%%%%%%%%%%%%%%%%%%%%%%%%%%%%%%%%%%%%%%%
\subsection{Projected diversity}
\label{subsec:projected_diversity}

Using projected edges $E_{\Pi}$ of a \metapath{} $\Pi$  (cf. Definition~\ref{def:metapath_projection} in Section~\ref{subsec:restricted_random_walks}), we can also define the diversity of the distribution on $\Vend$ of a constrained random walk starting at $v_0\in \Vstart$ and following the edges in $E_{\Pi}$.

\begin{definition}[Projected diversity]
\label{def:projected_diversity}
Let $E_\Pi$ be the set of projected edges of \metapath{} $\Pi$.
We define the \emph{projected $\Vend$ diversity of $v_0\in \Vstart$ along $\Pi$} as the true diversity of the distribution of the ending vertices in $\Vend$ of a random walk starting at $v_0\in \Vstart$ and following the edges in $E_{\Pi}$. We denote it by $\D{\alpha} \left( X_0 \xrightarrow{E_{\Pi}} X_k \given X_0 = v_0 \right)$ and compute it as follows:
$$
\D{\alpha} \left( X_0 \xrightarrow{E_{\Pi}} X_k \given X_0 = v_0 \right) = \D{\alpha}\left(p_{E_\Pi|v_0} \right)\text{.}
$$
\end{definition}

\begin{figure}[!h]
  \centering
  % \begin{minipage}[t]{0.45\linewidth}
\centering

\begin{tikzpicture}
  \node [vertexset] (V0) {$V_0$};
  \node [vertexset, above = 0.5cm of V0] (V1) {$V_1$}; 
  \node [vertexset, above = 0.5cm of V1] (V2) {$V_2$};
  
  \node [vertex, marked = 1, right = 2cm of V0] (u1) {$1$};

  \node [vertex, white, above = 0.5cm of u1] (i) {};
  \node [vertex, marked = 1/2, left = 0.25cm of i] (i1) {$\frac{1}{2}$};
  \node [vertex, marked = 1/2, right = 0.25cm of i] (i2) {$\frac{1}{2}$};

  \node [vertex, marked = 1/2, above = 0.5cm of i1] (t1) {$\frac{1}{2}$};
  \node [vertex, marked = 1/6, above = 0.5cm of i2] (t3) {$\frac{1}{6}$};
  \node [vertex, marked = 1/6, left = 0.10cm of t3] (t2) {$\frac{1}{6}$};
  \node [vertex, marked = 1/6, right = 0.10cm of t3] (t4) {$\frac{1}{6}$};

  \drawLink[marked = 1/2]{u1}{i1}{1};
  \drawLink[marked = 1/2]{u1}{i2}{1};
  \drawLink[marked = 1/2]{i1}{t1}{1};
  \drawLink[marked = 1/6]{i2}{t2}{1};
  \drawLink[marked = 1/6]{i2}{t3}{1};
  \drawLink[marked = 1/6]{i2}{t4}{1};

  \node [vertex, marked = 1, right = 7cm of V0] (u1b) {$1$};

  \node [vertex, white, above = 0.5cm of u1b] (ib) {};
  \node [vertex, white, left = 0.25cm of ib] (i1b) {};
  \node [vertex, white, right = 0.25cm of ib] (i2b) {};

  \node [vertex, marked = 1/4, above = 0.5cm of i1b] (t1b) {$\frac{1}{4}$};
  \node [vertex, marked = 1/4, above = 0.5cm of i2b] (t3b) {$\frac{1}{4}$};
  \node [vertex, marked = 1/4, left = 0.10cm of t3b] (t2b) {$\frac{1}{4}$};
  \node [vertex, marked = 1/4, right = 0.10cm of t3b] (t4b) {$\frac{1}{4}$};

  \drawLink[marked = 1/4]{u1b}{t1b}{1};
  \drawLink[marked = 1/4]{u1b}{t2b}{1};
  \drawLink[marked = 1/4]{u1b}{t3b}{1};
  \drawLink[marked = 1/4]{u1b}{t4b}{1};

  \node [below = 0.2cm of u1] (seq) {Individual diversity $\approx$ 3.46};
  \node [below = 0.2cm of u1b] (proj) {Projected diversity = 4 };
  \node [font=\huge] at ($0.5*(seq)+0.5*(proj)$) {$<$};
\end{tikzpicture}

\vspace{2em}

\begin{tikzpicture}
  \node [vertexset] (V0) {$V_0$};
  \node [vertexset, above = 0.5cm of V0] (V1) {$V_1$}; 
  \node [vertexset, above = 0.5cm of V1] (V2) {$V_2$};
  
  \node [vertex, marked = 1, right = 2cm of V0] (u1) {$1$};

  \node [vertex, white, above = 0.5cm of u1] (i) {};
  \node [vertex, marked = 1/2, left = 0.25cm of i] (i1) {$\frac{1}{2}$};
  \node [vertex, marked = 1/2, right = 0.25cm of i] (i2) {$\frac{1}{2}$};

  \node [vertex, marked = 1/2, above = 0.5cm of i1] (t1) {$\frac{1}{2}$};
  \node [vertex, marked = 1/2, above = 0.5cm of i2] (t3) {$\frac{1}{2}$};

  \drawLink[marked = 1/2]{u1}{i1}{1};
  \drawLink[marked = 1/2]{u1}{i2}{1};
  \drawLink[marked = 1/2]{i1}{t1}{1};
  \drawLink[marked = 1/6]{i2}{t3}{3};

  \node [vertex, marked = 1, right = 7cm of V0] (u1b) {$1$};

  \node [vertex, white, above = 0.5cm of u1b] (ib) {};
  \node [vertex, white, left = 0.25cm of ib] (i1b) {};
  \node [vertex, white, right = 0.25cm of ib] (i2b) {};

  \node [vertex, marked = 1/4, above = 0.5cm of i1b] (t1b) {$\frac{1}{4}$};
  \node [vertex, marked = 3/4, above = 0.5cm of i2b] (t3b) {$\frac{3}{4}$};

  \drawLink[marked = 1/4]{u1b}{t1b}{1};
  \drawLink[marked = 1/4]{u1b}{t3b}{3};

  \node [below = 0.5cm of u1] (seq) {Individual diversity = 2};
  \node [below = 0.5cm of u1b] (proj) {Projected diversity $\approx$ 1.75};
  \node [font=\huge] at ($0.5*(seq)+0.5*(proj)$) {$>$};
\end{tikzpicture}

%%% Local Variables:
%%% mode: latex
%%% TeX-master: "../main"
%%% End:
  \caption{Individual Shannon diversities of a \metapath{} on two heterogeneous information networks, compared with the resulting projected diversities. We illustrate two situations: one in which projected diversity is greater than individual diversity (top), and one where individual diversity is greater than projected diversity (bottom). }
  \label{fig:sequential_vs_projected}
\end{figure}
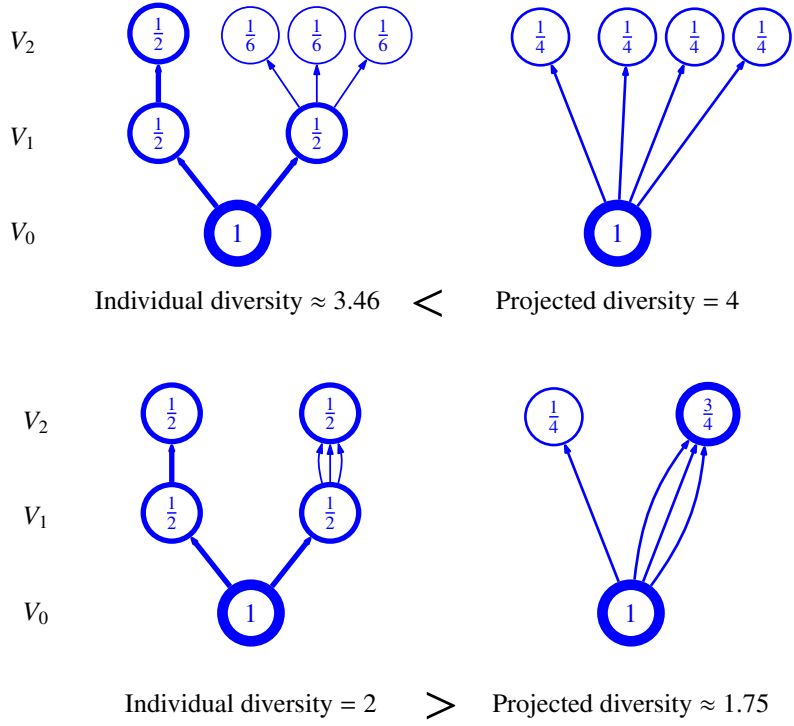

Note that in the previous definition, $p_{E_\Pi|v_0}$ is the probability distribution from Definition~\ref{def:conditional_probability_random_walk} when the \metapath{} is made only of projected edges in $E_\Pi$.
Figure~\ref{fig:sequential_vs_projected} illustrates the comparison between individual and projected diversities for two cases using Shannon diversity.
One of these cases results in a projected diversity that is lower than individual diversity, while the other results in a projected diversity that is higher than individual diversity.

%%%%%%%%%%%%%%%%%%%%%%%%%%%%%%%%%%%%%%%%%%%%%%%%%%%%%%%%%%%%%%
\subsection{The relation between network diversity measures}
\label{subsec:relations_between_diversities}

The network diversity measures presented here are not independent.
In this section we show a relation involving collective, backward, and mean individual diversities.
In order to do so, we first need to consider \emph{parts of \metapath{}s}.

\begin{definition}[Parts of \metapath{}s]
\label{not:parts_of_metapaths}
Given a \metapath{} $\Pi=(E_{r_1},\ldots,E_{r_k})$ of length $k$, we denote by $\Pi_{(i,j)}$, for $1\leq i \leq j \leq k$, its restriction 
$$
\Pi_{(i,j)} = \left(E_{r_i},E_{r_{i+1}},\ldots E_{r_{j-1}}, E_{r_j} \right).
$$
\end{definition}

\begin{theorem}[Bound for collective Shannon diversity]
\label{thm:bound_collective}
The following inequality holds for Shannon diversity, that is 1-order true diversity,
$$\D{1}(X_0 \xrightarrow{\Pi} X_k) \quad \leq \quad \D{1}(X_0 \xrightarrow{\subPi{1}{i}} X_i) \quad \D{1}(X_i \xrightarrow{\subPi{i+1}{k}} X_k \given X_0\xrightarrow{\subPi{1}{i}} X_i) \text{,}$$
with equality if and only if $\D{1} (X_0 \xrightarrow{\subPi{1}{i}} X_i \given X_i \xrightarrow{\subPi{i+1}{k}} X_k) = 1$.
\end{theorem}

In other words, collective diversity along a \metapath{} is bounded by two factors: 
(1)~collective diversity at any step of the \metapath{},
multiplied by (2)~mean individual diversity along the remaining part of the \metapath{}.
This bound is achieved if an only if all individual backward diversities along this remaining part are minimal.

Before proving Theorem~\ref{thm:bound_collective}, let us first prove the following result linking collective and individual 1-order true diversities for a single edge type.

\begin{lemma}[The relation between collective, backward, and mean individual diversities]
\label{lemma:collective_meanindividual_single_edgetype}
Let us consider an edge type $E$, with $\sV{E}=V_0$ and $\dV{E}=V_1$, and the constrained random walk $X_0 \xrightarrow{E}X_1$, with $X_0\in V_0$ and $X_1\in V_1$. The following identity relation between collective, backward, and mean individual 1-order true diversities holds:
$$
\underbrace{\D{1}\left(X_0\xrightarrow{E} X_1\right)}_{\text{collective div.}} \quad \underbrace{\D{1}\left(X_0\given X_0\xrightarrow{E} X_1\right)}_{\text{mean backward div.}} \quad = \quad \underbrace{\D{1}\left(\Pr(X_0)\right)}_{\text{initial div.}} \quad \underbrace{\D{1}\left(X_0\xrightarrow{E}X_1 \given X_0\right)}_{\text{mean individual div.}},
$$
\noindent where $\D{1}\left(\Pr(X_0)\right)$, the \emph{initial diversity}, is the 1-order true diversity of the distribution for the starting vertex of the random walk.
\end{lemma}

\begin{pf}
Let us consider the 1-order true diversity of the joint probability $\Pr(X_0,X_1)$ of the starting vertex in $V_0$ and the ending vertex in $V_1$. Despite $X_0$ and $X_1$ being dependent, by the principle of strong additivity of 1-order true diversity (cf. Theorem~\ref{thm:true_diversities_are_strongly_additive}), we have

\begin{align*}
\D{1}\left(\Pr (X_0,X_1) \right) & \overset{\text{Thm.}~\ref{thm:true_diversities_are_strongly_additive}}{=} \D{1}\left(\Pr (X_0)\right) \;  \prod\limits_{v_0\in V_0} \D{1}\left(\Pr (X_1 \given X_0=v_0) \right)^{\Pr(X_0=v_0)}\\
& \overset{\text{Def.}~\ref{def:mean_individual_diversity}}{=} \D{1}\left(\Pr (X_0)\right) \;  \D{1}\left( X_0\xrightarrow{E}X_1\given X_0 \right).
\end{align*}

Also by the principle of strong additivity of 1-order true diversity we have

\begin{align*}
\D{1}\left(\Pr (X_0,X_1) \right) & \overset{\text{Thm.}~\ref{thm:true_diversities_are_strongly_additive}}{=} \D{1}\left(\Pr (X_1)\right) \; \prod\limits_{v_1\in V_1} \D{1}\left(\Pr (X_0 \given X_1=v_1) \right)^{\Pr(X_1=v_1)}\\
& \overset{\text{Def.}~\ref{def:backward_diversity}}{=} \D{1}\left(X_0\xrightarrow{E}X_1\right) \;  \D{1}\left( X_0\given X_0 \xrightarrow{E}X_1\right).\quad\blacksquare
\end{align*}

\end{pf}

Since true diversities are greater or equal to 1 (cf. Theorem~\ref{thm:bounds_for_diversities}), it is clear that 

$$
\underbrace{\D{1}\left(X_0\xrightarrow{E} X_1\right)}_{\text{collective}} \quad \leq \quad \underbrace{\D{1}\left(\Pr(X_0)\right)}_{\text{initial}} \quad \underbrace{\D{1}\left(X_0\xrightarrow{E}X_1 \given X_0\right)}_{\text{mean individual}},
$$
with equality when mean backward diversity is minimal, $\D{1}\left(X_0 \given X_0\xrightarrow{E}X_1\right)=1$. This can only happen when each ending vertex in $V_1$ is reachable from only one starting vertex in $V_0$.

Using the same procedure as in Lemma~\ref{lemma:collective_meanindividual_single_edgetype}, we may now prove Theorem~\ref{thm:bound_collective}. 

\newproof{pot}{Proof of Theorem~\ref{thm:bound_collective}}
\begin{pot}
Given a \metapath{} $\Pi=\left(E_{r_1},\ldots,E_{r_k} \right)$ and a constrained random walk  $X_0\xrightarrow{\Pi}X_k$ along it, let us split it in two parts, dividing our walk in two parts: 
$$
\subPi{1}{i}=\left(E_{r_1},\ldots,E_{r_i} \right),\; \text{ for random walk }X_0\xrightarrow{\subPi{1}{i}}X_i,\quad \text{and}
$$
$$
\subPi{i+1}{k}=\left(E_{r_{i+1}},\ldots,E_{r_k} \right),\; \text{ for random walk }X_i\xrightarrow{\subPi{i+1}{k}}X_k.
$$

Following the same argument than in the proof of Lemma~\ref{lemma:collective_meanindividual_single_edgetype}, we compute the 1-order diversity of distribution $\Pr\left(X_i,X_k\right)$, using the strong additivity principle to obtain two different expressions.

A first application of the strong additivity principle yields
$$
\D{1}\left(\Pr(X_i,X_k)\right) \quad = \quad \D{1}\left(X_0\xrightarrow{\subPi{1}{i}} X_i\right) \quad \D{1}\left(X_i\xrightarrow{\subPi{i+1}{k}}X_k\given X_0\xrightarrow{\subPi{1}{i}} X_i\right),
$$
where $\D{1}\left(X_i\xrightarrow{\subPi{i+1}{k}}X_k\given X_0\xrightarrow{\subPi{1}{i}} X_i\right)$ is the mean individual diversity along meta path $\subPi{i+1}{k}$ using probabilities resulting from random walk $X_0\xrightarrow{\subPi{1}{i}}X_k$ for the weighted geometric mean.

A second application of the strong additivity principle yields
$$
\D{1}\left(\Pr(X_i,X_k)\right) \quad = \quad \D{1}\left(X_i\xrightarrow{\subPi{i+1}{k}} X_k\right) \quad \D{1}\left(X_i\given X_i\xrightarrow{\subPi{i+1}{k}}X_k\right).
$$

Since starting probabilities $\Pr(X_i)$ in collective diversity $\D{1}\left(X_i\xrightarrow{\subPi{i+1}{k}} X_k\right)$ are those resulting from random walk $X_0\xrightarrow{\subPi{1}{i}}X_i$, we also have $\D{1}\left(X_i\xrightarrow{\subPi{i+1}{k}} X_k\right)=\D{1}\left(X_0\xrightarrow{\Pi}X_k\right)$.

This gives the desired result
$$
\D{1}\left(X_0\xrightarrow{\Pi} X_k\right) \quad \D{1}\left(X_i\given X_i\xrightarrow{\subPi{i+1}{k}}X_k\right) \quad = \quad \D{1}\left(X_0\xrightarrow{\subPi{1}{i}} X_i\right) \quad \D{1}\left(X_i\xrightarrow{\subPi{i+1}{k}}X_k\given X_0\xrightarrow{\subPi{1}{i}} X_i\right),
$$
from which it follows that
$$
\D{1}\left(X_0\xrightarrow{\Pi} X_k\right) \quad \leq \quad \D{1}\left(X_0\xrightarrow{\subPi{1}{i}} X_i\right) \quad \D{1}\left(X_i\xrightarrow{\subPi{i+1}{k}}X_k\given X_0\xrightarrow{\subPi{1}{i}} X_i\right)
$$
if mean backward diversity is not equal to 1.\eop

\end{pot}

%%%%%%%%%%%%%%%%%%%%%%%%%%%%%%%%%%%%%%%%%%%%%%%%%%
\subsection{Summary of network diversity measures}
\label{subsec:summary_metapath_diversities}

In this Section~\ref{sec:part4}, we have used the definitions developed within the proposed formalism for heterogeneous information networks, in particular that of \metapath{} constrained random walk, to propose different network diversity measures.
These include collective, individual, backward, relative, and projected diversities along a \metapath{}.
For each one, we have proposed a notation and we have defined a computation using the definitions established in Section~\ref{sec:part3}.
Table~\ref{tab:summary_network_diversities} summarizes the notations and computations of each of the proposed network diversity measures.

\begin{table}
\small
\centering
\caption{Summary of defined diversities along a \metapath{} $\Pi$, with $X_0\in\sV{\Pi}$ and $X_k\in\dV{\Pi}$.}
\label{tab:summary_network_diversities}
\begin{tabular}{|l|l|l|}
\hline
Diversity & Notation & Expression \\
\hline
Collective & $\D{\alpha}\left(X_0 \xrightarrow{\Pi} X_k \right)$ & $\D{\alpha}(p_\Pi)$ \\
\hline
Individual & $\D{\alpha}\left(X_0 \xrightarrow{\Pi} X_k \given X_0=v_0 \right)$ & $\D{\alpha}(p_{\Pi|v_0})$ \\
\hline
Mean individual & $\D{\alpha}\left(X_0 \xrightarrow{\Pi} X_k \given X_0 \right)$ & $\prod\limits_{v_0\in V_0}\D{\alpha}\left(p_{\Pi|v_0} \right)^{\Pr(X_0=v_0)}$ \\
\hline
Relative individual & $\D{\alpha}\left(X_0 \xrightarrow{\Pi} X_k \given X_0 = v_0 \relativeto X_0 \xrightarrow{\Pi} X_k \right)$ & $\D{\alpha}\left( p_{\Pi|v_0} \relativeto p_\Pi \right)$ \\
\hline
Backward individual & $\D{\alpha}\left(X_0 \given X_0 \xrightarrow{\Pi} X_k=v_k \right)$ & $\D{\alpha}\left(p_{\Pi^\intercal|v_k} \right)$ \\
\hline
Projected individual & $\D{\alpha}\left(X_0 \xrightarrow{E_\Pi} X_k \given X_0=v_0 \right)$ & $\D{\alpha}\left(p_{E_\Pi|v_0}\right)$ \\
\hline
\end{tabular}
\end{table}

In the next section, we present different domains of application for which modeling using heterogeneous information networks is useful.
We show that some quantitative measures traditionally computed in different domains are closely related to the network diversity measures we defined, and that their use allows for the consideration of other useful quantitative observables in modeled systems.

\section{Applications}
\label{sec:part5}

The network diversity measures we proposed find numerous applications in many domains where diversity provides relevant information.
Information retrieval, and in particular algorithmic recommendation, is one of the areas with the most direct applications that best illustrates applicability in general.
We first illustrate the use of network diversity measures by means of a simple example from recommender systems in Section~\ref{subsec:simple_example}.
Recommender systems, closely related to information retrieval, intersects with many research topics in artificial intelligence, machine learning, and data mining, and will help us provide illustrative examples for these domains.
The first example introduces notations used in this section, and the approach chosen to illustrate the application of these measures.
This approach consists of considering a particular heterogeneous information network providing an ontology for data in each domain of application, and showing its network schema (cf. Definition~\ref{def:network_schema}).
Then, for each application case in each domain, we list several concepts of interest for research questions that are traditionally relevant in the literature, together with the explicit expression of the corresponding network diversity measures.
These concepts will be referred to previous work where they find pertinence.
We highlight how these proposed measures can address existing research questions and current practices in different research areas, and how they allow for positing new ones. 

After having introduced a simple first example, we provide a numerical example of application of the network diversity measures to real datasets in Section~\ref{subsec:numerical_example}.
In a third example (Section~\ref{subsec:recommender_systems}), we provide a detailed application case in a recommender system setting, explaining the relation between network diversity measures and several existing practices and concepts while also highlighting possible new uses.
We then illustrate the use of network diversity measures for the analysis of social networks and media in Section~\ref{subsec:social_media}.
Finally, we provide other examples of applications in ecology in Section~\ref{subsec:ecology}, antitrust regulation in Section~\ref{subsec:antitrust}, and scientometrics in Section~\ref{subsec:scientometrics}.

%%%%%%%%%%%%%%%%%%%%%%%%%%%%%%%%%%%%%%%%%%%%%%%%%%%%%%%%
\subsection{A simple example}
\label{subsec:simple_example}
  
Let us consider an example heterogeneous information network with three vertex types: users, items, and types of items.
Similar to the notation established in Section~\ref{subsec:notations}, for the sake of readability, let us denote these vertex types respectively by $\susers{V}$, $\sitems{V}$, and $\stypes{V}$.
An example of entities represented by items are films, and an example of types are then film genres (\eg, comedy, thriller).

\begin{figure}[!h]
    \centering
      \begin{center}
    \begin{tikzpicture}
      
      \node [draw, circle, minimum width = 1cm ] (U) {$\susers{V}$};
      \node [draw, circle, minimum width = 1cm , right = 1.5cm of U] (I) {$\sitems{V}$};
      \node [draw, circle, minimum width = 1cm , right = 1.5cm of I] (T) {$\stypes{V}$};

      \draw [->] (U) to[out=30, in=150,edge node={node [midway,above] {$\srec$}}] (I);
      \draw [->] (U) to[out=-30,in=-150,edge node={node [midway,below] {$\schosen$}}] (I);
      \draw [->] (I) to[edge node={node [midway,above] {$\styped$}}] (T);

    \end{tikzpicture}
  \end{center}  
    \caption{Network schema of a simple heterogeneous information network, where users in $\susers{V}$ have chosen and have been recommended items in $\sitems{V}$, which are classified into types in $\stypes{V}$.}
    \label{fig:simple_example}
  \end{figure}
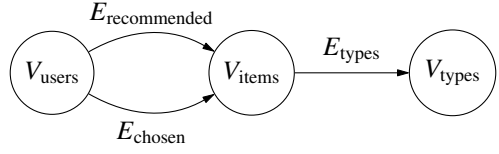

Let us now consider three edge types, indicating items chosen by users, items recommended to users, and classification of items into types.
We respectively denote these edge types as $\schosen$, $\srec$, and $\styped$.
Figure~\ref{fig:simple_example} illustrates the network schema of the described heterogeneous information network.

In order to consider random walks constrained to \metapath{}s in this network, let us denote by the capital letter $X$ random vertices in vertex types.
Thus, for example, $\susers{X}$ is a random vertex in $\susers{V}$, \ie, a random user.
This allows for the consideration of random walks such as 
$$\susers{X}\xrightarrow{\schosen}\sitems{X}\xrightarrow{\styped}\stypes{X},$$
for some starting probability distribution $\Pr(\susers{X})$, and constrained to the \metapath{} $\Pi=(\schosen,\styped)$.
Throughout this section, we denote a random walk by explicitly writing the vertex types and edge types, as in Definition~\ref{def:random_walk}, rather than by its shorter notation $\susers{X}\xrightarrow{\Pi}\stypes{X}$.

Using this notation, we identify some concepts of interest related to diversity, and their corresponding network diversity measures.
Indeed, we might take interest in the collective type diversity of items recommended to users (cf. Definition~\ref{def:collective_diversity}), $$\D{\alpha}\left(\susers{X}\xrightarrow{\srec}\sitems{X}\xrightarrow{\styped}\stypes{X} \right),$$ that quantifies the {type} diversity of items that are recommended to the users.
We might also take interest in the mean individual type diversity of items recommended to users (cf. Definition~\ref{def:mean_individual_diversity}) $$\D{\alpha}\left(\susers{X}\xrightarrow{\srec}\sitems{X}\xrightarrow{\styped}\stypes{X} \given \susers{X} \right),$$
which quantifies the mean of the {type} diversity of items recommended to each user.
Network diversity measures allow for the evaluation, for example, of the collective type diversity of items that are \emph{recommended to} users with respect to items that are \emph{chosen by} users using relative diversity (cf. Definition~\ref{def:relative_individual_diversity}):
$$
\D{\alpha}\left(\susers{X}\xrightarrow{\srec}\sitems{X}\xrightarrow{\styped}\stypes{X}\relativeto 
\susers{X}\xrightarrow{\schosen}\sitems{X}\xrightarrow{\styped}\stypes{X} \right).
$$
Such a measure would reveal how diverse recommendations are (according to item's types) while taking the general landscape of users' consumption as a baseline to measure this diversity. In other words, such a measure would reveal how recommendations may increase or decrease the diversity of what is consumed.
  
The use of transpose edge types (cf. Definition~\ref{def:transpose_edge_type}) allows for the referencing and computing of more complex concepts, such as 
$$\D{\alpha}\left(\susers{X}\xrightarrow{\srec}\sitems{X}\xrightarrow{\schosen^\intercal}\susers{X}'\xrightarrow{\schosen}\sitems{X}'\xrightarrow{\styped}\stypes{X}\given\susers{X}=u\right),$$
which would otherwise be referred to as the \emph{individual type diversity of items chosen by users that chose items recommended to user $u\in\susers{V}$}.
Some random variables are marked with an apostrophe (\eg, $\susers{X}'$) to indicate that, while they have the same support as the unmarked ones ($\text{supp}(\susers{X}')=\text{supp}(\susers{X})=\susers{V}$), they are not the same variable.
This is needed in \metapath{}s that include the same vertex type two or more times.

The following examples of application use this approach: to identify, referentiate, and provide computable expressions for concepts from different domains of research interested in both diversity measures and network representations. 

%%%%%%%%%%%%%%%%%%%%%%%%%%%%%%%%%%%%%%%%%%%%%%%%%%%%%%%%

\subsection{A numerical example}
\label{subsec:numerical_example}
\def\msd{\textsc{ms}\xspace}
\def\amz{\textsc{amazon}\xspace}

We turn now to an empirical example that will show how our network diversity
measures can be used and interpreted in order to analyze a specific
dataset.  
The following study deals with the behaviour
of users on online musical platforms. 
In such platforms, users listen to songs and songs are usually tagged by musical categories.
This situation can be modeled by a heterogeneous information network with three
vertex types (users $\numusers{V}$, songs $\numsongs{V}$, and tags $\numtags{V}$) and two types of edges ($\numconsumed$ connecting $\numusers{V}$ to songs $\numsongs{V}$, and $\numtagged$ connecting $\numsongs{V}$ to tags $\numtags{V}$). 
We use network diversity measures to investigate two different questions.
First we analyze the diversity of the distribution of users that listen to songs tagged with a given tag $t\in \numtags{V}$: the \emph{diversity of the tag audience}.  
Second, we analyze, for a given user $u\in \numusers{V}$, the diversity of the distribution of tags associated with the songs she listens to: the \emph{diversity of a user's attention}.
Translated into our framework, we consider the meta-paths 
$\Pi_{\text{audience}}=( \numtagged^\intercal, \numconsumed^\intercal )$ and
$\Pi_{\text{attention}}=( \numconsumed, \numtagged )$ to analyze the following diversities:

\begin{itemize}
\item $\forall t\in \numtags{V}, \D{\alpha}(p_{\Pi_{\text{audience}}|t})$: the individual diversities of audiences of tags in $\numtags{V}$  (see Section~\ref{sec:divaud});
\item $\forall u\in \numusers{V}, \D{\alpha}(p_{\Pi_{\text{attention}}|u})$: the individual diversities of tags of users in $\numusers{V}$ (see Section~\ref{sec:divatt}).
\end{itemize}

It is worth noting that while the numerical analyses presented in this
section are new, a complete study dedicated to this context and dataset has been
published in~\cite{poultar2020}, which can provide a useful complement
to the reader.

\subsubsection{Datasets used}

In this numerical example we use the same above-described network schema (cf. Definition~\ref{def:network_schema}) for $\numusers{V}$, $\numsongs{V}$, and $\numtags{V}$ to analyze data from two different datasets that can be modeled by this heterogeneous information network.

For the first dataset, we use data from the \emph{Million Song Dataset} project~\cite{msd}.
In particular, we use the \emph{user-taste-profile} data\footnote{Available at \url{https://labrosa.ee.columbia.edu/millionsong/tasteprofile}.}, that
contains 48 million events of users listening to songs, to determine $\numusers{V}$, $\numsongs{V}$, and $\numconsumed$ parts of a heterogeneous information network model.
For the $\numtags{V}$ and $\numtagged$ parts, we use data from the \emph{last.fm}
dataset\footnote{Available at \url{https://labrosa.ee.columbia.edu/millionsong/lastfm}.} that
provides a list of tags for each song. 
Using these two sources of data resulted in a dataset with \np{1 019 190} users in vertex type $\numusers{V}$, \np{234 379} songs in vertex type $\numsongs{V}$, and \np{1 000} tags in vertex type $\numtags{V}$. 
We refer to this dataset as the \msd data.

For the second dataset, we consider a collection of reviews made on
 \emph{Amazon} \cite{amz1, amz2}, and that contain musical items (\eg, CDs, vinyls, and digital music).
From these data, we only retain the link between a user and a product (a
song or an album here).
\emph{Amazon} provides a hierarchy of
categories for each product, which allows us to extract musical tags for each song. 
This resulted in a dataset with \np{465 248} users in vertex type $\numusers{V}$, \np{445 514} songs
in vertex type $\numsongs{V}$, and \np{250} tags in vertex type $\numtags{V}$. 
We will refer to this second dataset as \amz.

Using the specified heterogeneous information network schema to model the data from these two datasets, we may use our network diversity measures to compute diversities in the data.
For ease of analysis, we restrain our analysis in this Section~\ref{subsec:numerical_example} to the 1-order true diversity, \ie, the Shannon diversity (cf. Table~\ref{tab:summary_diversity_measures}).
The reader is refered to~\cite{poultar2020} for a study of these datasets using different diversity measures.

\subsubsection{1-order true diversity of the tag audience}
\label{sec:divaud}

\begin{figure}
  \begin{center}
  \begin{subfigure}[c]{0.4\textwidth}
        \centering 
      \caption{\msd dataset}\label{fig:divaudmsd}
      \includegraphics[width=\columnwidth]{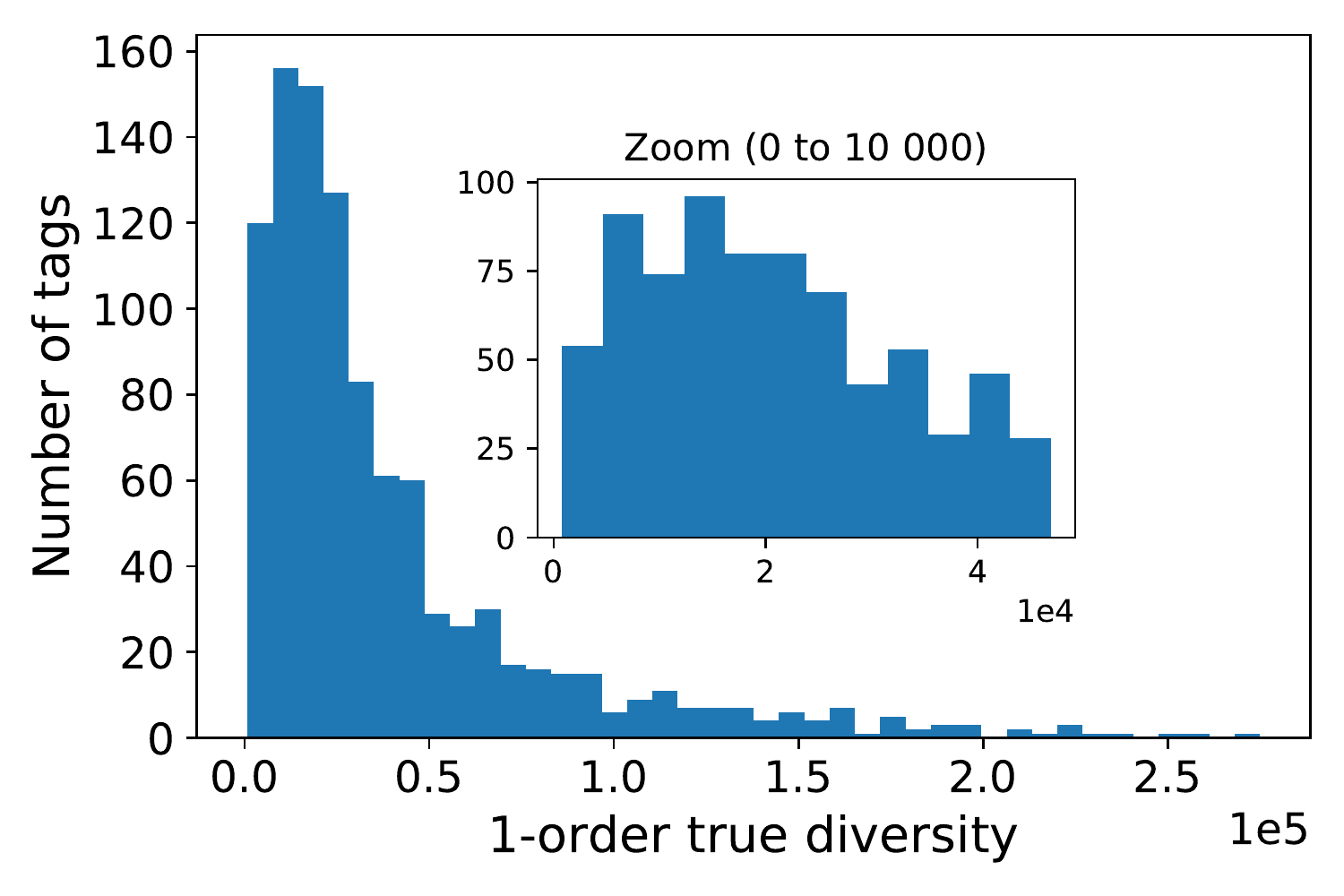}
    \end{subfigure}
    \begin{subfigure}[c]{0.4\textwidth}
        \centering 
      \caption{\amz dataset}\label{fig:divaudam}
      %\raisebox{-5cm}{
      \includegraphics[width=\columnwidth]{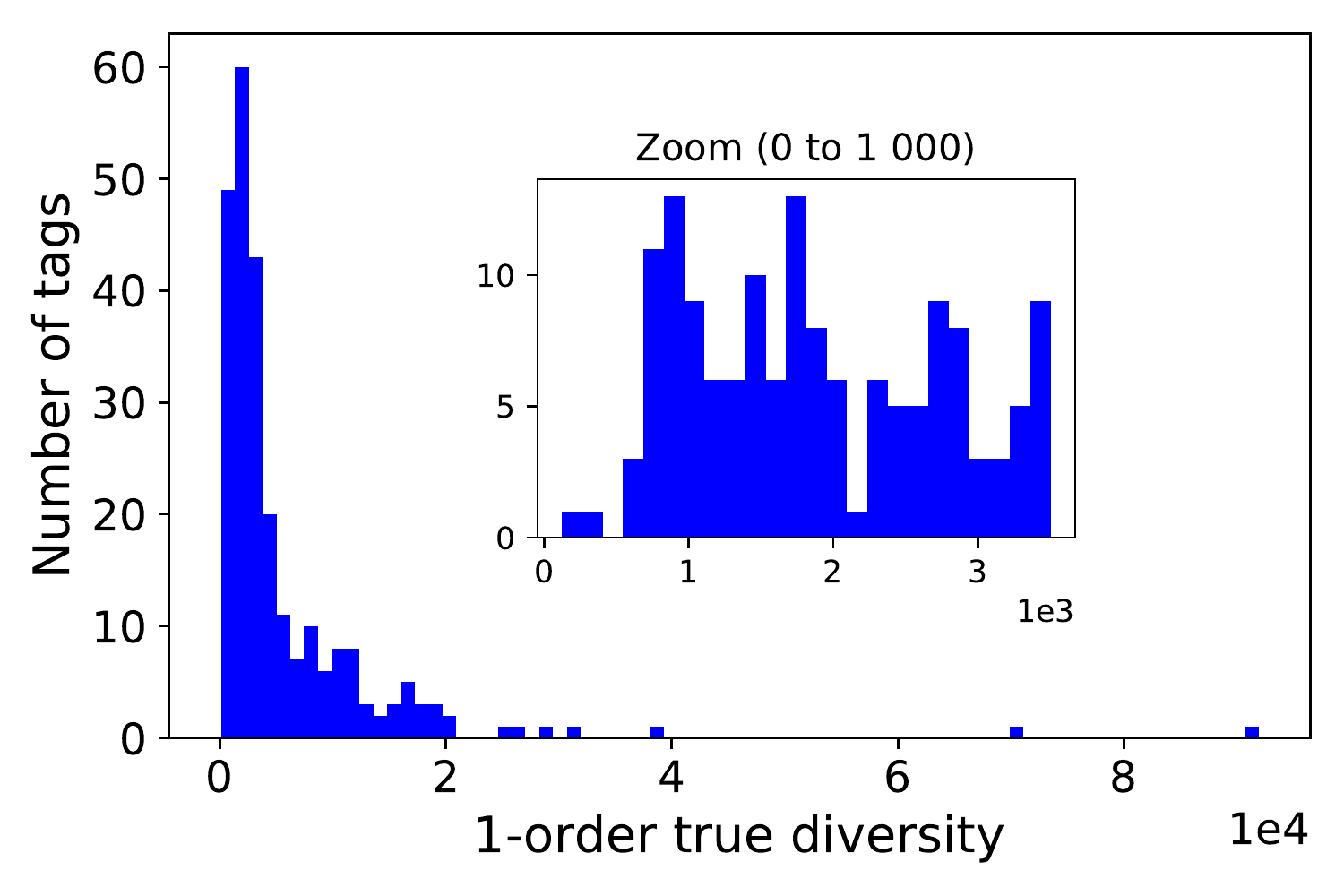}
      %}
    \end{subfigure}
  \end{center}
\caption{Histograms showing the distributions of the 1-order true diversities of the tag audience ($\D{1}(p_{\Pi_{\text{audience}}|t})$ for all tags in $\numtags{V}$) for the \msd (left) and the \amz (right) datasets.}
\label{fig:divaud}
\end{figure}

First we focus on the diversity of the audiences of tags: the individual user-diversities of the tags.
%We want to measure how diverse is the audience of a musical category.
Figure~\ref{fig:divaud} presents the distribution of the 1-order true
diversities $\D{1}(p_{\Pi_{\text{audience}}|t})$ of all tags $t\in\numtags{V}$.
We compute and present these values using the \msd (Figure~\ref{fig:divaudmsd}) and the \amz
(Figure~\ref{fig:divaudam}) datasets.

\begin{figure}
  \begin{subfigure}[c]{0.49\textwidth}
        \caption{\msd dataset}\label{fig:divaud25msd}
        \raisebox{1.1cm}{\includegraphics[width=\columnwidth]{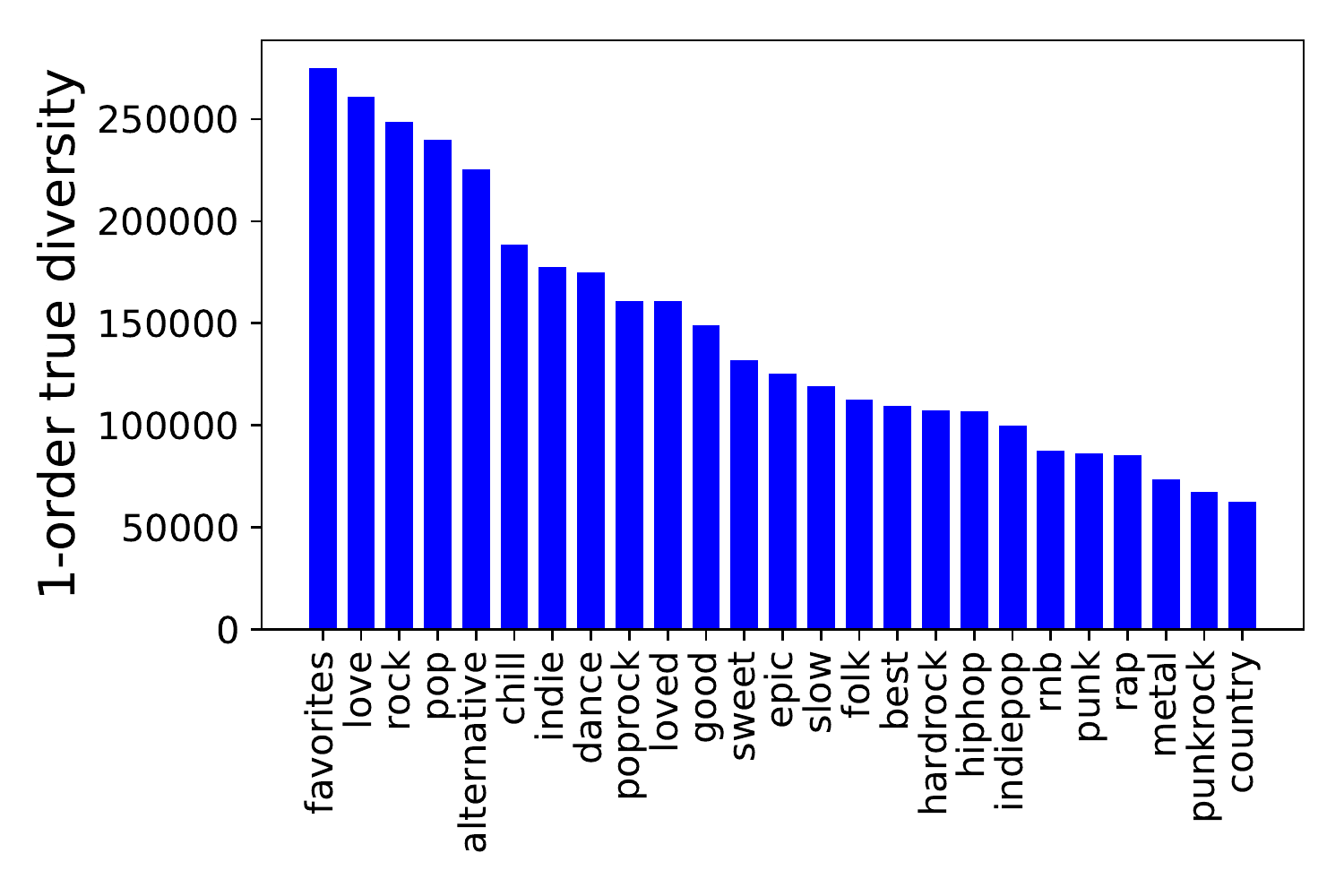}}
    \end{subfigure}
    \begin{subfigure}[c]{0.49\textwidth}
        \centering 
        \caption{\amz dataset}\label{fig:divaud25amz}
        \vspace{0.25cm}
        \includegraphics[width=\columnwidth,height=6.25cm]{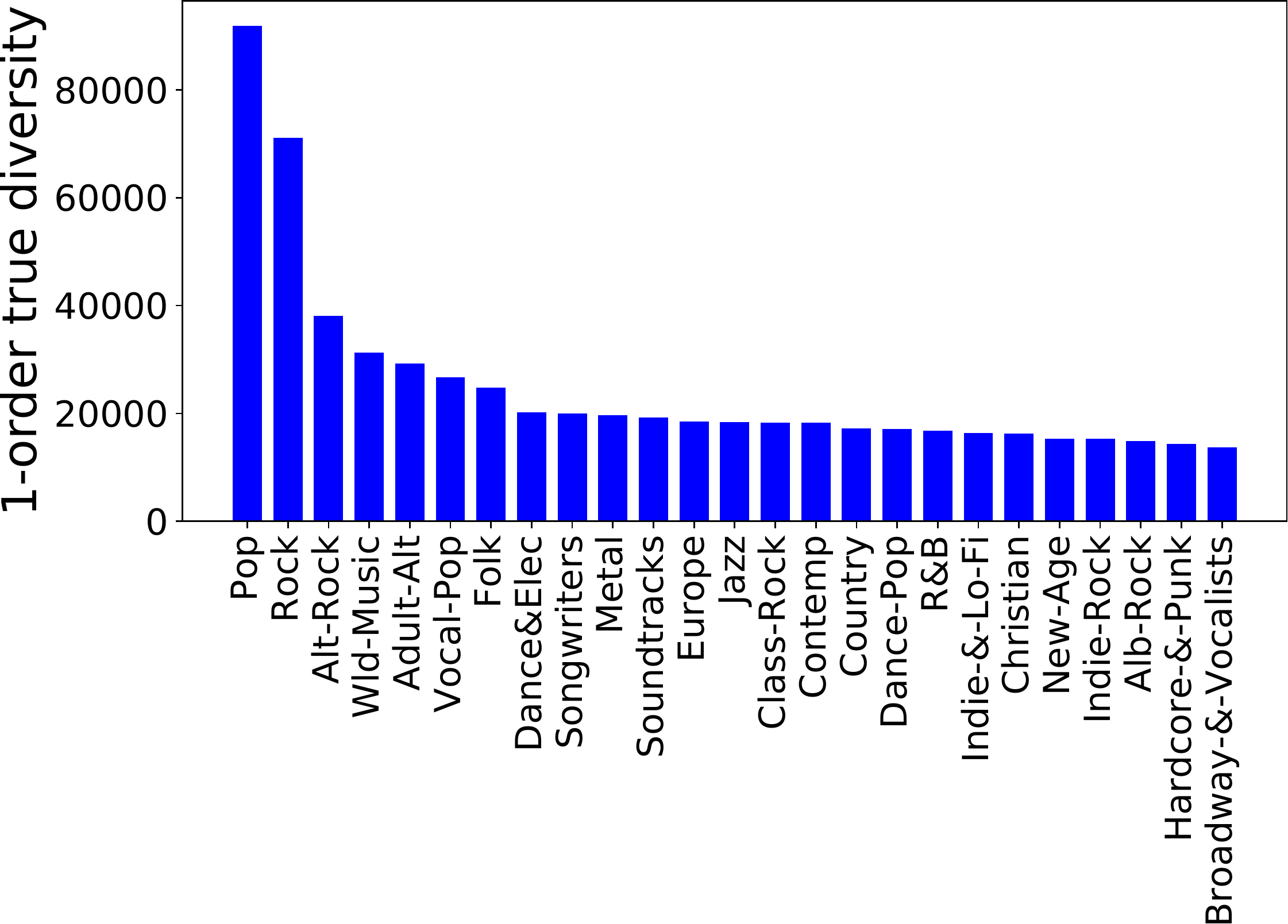}
    \end{subfigure}
\caption{Ordered 1-order true diversities of the tag audience $\D{1}(p_{\Pi_{\text{audience}}|t})$ for 25 selected tags for the \msd (left) and the \amz (right) datasets.}
\label{fig:divaud25}
\end{figure}

Both plots show strongly heterogeneous distributions of individual diversities: if most of the tags have a rather narrow audience, one can identify some tags with a particularly high
diversity.
This is the case for the tags \emph{Rock} and \emph{Pop} in
both datasets (see Figure~\ref{fig:divaud25}). 
But even for those tags,
their diversity value (around $10^5$ in \msd and $10^4$ in \amz) is
still one order of magnitude lower than the maximal theoretical values:
\np{1 019 190} for \msd and \np{465 249} for \amz (cf. Axiom~\ref{ax:normalization}).
One can, however, nuance this observation by noticing that small
diversity values are more homogeneously distributed. 
This is visible
in the insets of Figure~\ref{fig:divaud}, which focus on the
distribution of diversity values that are lower than \np{10 000} ($74\%$ of the nodes in
\msd, Figure~\ref{fig:divaudmsd}) and lower than \np{1 000} ($56\%$ of the nodes in
\amz, Figure~\ref{fig:divaudam}). 
One can see in particular that the values are well distributed around the mean value of the dataset
(respectively \np{24 850} for \msd and and \np{2 905} for \amz).  
This indicates that while one may spot some extremely diverse musical
contents (\emph{Rock} and \emph{Pop}, for instance), most of them are
narrowed towards a smaller and less diverse set of users (such as
\emph{Country} and \emph{Punkrock} in \msd or \emph{New-Age} in \amz).

In order to further investigate how such diversity measures can
be used to analyze specific categories, we show in
Figure~\ref{fig:divaud25} a selection of 25 tags for the two
datasets. 
It is worth mentioning here that for the \msd dataset, the tags are
actually provided by the users themselves that can decide to use any
word to tag any song (this is known as a \textit{folksonomy} \cite{peters2009folksonomies}). 
While most tags coincide with common music genres (like \emph{Rock}, \emph{Pop}, \emph{Folk}, \emph{Metal},
...), others are obviously meant to give an appreciation of the songs
(like \emph{Favorites}, \emph{Love}, \emph{Best}, ...) or even to
depict a moment at which a song is listened to (like
\emph{BeforeSleep} or \emph{InShower}).
The wide range of usage of the tags is an opportunity for us to
assess how our network diversity measures respond to those different
behaviors. 
For instance, one can expect tags like \emph{Favorites} to
be related to a broader and more diverse audience than \emph{Metal}
since the songs tagged by the former do not belong to a dedicated
musical category. 
This is indeed confirmed by Figure~\ref{fig:divaud25msd} which shows that popular tags like
\emph{Favorites} and \emph{Love} have a diversity higher than any
other tags of the dataset.

In contrast with the case of the \msd dataset, the classification imposed by \emph{Amazon} provides
only tags that describe musical genres. This allows for a direct
comparison of the musical categories presented in
Figure~\ref{fig:divaud25amz}, which provides interesting insights on
the way users commit to the different categories. 
For instance, if we compare \emph{Adult-Alternative} and \emph{World-Music} with
\emph{R\&B} and \emph{Dance-Pop}\footnote{We discard in the comparison
  \emph{Rock} and \emph{Pop} that have a particularly large number of
  users posting reviews to their songs, at least ten times higher than
  the number of users for any other tag in the dataset.}, it is
remarkable that the two former ones have a diversity twice higher
although the four tags have songs reviewed by the same number of users
(approximately \np{300 000} users). 
This is a clear indication that users posting reviews on \emph{Adult-Alternative} and
\emph{World-Music} songs are much more committed (the reviews are more
uniformly distributed among the users) than the ones of \emph{R\&B}
and \emph{Dance-Pop}.

\subsubsection{1-order true diversity of users' attention.}
\label{sec:divatt} 

\begin{figure}[h]
  \begin{subfigure}[c]{0.49\textwidth}
        \centering 
        \caption{\msd dataset}\label{fig:divattmsd}
        \includegraphics[width=\columnwidth]{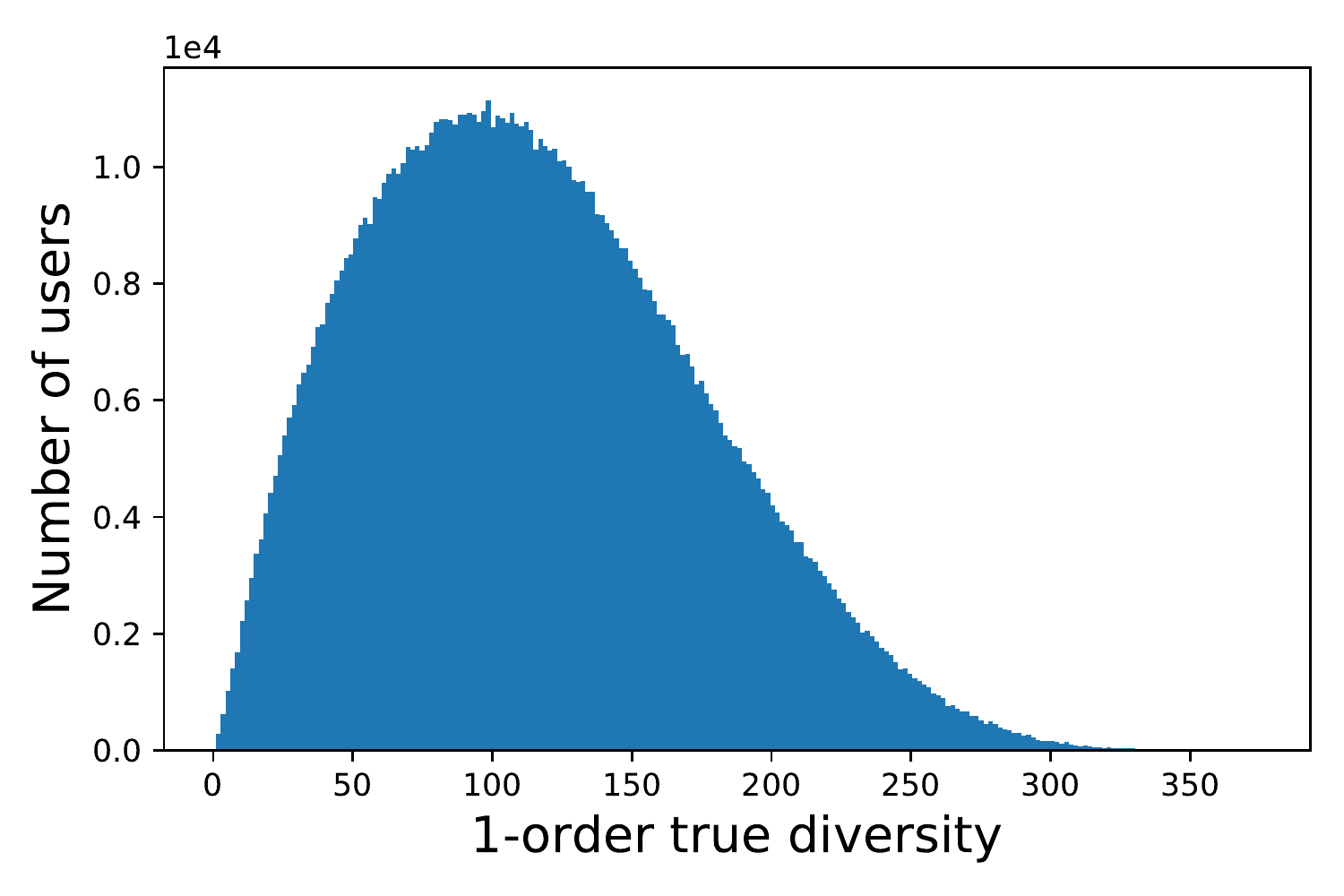}
    \end{subfigure}
    \begin{subfigure}[c]{0.49\textwidth}
        \centering 
        \caption{\amz dataset}\label{fig:divattamz}
        \vspace{0.2cm}
        \raisebox{0.25cm}{\includegraphics[width=\columnwidth,height=4.8cm]{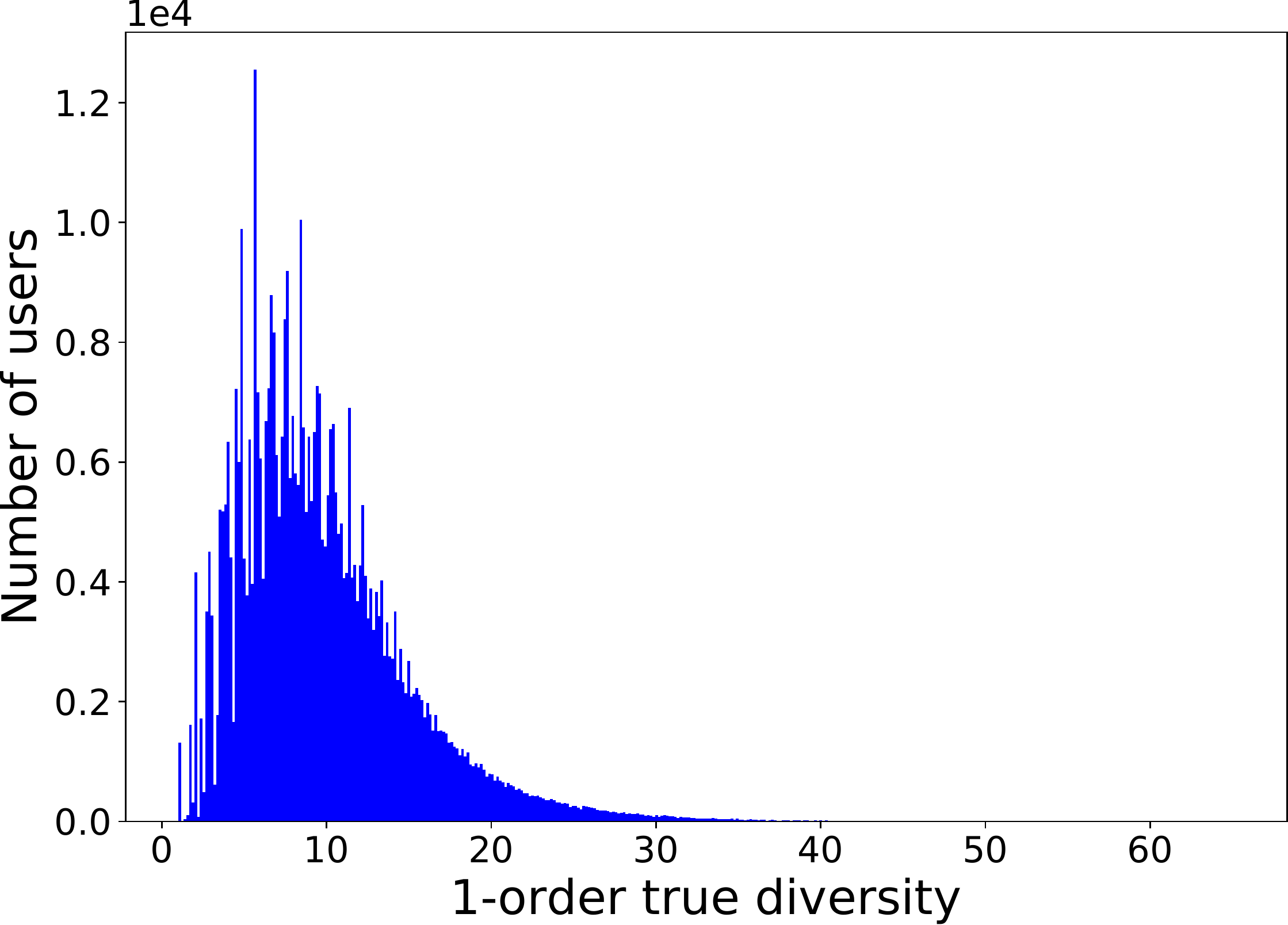}}
    \end{subfigure}
\caption{Histograms showing the distributions of the 1-order true diversities of the attention of users ($\D{1}(p_{\Pi_{\text{attention}}|u})$) for users $u\in\numusers{V}$ for the \msd (left) and \amz (right) datasets.}
\label{fig:divatt}
\end{figure}

We now turn to the diversity of users' attention, the diversity of tags listened by users. 
Figure~\ref{fig:divatt} presents the distribution of the 1-order true diversities $\D{1}(p_{\Pi_{\text{audience}}|t})$ for all users $u\in\numusers{V}$.
We compute and present these values using the \msd
(Figure~\ref{fig:divattmsd}) and \amz (Figure~\ref{fig:divattamz}) datasets. 
In contrast with the distributions presented in Figure~\ref{fig:divaud},
the diversity of users' attention is clearly homogeneous and centered
around small values (compared to the maximal theoretical ones). This
indicates that even if some users have a particularly high diversity,
the vast majority of them have a relatively narrow consumption of
the musical products.
It is worth noting that, compared to the study of tags, 
that often had a meaningful name, we have no
information regarding the profile of a user who is just an anonymized
value in the dataset. 
Thus we cannot focus on specific users to
provide an interpretation of the diversity values like we did in the
previous section.

However, it is possible to study how the diversity of the users
depends on their activity on the platform. More precisely, let us
define the \emph{volume} of a user $u\in\numusers{V}$ as the sum of the number of
tags for all songs listened to (in the case of the \msd dataset) or reviewed (in the case of the \amz dataset) by $u$, multiplied by their play count (the number of songs consumed by user $u$).
Then we can investigate whether there is a correlation
between volume and diversity. 
Intuitively, the highest the volume, the
highest the diversity: as its volume increases, a user has indeed more
opportunities to explores new musical categories, thus diversifying
its activity on the platform.

\begin{figure}[h]
  \begin{subfigure}[c]{0.49\textwidth}
        \centering 
        \caption{\msd dataset}\label{fig:evolattmsd}
        \includegraphics[width=\columnwidth]{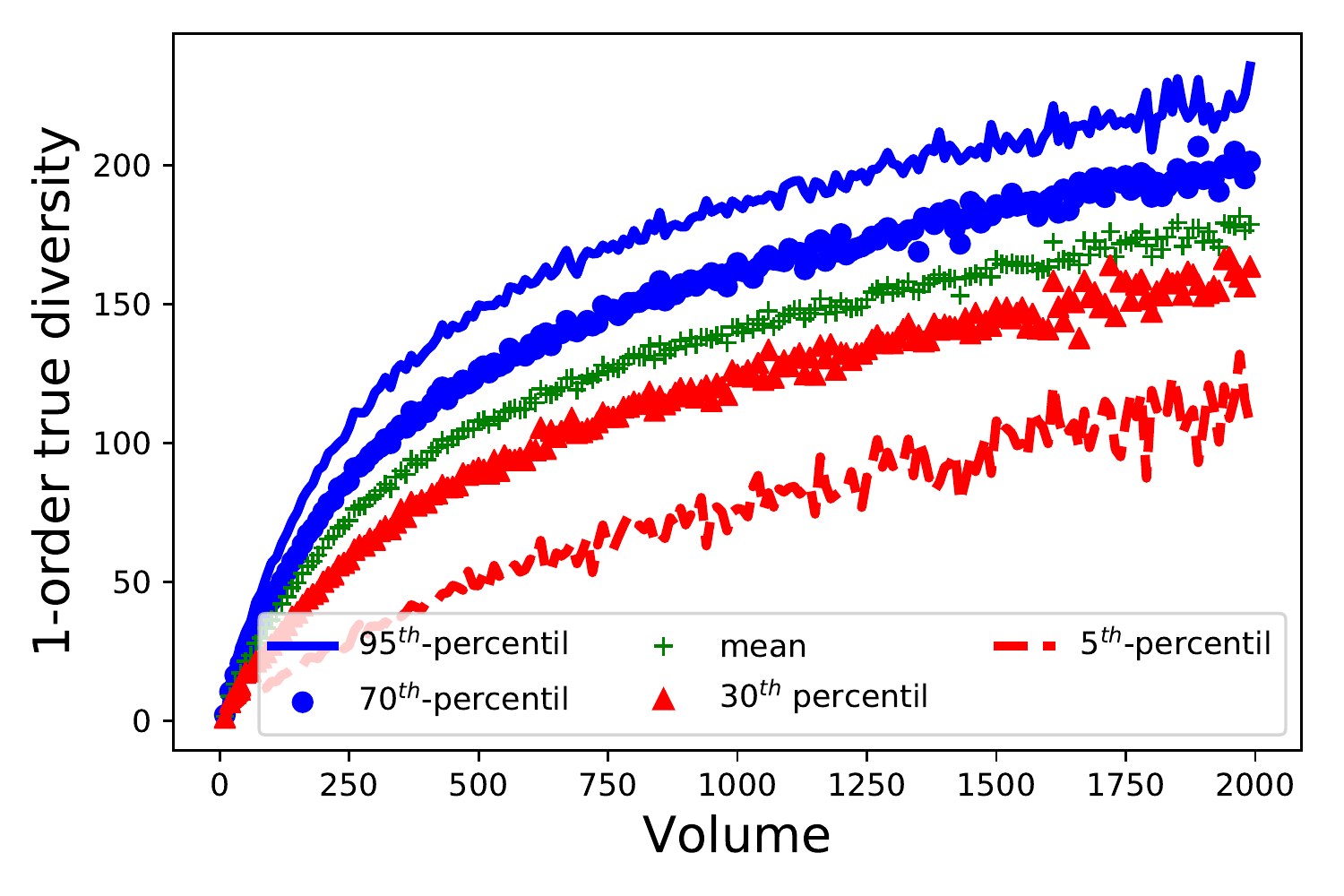}
    \end{subfigure}
    \begin{subfigure}[c]{0.49\textwidth}
        \centering 
        \caption{\amz dataset}\label{fig:evolattamz}
        \vspace{0.2cm}
        \raisebox{0.28cm}{\includegraphics[width=\columnwidth,height=4.8cm]{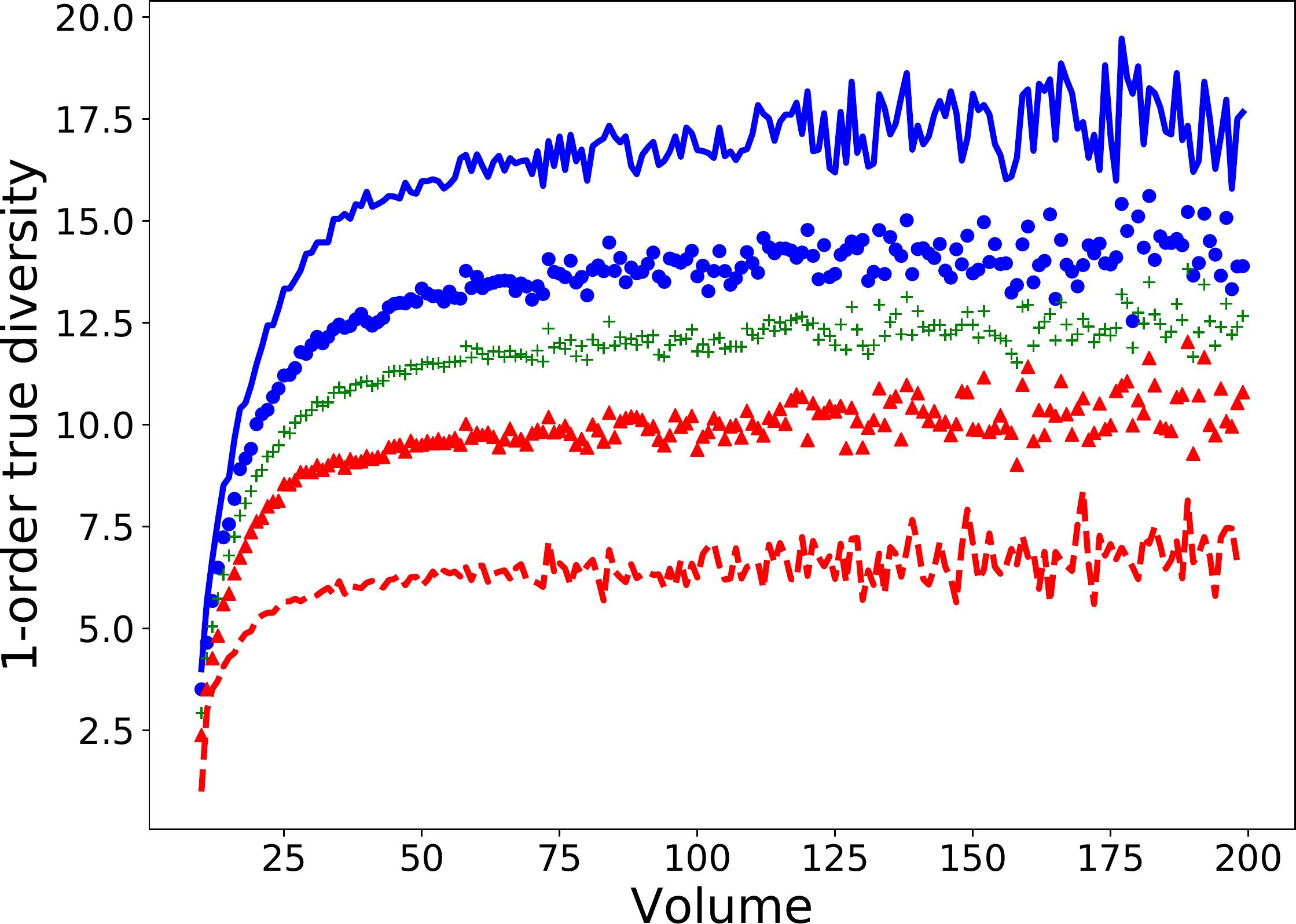}}
    \end{subfigure}
\caption{Evolution  of the 1-order true diversity of users' attention as a function of its volume for the \emph{Million Song Dataset} (left) and \emph{Amazon Dataset} (right).}
\label{fig:evolatt}
\end{figure}

To see this more clearly, Figure~\ref{fig:evolatt}
presents the mean value of the 1-order true diversity of a user as a
function of its volume, along with the $5^{th}$-, $30^{th}$-,
$70^{th}$- and $95^{th}$-percentile. 
For both dataset, we can observe
that the diversity increases along with the volume. However, we can
also notice that the influence of the volume is clearly lower after a
given threshold, highlighting a \emph{saturation process} in the
diversity of users' attention: while the growth of the diversity is
initially sharp as the volume increases, after a given threshold
(around $250$ in \msd and $25$ in \amz), the users listen repeatedly to, or review similar contents proposed by the platform. This
is particularly obvious in \amz (Figure~\ref{fig:evolattamz}) but one
can also spot this phenomenon on \msd (Figure~\ref{fig:evolattmsd}).

After having presented a simple example of application of the network diversity measures in Section~\ref{subsec:simple_example} and a numerical example with real datasets in this Section~\ref{subsec:numerical_example}, we present, in the rest of Section~\ref{sec:part5} application examples for different research domains

%%%%%%%%%%%%%%%%%%%%%%%%%%%%%%%%%%%%%%%%%%%%%%%%%%%%%%%%
\subsection{Recommender Systems}
\label{subsec:recommender_systems}

Diversity and diversification of algorithmic recommendations has become one of the leading topics of the recommender systems research community \cite{kunaver2017diversity,zhou2010solving}.
Through a variety of means, users have access today to large numbers of items (\eg, products and services in e-commerce, messages and posts in social media, or news articles in aggregators).
While users enjoy an ever-growing offer, it can also become unmanageable for them to consider enough items, or to effectively explore all that is offered.
Recommender systems, developed as early as in the 1980s \cite{salton1983introduction}, help solve this problem by filtering all possible items down to a recommended set tailored for each user or group.
One recent advance in this field is the recognition of the importance of diversity and its introduction in recommendations \cite{mcnee2006being,bradley2001improving}.

In recommender systems, diversity can help improve users' appreciation of the quality of recommendations \cite{silveira2019good,bobadilla2013recommender}.
It also has other applications, such as detecting changes in consumption behavior for context-aware recommenders \cite{l2016modeliser}.
As a property of recommendations, diversity has been traditionally captured by a set of related indicators proposed on intuitive bases, called \emph{serendipity}, \emph{discovery}, \emph{novelty}, \emph{dissimilarity} (see Section 8.3 of~\cite{kantor2010recommender}, or~\cite{silveira2019good,bobadilla2013recommender} for a discussion of terminology and definitions).
These indicators are often computed using past collective choices of items made by users \cite{zhou2010solving}, or classifications of items into types \cite{ziegler2005improving}.
To this date, no general framework exists to account for all proposed diversity indices in recommender systems, nor alternatives for exploiting richer meta-data structures such as those encodable by heterogeneous information networks.
This is where our proposed network diversity measures find valuable applications.
They accommodate some of the existing concepts from the literature, extend the measurement of diversity to more complex data structures that can include meta-data on users and items, and give formal explicit expressions to computable quantities related to new and existing research questions in this field.

\begin{figure}[h!]
    \centering
      \begin{center}
    \begin{tikzpicture}
      
      \node [draw, circle, minimum width = 1cm ] (U) {$\U$};
      \node [draw, circle, minimum width = 1cm , right = 1.5cm of U] (I1) {$\IOne$};
      \node [draw, circle, minimum width = 1cm , below = 1.5cm of U] (I2) {$\ITwo$};
      \node [draw, circle, minimum width = 1cm , right = 1.5cm of I1] (T1) {$\TOne$};
      \node [draw, circle, minimum width = 1cm , right = 1.5cm of I2] (T2) {$\TTwo$};
      \node [draw, circle, minimum width = 1cm , above left =  0.5cm and 1.5cm of U] (G1) {$\GOne$};
      \node [draw, circle, minimum width = 1cm , below = 1.5cm of G1] (G2) {$\GTwo$};

      % \draw [->] (U) to[out=60,in=100,distance=1cm] (I1);
      \draw [->] (U) to[out=45,in=135,edge node={node [midway,above] {$\UlikeIOne$}}] (I1);
      \draw [->] (U) to[out=0,in=180,edge node={node [midway,above] {$\UseenIOne$}}] (I1);
      \draw [->] (U) to[out=-45,in=-135,edge node={node [midway,above] {$\UrecIOne$}}] (I1);

      \draw [->] (U) to[out=-110,in=110,edge node={node [midway,left] {$\UrateITwo$}}] (I2);
      \draw [->] (U) to[out=-70,in=70,edge node={node [midway,right] {$\UrecITwo$}}] (I2);

      \draw [->] (I1) to[edge node={node [midway,above] {$\IOneTOne$}}] (T1);
      \draw [->] (I1) to[edge node={node [midway,right] {$\IOneTTwo$}}] (T2);
      \draw [->] (I2) to[edge node={node [midway,above] {$\ITwoTTwo$}}] (T2);

      \draw [->] (U) to[edge node={node [midway,above] {$\GOneU$}}] (G1);
      \draw [->] (U) to[edge node={node [midway, above] {$\GTwoU$}}] (G2);

      \draw [->] (U) to[out=110,in=70,distance=1cm,edge node={node [midway,above] {$\EU$}}] (U);

    \end{tikzpicture}
  \end{center}
%\end{minipage}  
    \caption{Network schema of a heterogeneous information network in a setting from recommender systems, where users in $\U$, belonging to groups $\GOne$ and $\GTwo$ interact and are recommended two different sets of items $\IOne$ and $\ITwo$, which are classified using types in $\TOne$ and $\TTwo$.}
    \label{fig:rs}
  \end{figure}
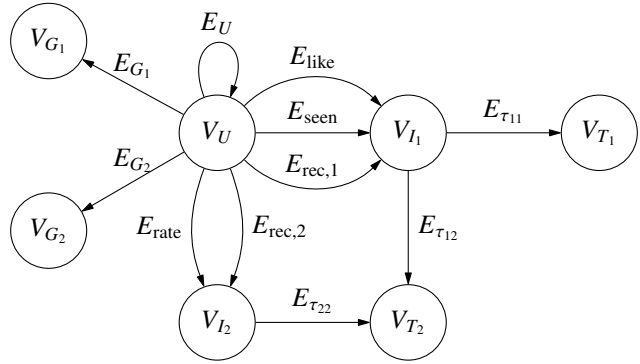

For illustrative purposes, let us consider a heterogeneous information network giving an ontology to complex data related to a situation in which we have recommended different types of items to users.
Figure~\ref{fig:rs} shows the network schema of the heterogeneous information network to be considered in this example.
Let us consider the following vertex types for the example:
\begin{itemize}
\item A vertex type of users $\U$;
\item Two vertex types for items: $\IOne$ (\eg, films) and $\ITwo$ (\eg, series);
\item Two vertex types for item classification: $\TOne$ (\eg channels/distributors) and $\TTwo$ (\eg genre);
\item Two vertex types of user groups: $\GOne$ (\eg, demographic group) and $\GTwo$ (\eg, location).
\end{itemize}

In order to consider random walks constrained to \metapath{}s, and following the example in Section~\ref{subsec:simple_example}, we denote with capital letter $X$ the random variables supported by a vertex type.
For example, $\XU$ is a random vertex in $\U$ and $\XIOne$ is a random vertex in $\IOne$.

In the heterogeneous information network illustrated in Figure~\ref{fig:rs}, we also consider different edge types:
\begin{itemize}
\item An edge type between users $\EU$ (\eg, users following or friending each others on a social network);
\item Edge types from groups to users: $\GOneU$ (\eg, associating users with demographic groups) and $\GTwoU$  (\eg, associating users with locations);
\item Edge types indicating classification of items into types: $\IOneTOne$ and $\IOneTTwo$ for $\IOne$, and $\ITwoTTwo$ for $\ITwo$;
\item Edge types representing when users have liked ($\UlikeIOne$), seen ($\UseenIOne$), or rated ($\UrateITwo$) an item, or representing when users have been recommended items ($\UrecIOne$ and $\UrecITwo$).
\end{itemize}

All elements in the proposed example are useful for representing common practices in recommender systems.
Settings for recommendation where there are two --or more-- types of items ($\IOne$ and $\ITwo$ in our example) are common in cross-domain recommendation \cite{tang2012cross}, and more generally in heterogeneous information network recommendation \cite{yu2014personalized}.
Relations between users and items can be of different kinds in recommendation settings:
edges can be used to indicate that a user has rated an item in \emph{explicit feedback} --or scoring, or noting-- systems ($\UrateITwo$ in the example), or to indicate that a user has liked an item in \emph{implicit feedback} systems ($\UlikeIOne$ in the example).
Some recommender systems and diversity measures can take into account whether a user has previously seen an item \cite{vargas2011rank} ($\UseenIOne$ in the example).
Also, settings where meta-data are associated with users is very common in demographic or location filtering \cite{safoury2013exploiting}, and are represented in the example by using vertex types $\GOne$ and $\GTwo$.
Finally, edges between users signaling relations such as friendship of a user \emph{following} another one on social networks ($\EU$ in the example) may also be exploited for recommendations \cite{groh2007recommendations,bernardes2015social}, and certainly in diversity computations.
As stated before, Figure~\ref{fig:rs} represents the network schema (cf. Definition~\ref{def:network_schema}) of our example.

Most diversity computations in recommender systems consist in providing a measure of the diversity of items recommended to a user, or an aggregation of this quantity for all users.
Diversity between items can be computed, for example, with respect to a classification of items \cite{ziegler2005improving} (\eg, genres for films).
In the proposed framework, this concept would be captured by the individual diversity.
Let us imagine that $\U$ are users, that $\ITwo$ are films, and that $\TTwo$ are film genres (\eg, comedy, thriller, etc.).
The individual genre ($\TTwo$) diversity of films ($\ITwo$) recommended to a user $u\in\U$ is 
$$\D{\alpha}\left(\XU\xrightarrow{\UrecITwo}\XITwo\xrightarrow{\IOneTTwo}\XTOne\given \XU=u \right).$$
\noindent Similarly, the mean genre ($\TTwo$) diversity of films ($\ITwo$) recommended to all users $\U$ is the mean individual diversity
$$\D{\alpha}\left(\XU\xrightarrow{\UrecITwo}\XITwo\xrightarrow{\IOneTTwo}\XTOne\given \XU \right),$$
which can be computed as a geometric mean by choosing $\XU\sim\text{Uniform}(\U)$ for the starting point of the \metapath{} constrained random walk.

In another classic setting, the diversity of an item is computed according to the number of users that have previously chosen or liked it (sometimes called \emph{novelty} \cite{hurley2011novelty}).
In the proposed framework, an aggregation of this quantity for items proposed to all users corresponds to the following network diversity measure:
$$
\D{\alpha}\left(\XU\xrightarrow{\UrecIOne}\XIOne\xrightarrow{\UlikeIOne^\intercal}\XU' \given \XU \right).
$$

More interestingly, other relevant quantities expressible as network diversity measures have no explicit expression in other frameworks of the literature.
The clearest example is the comparison between the mean individual and collective recommended diversities: for example, $\D{\alpha}\left(\XU\xrightarrow{\UrecIOne}\XIOne\xrightarrow{\IOneTOne}\XTOne\given \XU \right)$ versus $\D{\alpha}\left(\XU\xrightarrow{\UrecIOne}\XIOne\xrightarrow{\IOneTOne}\XTOne\right)$.
Distinguishing between these two concepts (cf. Figure~\ref{fig:individual_vs_collective}) is important when taking interest in diversity beyond its use as a quality of recommendations; for example, when studying phenomena such as filter bubbles, which could manifest at some level of aggregation of users, while still having a high collective diversity.
Some concepts at the core of Recommender Systems could also be expressed in our network diversity framework, such as the so-called \emph{User-Based Collaborative Filtering} (see Section 4.2 of \cite{kantor2010recommender}):
$$
\D{\alpha}\left(\XU\xrightarrow{\UrecIOne}\IOne\xrightarrow{\UlikeIOne^\intercal}\XU'\xrightarrow{\UlikeIOne}\IOne'\xrightarrow{\IOneTOne}\TOne \right),
$$
which corresponds to the \emph{collective type diversity of items chosen by users that chose items recommended to users}.

Let us present in a schematic fashion, in Table~\ref{tab:mapping_rs_concepts_to_networks_diversities},  different examples of concepts related to diversity that are of interest for research questions in the domain of recommender systems, along with the respective quantities that can be identified, expressed, and computed as network diversity measures.
% [Next lines are added in December 2020 after 2nd revision]
The reader is referred to \cite{morales2020testing} for an example of the application of the network diversity measures in conjunction with recommendation tasks. 
In the cited article, after presenting different experimental protocols and datasets known to the Recommender Systems community, the authors examine the performance of recommendations using the networks diversity measures, whose theoretical development and properties are the object of this article. 

\begin{table}[h!]
\centering
\caption{Schematic representation of examples of concepts related to diversity in recommender systems and the network diversity measures that can be used to address them in quantitative studies.}
\scriptsize
\label{tab:mapping_rs_concepts_to_networks_diversities}
\begin{tabular}{p{8.2cm}p{6cm}}
Examples of concepts expressible in research questions & Corresponding network diversity measures \\
\hline

% Diversity of items $\IOne$ recommended to user $u\in\U$ according to types $\TOne$ &
% $\D{\alpha}\left(\XU\xrightarrow{\UrecIOne}\XIOne\xrightarrow{\IOneTOne}\XTOne\given \XU=u \right)$\\

% Mean diversity of items $\IOne$ recommended to users $\U$ according to types $\TOne$ &
% $\D{\alpha}\left(\XU\xrightarrow{\UrecIOne}\XIOne\xrightarrow{\IOneTOne}\XTOne\given \XU \right)$\\

% Collective diversity of items $\IOne$ recommended to users $\U$ according to types $\TOne$ &
% $\D{\alpha}\left(\XU\xrightarrow{\UrecIOne}\XIOne\xrightarrow{\IOneTOne}\XTOne \right)$\\

% Diversity of items $\IOne$ recommended to user $u\in\U$ according to users that liked those items &
% $\D{\alpha}\left(\XU\xrightarrow{\UrecIOne}\XIOne\xrightarrow{\UlikeIOne^\intercal}\XU\given \U=u \right)$\\

Mean individual diversity of recommendation of items $\IOne$ according to types $\TOne$ relative to the corresponding collective diversity &
$\D{\alpha}\left(\XU\xrightarrow{\UrecIOne}\XIOne\xrightarrow{\IOneTOne}\XTOne \relativeto \XU\xrightarrow{\UrecIOne}\XIOne\xrightarrow{\IOneTOne}\XTOne \given \XU \right)$\\

Collective diversity of recommended items $\IOne$ according types $\TOne$ relative to the distribution of types of liked times &
$\D{\alpha}\left(\XU\xrightarrow{\UrecIOne}\XIOne\xrightarrow{\IOneTOne}\XTOne \relativeto \XU\xrightarrow{\UlikeIOne}\XIOne\xrightarrow{\IOneTOne}\XTOne \right)$\\

Collective diversity of recommendations of items in $\IOne$ (\eg, films) according to types $\TTwo$ (\eg, genres) relative to the one of $\ITwo$ (\eg, series) &
$\D{\alpha}\left(\XU\xrightarrow{\UrecIOne}\XIOne\xrightarrow{\IOneTTwo}\XTTwo \relativeto \XU\xrightarrow{\UrecITwo}\XITwo\xrightarrow{\ITwoTTwo}\XTTwo \right)$\\

Diversity of items $\IOne$ recommended to friends of $u\in\U$ according to types $\TOne$ &
$\D{\alpha}\left(\XU\xrightarrow{\EU}\XU'\xrightarrow{\UrecIOne}\XIOne\xrightarrow{\IOneTTwo}\XTOne \given \XU=u\right)$\\

Diversity of items $\IOne$ liked by group $g\in\GOne$ according to types $\TOne$ &
$\D{\alpha}\left(\XGOne\xrightarrow{\GOneU^\intercal}\XU\xrightarrow{\UlikeIOne}\XIOne\xrightarrow{\IOneTTwo}\XTOne \given \XGOne=g\right)$\\

Diversity of users $\U$ that liked items $\IOne$ of type $t\in\TTwo$ &
$\D{\alpha}\left(\XU \given \XU\xrightarrow{\UlikeIOne}\XIOne\xrightarrow{\IOneTTwo}\XTTwo = t\right)$\\

Diversity of types $\TTwo$ chosen by user $u\in\U$ through their choices of items in $\IOne$   &
$\D{\alpha}\left( \XU\xrightarrow{E_{\left(\UlikeIOne,\IOneTTwo\right)}}\XTTwo \given \XU=u\right)$\\

\hline
\end{tabular}

\end{table}

%%%%%%%%%%%%%%%%%%%%%%%%%%%%%%%%%%%%%%%%%%%%%%%%%%%%%%%%%%%%%%%%%%%%%%%
\subsection{Social media studies, echo chambers, and filter bubbles}
\label{subsec:social_media}

The study of social media has been developed into a large and ever growing wealth of results.
The importance of studies about the creation, transmission, and consumption of information on social networks has become crucial.
Heterogeneous information networks provide a natural formalism for the treatment of these objects, as they can accommodate a variety of entities (\eg, posts, accounts, media outlets, tags, keywords) interacting through many different relations (\eg, users publishing posts, mentioning or following other users, using tags in publication).
More complex and abstract data is often analyzed in these studies, such as the political affiliations of users and media outlets \cite{gaumont2018reconstruction,flaxman2016filter}.
The analysis of phenomena such as echo chambers and filter bubbles through the measurement of diversity of information consumption is an established practice~\cite{div2, recdiv, sha2016framework, impact2, divmusic2}.
The settings of different social media studies vary.
Concrete examples are the study of the \emph{Leave} and \emph{Remain} Brexit campaigns on Twitter \cite{bastos2018geographic} and the exchange of information between US Democrats and Republicans on Facebook \cite{bakshy2015exposure}.

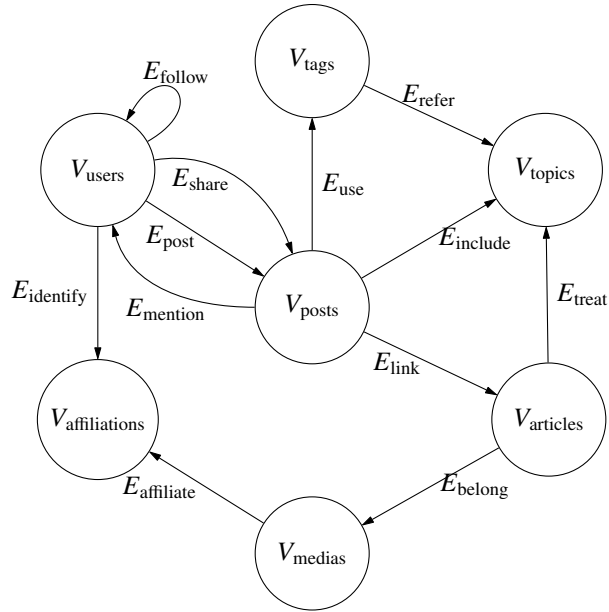
\begin{figure}[!h]
    \centering
    \newlength{\mywidth}
\setlength{\mywidth}{1.5cm}
\newlength{\mysep}
\setlength{\mysep}{1.75cm}

  \begin{center}
    \begin{tikzpicture}
      
      \node [draw, circle, minimum width = \mywidth ] (U) {$\users$};
      \node [draw, circle, minimum width = \mywidth , below right = 0.75cm and \mysep of U] (Posts) {$\posts$};
      \node [draw, circle, minimum width = \mywidth , below = \mysep of U] (Affiliations) {$\affiliations$};

      \node [draw, circle, minimum width = \mywidth , below = \mysep of Posts] (Medias) {$\medias$};

      \node [draw, circle, minimum width = \mywidth , right = 2.5*\mysep of U] (Topics) {$\topics$};
      \node [draw, circle, minimum width = \mywidth , right = 2.5\mysep of Affiliations] (Articles) {$\articles$};
      \node [draw, circle, minimum width = \mywidth , above = \mysep of Posts] (Tags) {$\tags$};

      \draw [->] (U) to[edge node={node [midway, left] {$\Upub$}}] (Posts);
      \draw [->] (U) to[out=30,in=60,distance=1cm,edge node={node [midway, above] {$\Ufollow$}}] (U);
      \draw [->] (U) to[out=10,in=110, edge node={node [midway, left] {$\Ushares$}}] (Posts);
      \draw [->] (Posts) to[out=180,in=-75,edge node={node [midway, below] {$\Pmentions$}}] (U);
      \draw [->] (U) to[edge node={node [midway, left] {$\Uaff$}}] (Affiliations);
      \draw [->] (Posts) to[edge node={node [midway, left] {$\Part$}}] (Articles);
      \draw [->] (Articles) to[edge node={node [midway, right] {$\Amed$}}] (Medias);
      \draw [->] (Medias) to[edge node={node [midway, left] {$\Maff$}}] (Affiliations);
      \draw [->] (Posts) to[edge node={node [midway, right] {$\Ptop$}}] (Topics);
      \draw [->] (Articles) to[edge node={node [midway, right] {$\Atop$}}] (Topics);
      \draw [->] (Posts) to[edge node={node [midway, right] {$\Ptag$}}] (Tags);
      \draw [->] (Tags) to[edge node={node [midway, above] {$\Ttop$}}] (Topics);

    \end{tikzpicture}
  \end{center}
%\end{minipage}  
    \caption{Network schema of a heterogeneous information network in a setting from social networks and media studies, suited for the study of echo chambers and filter bubbles.}
    \label{fig:bubbles}
\end{figure}

In this section, we illustrate the use of network diversity measures for the study of information exchange on social networks.
We consider, as before, a heterogeneous information network created from activity traces of social networks and media.
Figure~\ref{fig:bubbles} shows its network schema, with which we may illustrate the use of network diversity measures in this context.
In this example, we consider:
users that post or share posts (or \emph{tweets}, or blog entries, or comments in forums), users that can follow (or befriend) other users, posts that may mention users, include tags (\eg, \emph{hashtags}), include topics (detectable, for example, by matching strings or using topic discovery methods), and even link to articles through a URL address. 
In many contexts, articles may be associated with media outlets, which may in turn be identified with groups or affiliations (\eg, political parties).

The considered heterogeneous information network can accommodate different aspects considered in social media studies.
For example, Gaumont et al.~\cite{gaumont2018reconstruction} consider relations of political affiliation of users and interactions between them, and analyze the notion of diversity of Twitter posts according to the political communities they have reached.
Other studies also consider the use of entropy measures over distributions representing the proportion of users that browse given information sources~\cite{nikolov2015measuring}.
Some studies, for example~\cite{roth2013socio}, explicitly consider networks of information items (\eg, blog posts) and the concepts that they use.

As with the example of a generic recommender systems, we present in a schematic fashion (in Table~\ref{tab:mapping_new_media_concepts_to_networks_diversities}) different diversity-related concepts of interest for research questions in the field of social networks and media studies, along with quantities that can be computed as network diversity measures.

\begin{table}[!h]
\centering
\caption{Schematic representation of diversity-related concepts in social networks and media studies, and the corresponding network diversity measures that can be used to address them in quantitative studies.}
\scriptsize
\label{tab:mapping_new_media_concepts_to_networks_diversities}
\begin{tabular}{p{7cm}p{8cm}}
Examples of concepts expressible in research questions & Corresponding network diversity measures \\
\hline

Collective diversity of affiliations of users &
$\D{\alpha}\left(\Xusers\xrightarrow{\Uaff}\Xaffiliations \right)$\\

% Mean individual affiliation $\affiliations$ diversity of users $\users$ &
% $\D{\alpha}\left(\Xusers\xrightarrow{\Uaff}\Xaffiliations \given \Xusers \right)$\\

Collective affiliation diversity of users through the contents they share &
$\D{\alpha}\left(\Xusers\xrightarrow{\Ushares}\Xposts\xrightarrow{\Part}\Xarticles\xrightarrow{\Amed}\Xmedias\xrightarrow{\Maff}\Xaffiliations \right)$\\

Individual topic diversity of user $u\in\users$ through posting &
$\D{\alpha}\left(\Xusers\xrightarrow{\Upub}\Xposts\xrightarrow{\Ptop}\Xtopics \given \Xusers=u \right)$\\

Individual topic diversity of user $u\in\users$ through posting of followers &
$\D{\alpha}\left(\Xusers\xrightarrow{\Ufollow^\intercal}\Xusers'\xrightarrow{\Upub}\Xposts\xrightarrow{\Ptop}\Xtopics \given \Xusers=u \right)$\\

Individual affiliation diversity of users mentioned by user $u\in\users$ &
$\D{\alpha}\left(\Xusers\xrightarrow{\Upub}\Xposts\xrightarrow{\Pmentions}\Xusers'\xrightarrow{\Uaff}\Xaffiliations \given \Xusers=u \right)$\\

Diversity of affiliation groups that treat topic $t\in\topics$ in articles &
$\D{\alpha}\left(\Xaffiliations \given \Xaffiliations\xrightarrow{\Maff^\intercal}\Xmedias \xrightarrow{\Amed^\intercal}\Xarticles\xrightarrow{\Atop}\topics=t \right)$\\

Affiliation diversity of users according to who they mention in their posts relative to their own identified political affiliation &
$\D{\alpha}\left(\Xusers\xrightarrow{\Upub}\Xposts\xrightarrow{\Pmentions}\Xusers'\xrightarrow{\Uaff}\Xaffiliations \relativeto \Xusers\xrightarrow{\Uaff}\Xaffiliations \right)$\\
\hline
\end{tabular}

\end{table}

%%%%%%%%%%%%%%%%%%%%%%%%%%%%%%%%%%%%%%%%%%
\subsection{Ecology}
\label{subsec:ecology}

Diversity is useful in ecology, as identified and commented in Section~\ref{subsec:diversity_of_diversities}.
Many advances in diversity measures come from this community (\eg, \cite{hill1973diversity}).
One prominent concept in this domain is the diversity of species in a habitat.
For the computation of quantitative indices of this diversity, individuals from different species are counted or their number is estimated.
From their apportionment into the species present in a habitat, diversity is then computed and reported.

Interactions among organisms are also of interest in ecology.
These can be treated using graph representations and models.
One of such interactions, also related to diversity, is represented by so-called \emph{food webs} \cite{paine1966food}: network models that describe species that feed on other species.
In the past, there have been efforts to use graph formalisms to treat food webs \cite{may1983ecology,lundgren1989food}.
Similarly, other relations between species have been described using graphs, such as parasitation \cite{poulin2010network}.
Another subject of interest in ecology is the description of habitats and their interconnectedness; there have been several approaches using graph theory to describe these connections \cite{urban2001landscape,wirth1966graph}.

We suggest that all these elements present in ecology can be treated using heterogeneous information networks.
Using network diversity measures, different concepts related to diversity can be computed.
Let us consider for example a heterogeneous information network with vertex types for habitats ($\habitats{V}$), for individuals ($\individuals{V}$), for species ($\species{V}$), for genera ($\genera{V}$), for families ($\families{V}$), and so on as needed.

Let us also consider for our example several edge types.
Edge type $\connect$ is that of edges between habitats, indicating whether an individual can access a given habitat from another one.
Edge type $\inhabit$ is used to represent which individuals inhabit which habitats.
Edge types $\eat$ and $\parasite$ are used to represent relations between species; which species eat which species, and similarly for parasitation.
Edge type $\belongS$ is used to represent which individual belongs to which species.
Finally, edge types $\belongG$ and $\belongF$ contain edges indicating which specie belongs to which genus, and which genus belongs to which family (species, genera, and families are three of the eight major taxonomic ranks in biological classification).

This setting can accommodate the common practices of measurement of --bio-- diversity of a habitat $h\in\habitats{V}$, which in network diversity measures finds the expression
$$\D{\alpha}\left(\habitats{X}\xrightarrow{\inhabit^\intercal}\individuals{X}\xrightarrow{\belongS}\species{X} \given \habitats{X}=h \right),$$
with $\alpha=0$ giving the richness biodiversity, and $\alpha=\infty$ the Berger-Parker biodiversity of the  habitat.
Figure~\ref{fig:ecology} illustrates the network schema of the described heterogeneous information network, along with a table of diversity-related concepts and their expressions as network diversity measures.

\begin{figure}[!h]
    \centering
    %\begin{minipage}[t]{0.45\linewidth}
\newlength{\widtheco}
\setlength{\widtheco}{1.5cm}
\newlength{\sepeco}
\setlength{\sepeco}{1.3cm}
  \begin{center}
    \begin{tikzpicture}

      \node [draw, circle, minimum width = \widtheco ] (habitats) {$\habitats{V}$};
      \node [draw, circle, minimum width = \widtheco , right = \sepeco of habitats] (individuals) {$\individuals{V}$};
      \node [draw, circle, minimum width = \widtheco , right = \sepeco of individuals] (species) {$\species{V}$};
      \node [draw, circle, minimum width = \widtheco , right = \sepeco of species] (genera) {$\genera{V}$};
      \node [draw, circle, minimum width = \widtheco , right = \sepeco of genera] (families) {$\families{V}$};
      \node [ circle,minimum width = \widtheco , right = \sepeco of families] (uptree) {$\cdots$};

      \draw [->] (individuals) to[edge node={node [midway, above] {$\inhabit$}}] (habitats);
      \draw [->] (individuals) to[edge node={node [midway, above] {$\belongS$}}] (species);
      \draw [->] (species) to[edge node={node [midway, above] {$\belongG$}}] (genera);
      \draw [->] (genera) to[edge node={node [midway, above] {$\belongF$}}] (families);
      \draw [->] (families) to[edge node={node [midway, above] {$\belongBeyond$}}] (uptree);

      \draw [->] (species) to[out=110,in=70,distance=1cm,edge node={node [midway, above] {$E_\text{eat}$}}] (species);
      \draw [->] (species) to[out=-110,in=-70,distance=1cm,edge node={node [midway, below] {$E_\text{parasite}$}}] (species);
      \draw [->] (habitats) to[,out=110,in=70,distance=1cm,edge node={node [midway, above] {$E_\text{connect}$}}] (habitats);

    \end{tikzpicture}
  \end{center}
%\end{minipage}

%%% Local Variables:
%%% mode: latex
%%% TeX-master: "../main"
%%% End: 
    {\scriptsize
    \begin{tabular}{p{7cm}p{8cm}}
Examples of concepts expressible in research questions & Corresponding network diversity measures \\
\hline

Species diversity in habitat $h\in\habitats{V}$ &
$\D{\alpha}\left(\species{X} \given \species{X}\xrightarrow{\belongS^\intercal}\individuals{X}\xrightarrow{\inhabit}\habitats{X}=h \right)$\\

Genera diversity in habitat $h\in\habitats{V}$ &
$\D{\alpha}\left(\genera{X} \given\genera{X}\xrightarrow{\belongG^\intercal} \species{X}\xrightarrow{\belongS^\intercal}\individuals{X}\xrightarrow{\inhabit}\habitats{X}=h \right)$\\

Species diversity of habitats adjacent to those where a species $s\in\species{V}$ is present &
$\D{\alpha}\left(\species{X}\xrightarrow{\belongS^\intercal}\individuals{X}\xrightarrow{\inhabit}\habitats{X}\xrightarrow{\connect}\habitats{X}'\right.$ 
$\left.\quad\quad\quad\quad\quad\quad\quad\quad\quad\quad \xrightarrow{\inhabit^\intercal}\individuals{X}'\xrightarrow{\belongS}\species{X}'\given \species{X}=s \right)$\\

Species diversity of the predators of species that parasite a species $s\in\species{V}$ &
$\D{\alpha}\left(\species{X}\given\species{X}\xrightarrow{\eat}\species{X}'\xrightarrow{\parasite}\species{X}''=s\right)$\\

Diversity in habitat $h_1\in\habitats{V}$ relative to another habitat $h_2\in\habitats{V}$ &
$\D{\alpha}\left(\habitats{X}\xrightarrow{\inhabit^\intercal}\individuals{X}\xrightarrow{\belongS}\species{X}\given\habitats{X}=h_1 \relativeto \right.$ $\left.\quad\quad\quad\quad\habitats{X}'\xrightarrow{\inhabit^\intercal}\individuals{X}'\xrightarrow{\belongS}\species{X}'\given\habitats{X}'=h_2 \right)$\\

\hline
\end{tabular} 
    }
    \caption{Network schema of an example from ecology, and table with examples of diversity-related concepts and their expression as network diversity measures.}
    \label{fig:ecology}
  \end{figure}
  
%%%%%%%%%%%%%%%%%%%%%%%%%%%%%%%%%%%%%%%%%%
\subsection{Antitrust and competition law}
\label{subsec:antitrust}

Many developments and applications of concentration measures are found in the economics community, antitrust regulation, and competition law.
As shown in Section~\ref{subsec:diversity_of_diversities}, concentration is a concept for which indices are the reciprocal of those used for the concept of diversity.
Concentration or diversity indices are used to measure the degree to which some firms concentrate the production of units (or the provision of services) in an industry.
Let us consider, for example, the classification and apportionment of tons of steel produced --in a given period of time in a given country-- by the firms that produced them.
From this apportionment or distribution, concentration of the steel industry can be quantitatively measured with diversity measures.
This is the subject of the doctoral thesis of O. C. Herfindahl, for which he developed what is now known as the Herfindahl-Hirschman Index \cite{herfindahl1950concentration}.
The quantitative measurement of concentration of an industry allows for important comparisons to be made by industry regulators, such as, for example, the degree of concentration of an industry should a given merger or acquisition be allowed.

This exercise in measurement of industrial concentration, and the detection and limitation of monopolistic behavior, is made significantly more difficult by the existence of cross-ownership, or cross-control relation between firms.
Cross-ownership refers to situations in which firms from an industry are mutually owned in complex, network-like relations (in a simple example between two firms A and B, firm A owns a part of firm B, and firm B owns a part of firm A).
Cross-control refers, similarly, to situations where firms can name board members of other firms in the same industry producing complex relations of control in a network-like fashion.

Specialized economics literature accounts for many case studies that challenge the application of the aforementioned procedure to regulation \cite{kang1997ownership,kim2014concentrated},
and that address the complex structure of co-ownership networks and their importance in regulation \cite{compston2013network,vitali2011network}.
This makes graph-theoretical approaches good candidates for making advances in the measurement of concentration in industries \cite{levy2009control}.
In particular, the proposed network diversity measures provide tools that allow the measurement of many concepts of interest in antitrust regulation and competition law when dealing with network structures.

To illustrate this, let us consider a heterogeneous information network with three vertex types: that of vertices representing produced units of services $\units{V}$ (\eg, tons of steel, barrels of oil, or clients of portable phone services), that of vertices representing firms $\firms{V}$ that produce those units or provide those services, and that of persons $\persons{V}$ that own the firms.
To model the relations between these entities represented by vertex types, let us consider five edge types: $\produced$, linking each units to the firm that produced them, $\own$, linking firms to the persons that own them, $\crossown$, linking firms with each other according to cross-ownership, and similarly, $\control$ and $\crosscontrol$ linking firms with each other according to control and cross-control (for example, having the right to choose a member of the board of a given firm).
For edge types $\own$, $\crossown$, $\control$, and $\crosscontrol$, the multiplicity of edges can account for the units by which property or control is represented, such as, for example, shares or members of the boards of the firms.
For example, if ownership of a firm is represented by 10 shares, it will have 10 edges, that can belong to edge types $\own$ or $\crossown$.

In this setting the common measurement of industry diversity is expressed as
$$\D{\alpha}\left(\units{X}\xrightarrow{\produced}\firms{X}\right), $$
which becomes the Herfindahl-Hirschman Diversity (reciprocal of the Herfindahl-Hirschman Index) by choosing $\alpha=2$.
Figure~\ref{fig:crossownership} illustrates the heterogeneous information network described in this example, along with a table of concepts related to diversity and expressible using the proposed network diversity measures.

\begin{figure}[!h]
    \centering
    % \begin{minipage}[t]{0.45\linewidth}

  \begin{center}
    \begin{tikzpicture}
      % Names of parts
      \node [draw, circle, minimum width = 1.5cm ] (units) {$\units{V}$};
      \node [draw, circle, minimum width = 1.5cm , right = 1.5cm of units] (firms) {$\firms{V}$};
      \node [draw, circle, minimum width = 1.5cm , right = 1.5cm of firms] (persons) {$\persons{V}$};

      \draw [->] (units) to[edge node={node [midway,above] {$\produced$}}] (firms);
      \draw [->] (firms) to[out=30,in=150,edge node={node [midway,above] {$\control$}}] (persons);
      \draw [->] (firms) to[out=-30,in=-150,edge node={node [midway,below] {$\own$}}] (persons);
      \draw [->] (firms) to[out=110,in=70,distance=1cm,edge node={node [midway,above] {$\crossown$}}] (firms);
      \draw [->] (firms) to[out=-70,in=-110,distance=1cm,edge node={node [midway,below] {$\crosscontrol$}}] (firms);

    \end{tikzpicture}
  \end{center}
% \end{minipage}
    {\scriptsize
    \begin{tabular}{p{8cm}p{6cm}}
Examples of concepts expressible in research questions & Corresponding network diversity measures \\
\hline

Industry diversity with cross-ownership relations &
$\D{\alpha}\left(\units{X}\xrightarrow{\produced}\firms{X}\xrightarrow{\crossown}\firms{X}'\right)$\\

Industry diversity according to persons with cross-ownership relations &
$\D{\alpha}\left(\units{X}\xrightarrow{\produced}\firms{X}\xrightarrow{\crossown}\firms{X}'\xrightarrow{\own}\persons{X}\right)$\\

Diversity of cross-ownership of a firm $f\in\firms{V}$ &
$\D{\alpha}\left(\firms{X}\xrightarrow{\crossown}\firms{X}'\given \firms{X}=f\right)$\\

Diversity of ownership of firms relative to their diversity of control &
$\D{\alpha}\left(\firms{X}\xrightarrow{\own}\persons{X} \relativeto \firms{X}'\xrightarrow{\control}\persons{X}' \right)$\\

\hline
\end{tabular}  
    }
    \caption{Network schema of a cross-ownership and cross-control heterogeneous information network in a setting from \emph{antitrust} regulation or competition law, where products are apportioned in the firms that produced them, which can be cross-owned or cross-controlled by each others.}
    \label{fig:crossownership}
  \end{figure}

\subsection{Scientometrics}
\label{subsec:scientometrics}

Scientometrics, within the field of bibliometrics, studies the measurement and analysis of scientific literature.
Overlapping with information systems, scientometrics study, for example, the \emph{importance} of publications in networks of citations using metrics such as the \emph{Impact Factor} or the \emph{Science Citation Index}.
In networks including other entities such as authors, other measurements include the \emph{h-index}, an index for the productivity and citation impact of scholars.
Recent studies have used heterogeneous information networks to represent data including other entities, such as journals and conferences, in order to extract extended measurements \cite{yan2011p}.

The study of networks modeling and representing scientific production is of interest for other reasons too.
Diversity of topics explored by scientific communities is a concept of interest, for example, in public policy \cite{zitt2008challenges}, and in general for the understanding and description of the structure of scientific communities \cite{zitt2005facing,heimeriks2003mapping}.
Another practical application of the measurement of diversity in citation networks is the maintenance of classification systems~\cite{gomez1996coping}. 

Below, we illustrate the way proposed network diversity measures can address some of the concepts relevant to these areas of research by means of an example.
Let us consider a heterogeneous information network consisting of the following vertex types: authors $\authors{V}$, laboratories $\institutions{V}$ (or affiliation institutions), journals $\journals{V}$, scientific articles $\arts{V}$, keywords used by these articles $\keywords{V}$, and domains of research $\domains{V}$ (\eg, ecology, economics).
We also consider edge types for representing relations between these entities (see Figure~\ref{fig:scientometrics}): affiliation $\aff$ of authors to institutions, edition (or peer-review) $\edit$ of journals by authors, writing $\paperwrite$ of articles by authors, use of keywords $\usekey$ in articles, association of keywords $\belongdomain$ with domains, publication $\publish$ of articles by journals, and declared treatment $\treatdomain$ of research domains by journals.
Figure~\ref{fig:scientometrics} illustrates the corresponding heterogeneous information network, along with a table of concepts related to diversity and expressible using the proposed network diversity measures.

\begin{figure}[!h]
    \centering
    \begin{center}
    \begin{tikzpicture}
      % Names of parts
      \node [draw, circle, minimum width = 1.5cm ] (papers) {$\arts{V}$};
      \node [draw, circle, minimum width = 1.5cm , right = 3cm of papers] (authors) {$\authors{V}$};
      \node [draw, circle, minimum width = 1.5cm , above right = 1.5cm and 1.1cm of papers] (journals) {$\journals{V}$};
      \node [draw, circle, minimum width = 1.5cm , left = 1.5cm of papers] (kws) {$\keywords{V}$};
      \node [draw, circle, minimum width = 1.5cm , right = 1.5cm of authors] (affs) {$\institutions{V}$};
      \node [draw, circle, minimum width = 1.5cm , left = 1.5cm of journals] (doms) {$\domains{V}$};

      \draw [->] (authors) to[edge node={node [midway,above] {$\aff$}}] (affs);
      \draw [->] (authors) to[edge node={node [midway,above] {$\paperwrite$}}] (papers);
      \draw [->] (authors) to[edge node={node [midway,right] {$\edit$}}] (journals);
      \draw [->] (papers) to[out=-70,in=-110,distance=1cm,edge node={node [midway,below] {$\papercite$}}] (papers);
      \draw [->] (papers) to[edge node={node [midway,below] {$\usekey$}}] (kws);
      \draw [->] (kws) to[edge node={node [midway,left] {$\belongdomain$}}] (doms);
      \draw [->] (journals) to[edge node={node [midway,above] {$\treatdomain$}}] (doms);
      \draw [->] (journals) to[edge node={node [midway,left] {$\publish$}}] (papers);

    \end{tikzpicture}
  \end{center}  
    {\scriptsize
    \begin{tabular}{p{6cm}p{9cm}}
Examples of concepts expressible in research questions & Corresponding network diversity measures \\
\hline

Diversity of keywords used by author $a\in\authors{V}$ &
$\D{\alpha}\left(\authors{X}\xrightarrow{\paperwrite}\arts{X}\xrightarrow{\usekey}\keywords{X}\given\authors{X}=a\right)$\\

Diversity of domains addressed in publications by author $a\in\authors{V}$ relative to domains he or she addresses in editing (or peer-reviewing)&
$\D{\alpha}\left(\authors{X}\xrightarrow{\paperwrite}\arts{X}\xrightarrow{\usekey}\keywords{X}\xrightarrow{\belongdomain}\domains{X}\given\authors{X}=a \relativeto \right.$\\
&\hspace{1.5cm}$\quad\quad\quad\quad\left. \authors{X}'\xrightarrow{\edit}\journals{X}\xrightarrow{\treatdomain}\domains{X}'\given\authors{X}'=a \right)$\\

Diversity of domains addressed by citations by authors of laboratory $l\in\institutions{V}$ &
$\D{\alpha}\left(\institutions{X}\xrightarrow{\aff^\intercal}\authors{X}\xrightarrow{\paperwrite}\arts{X}\xrightarrow{\papercite}\arts{X}'\xrightarrow{\usekey}\keywords{X}\xrightarrow{\belongdomain}\domains{X}\given\institutions{X}=l\right)$\\

\hline
\end{tabular}  
    }
    \caption{Network schema of a heterogeneous information network in an example for scientometrics, and table of examples of concepts related to diversity and expressible using the proposed network diversity measures.}
    \label{fig:scientometrics}
  \end{figure}

\section{Conclusions}
\label{sec:part6}

This article presents a formal framework for the measurement of diversity in heterogeneous information networks. 
This allows for the extension of the application of diversity measures from classification modeled by apportioning into distributions, to data represented in network structures.

By presenting a concise theory resulting from the imposition of desirable properties of axioms, we organize diversity measures across a wide spectrum of domains into a family of functions defined by a single parameter $\alpha$: the {\em true diversities}.
Providing a formalism for heterogeneous information networks and constrained random walks on it, we consider different probability distributions on which diversity measures are computed.
These diversity measures are related to the structure of the heterogeneous information network, and thus to the phenomena or objects it represents.
Diversity measures are also related to the different ways in which distributions are computed, which allows us to distinguish several types of diversities: collective, individual, mean individual, backward, relative, and projected diversities.
Some of these network diversities relate to existing measures in the literature, that we framed into a comprehensive framework.
But they also allow for the treatment of new concepts related to diversity in networks.
We provide examples of applications in several domains.

The main contributions of this article are:
\begin{itemize}
\item The proposition of an axiomatic theory of diversity measures that allows us to present most of their uses across several domains with a single-parameter family of functions.
\item The formalization of concepts and tools to describe and process heterogeneous information networks, that have been gaining attention in representation learning and information retrieval communities (in particular in recommender systems).
\item The definition of several \emph{network diversity measures}, resulting from the application of true diversities to probability distributions that are computable with the heterogeneous information network formalism.
These network diversity measures allow for the referentiation, expression, and computation of concepts relevant to diversity in networks, extending the use of diversity from systems of classification and apportionment to systems best described by network-structured data.
\item The mapping of some of the \emph{network diversity measures} to pre-existing quantitative measurements that are widespread in different fields, and the development of new applications through examples in recommender systems, social media studies, ecology, competition law, and scientometrics.
\end{itemize}

In addition to providing means of referencing, expression, and computation on diversity-related concepts in complex data modeled by heterogeneous information networks, the \emph{network diversity measures} could be leveraged in different downstream tasks performed in data mining.
Expanding on those mentioned in the introduction (Section~\ref{sec:part1}), we can now hint at more precise examples of applications in such tasks.
In Recommender Systems, for example, previous works have used diversities associated with meta paths to avoid over-fitting in the training stage on data modeled with heterogeneous information networks \cite{liu2019user}. 
Network diversity measures could allow, in this case, to target different collective and individual diversities in the function to be optimized.
Other applications use meta paths to consider node similarities in strategies for node classification \cite{yin2017flexible}, and could leverage network diversity measures to consider parametrization of the weight given to balance and variety in diversity when used.
Research in representation learning could embed different quantities computable with network diversity measures in learning strategies where diversity would be leveraged for optimizing different objective function through training \cite{zhang2019shne}.
Finally, we hint to other applications that would need to be explored in greater detail than it is possible here, in domains related to heterogeneous information networks and diversity, such as community detection \cite{cruz2011entropy,nikolaev2015efficient,vzalik2018memetic}, clustering in networks \cite{wang2007entropy,kenley2011entropy}, and the analysis of time-series resulting from temporal networks \cite{gu2005detecting}.

% [next lines were added after revisions of december 2020]
Many relevant advances in computer sciences hinge on the improvement of performance metrics through the development of novel algorithms and methodologies.
This article seeks to contribute in the proposal of the metrics with which advacements are to be compared with the state of art.
It is for this purpose what we propose a detailed discussion of the properties and implications of the proposed network diversity measures.
% [end of addition]
We hope that this framework for the application of diversity measures to network structures will enrich research on diversity in the domains identified in Section~\ref{sec:part5} and beyond.
Future developments in this line of research might consider the identification of algebraic structures for network diversity measures, and their application to case studies.

%%% Local Variables:
%%% mode: latex
%%% TeX-master: "main"
%%% End:

\section*{Acknowledgement}

This work has been partially funded by the European Commission H2020 FETPROACT 2016-2017 program under grant 732942 (ODYCCEUS) and by the French National Agency of Research (ANR) under grant ANR-15-CE38-0001 (AlgoDiv).

The authors would like to sincerely thank H\^ong-Lan Botterman and Matthieu Latapy for their comments, as well as Claire Schaffer for her thorough proof-reading.

\bibliographystyle{elsarticle-num} 
\bibliography{references}

\end{document}